\documentclass{article}
\usepackage{ijcai21}

\usepackage{times}
\usepackage{soul}
\usepackage{url}
\usepackage{xr-hyper}
\makeatletter
\newcommand*{\addFileDependency}[1]{
  \typeout{(#1)}
  \@addtofilelist{#1}
  \IfFileExists{#1}{}{\typeout{No file #1.}}
}
\makeatother

\newcommand*{\myexternaldocument}[1]{
    \externaldocument{#1}
    \addFileDependency{#1.tex}
    \addFileDependency{#1.aux}
}

\myexternaldocument{supplementary}

\usepackage[hidelinks]{hyperref}
\hypersetup{
    colorlinks=true,    
    anchorcolor=black,
    citecolor=black,
    menucolor=black,
    linkcolor=black,
    filecolor=black,      
    urlcolor=cyan,
}
\usepackage[utf8]{inputenc}
\usepackage[small]{caption}
\usepackage{graphicx}
\usepackage{booktabs}
\usepackage{amssymb,dsfont,amsthm,amsmath}
\usepackage{mathtools}
\usepackage{xspace}
\usepackage{multirow}
\usepackage{listings}
\usepackage{bm}
\usepackage{subfig}
\usepackage{array}
\usepackage{eqparbox}
\usepackage{etoolbox}  
\usepackage{todonotes}
\presetkeys{todonotes}{inline,backgroundcolor=red!40,caption={}}{}

\DeclareMathOperator{\E}{\mathbb{E}}
\usepackage[linesnumbered, ruled]{algorithm2e}

\usepackage{floatrow}  
\usepackage[title]{appendix}

\usepackage{natbib}

\title{DEHB: Evolutionary Hyperband for Scalable, Robust and Efficient Hyperparameter Optimization\thanks{Proceedings of IJCAI-21}}

\author{
Noor Awad$^1$\and
Neeratyoy Mallik$^1$\and
Frank Hutter$^{1,2}$
\affiliations
$^1$Department  of  Computer  Science,  University  of  Freiburg, Germany\\
$^2$Bosch Center for Artificial Intelligence, Renningen, Germany
\emails
\{awad, mallik, fh\}@cs.uni-freiburg.de
}

\begin{document}

\maketitle

\begin{abstract}
Modern machine learning algorithms crucially rely on several design decisions to achieve strong performance, making the problem of Hyperparameter Optimization (HPO) more important than ever.
%
%
Here, we combine the advantages of the popular bandit-based HPO method Hyperband (HB) and the evolutionary search approach of Differential Evolution (DE) to yield a new HPO method which we call DEHB. 
%
Comprehensive results on a very broad range of HPO problems, 
as well as a wide range of tabular benchmarks from neural architecture search,
demonstrate that DEHB achieves strong performance far more robustly than all previous HPO methods we are aware of, especially for high-dimensional problems with discrete input dimensions. For example, DEHB is up to $1000\times$ faster than random search.
It is also efficient in computational time, conceptually simple and easy to implement, positioning it well to become a new default HPO method.     
\end{abstract}

\section{Introduction}
\label{sec:introduction}
Many algorithms in artificial intelligence rely crucially on good settings of their hyperparameters to achieve strong performance. This is particularly true for deep learning~\citep{henderson-aaai18,melis-iclr18a}, where dozens of hyperparameters concerning both the neural architecture and the optimization \& regularization pipeline need to be instantiated.
At the same time, modern neural networks continue to get larger and more computationally expensive, making the need for efficient hyperparameter optimization (HPO) ever more important.

We believe that a practical, general HPO method must fulfill many desiderata, including: (1) strong anytime performance, (2) strong final performance with a large budget, (3) effective use of parallel resources, (4) scalability w.r.t.\ the  dimensionality and (5) robustness \& flexibility.   
These desiderata drove the development of BOHB~\citep{falkner-icml18a}, which satisfied them by combining the best features of Bayesian optimization via Tree Parzen estimates (TPE)~\citep{bergstra-nips11a} (in particular, strong final performance), and the many advantages of bandit-based HPO via Hyperband~\citep{li-iclr17a}. While BOHB is among the best general-purpose HPO methods we are aware of, it still has problems with optimizing discrete dimensions and does not scale as well to high dimensions as one would wish. Therefore, it does not work well on high-dimensional HPO problems with discrete dimensions and also has problems with tabular neural architecture search (NAS) benchmarks (which can be tackled as high-dimensional discrete-valued HPO benchmarks, an approach followed, e.g., by regularized evolution (RE) ~\citep{real2019regularized}).

The main contribution of this paper is to further improve upon BOHB to devise an effective general HPO method, which we dub \emph{DEHB}. DEHB is based on a combination of the evolutionary optimization method of differential evolution (DE ~\citep{storn1997differential}) and Hyperband and has several useful properties:
\begin{enumerate}
    \item DEHB fulfills all the desiderata of a good HPO optimizer stated above, and in particular achieves more robust \textit{strong final performance} than BOHB, especially for high-dimensional and discrete-valued problems.
    \item DEHB is \textit{conceptually simple} and can thus be easily re-implemented in different frameworks.
    \item DEHB is \textit{computationally cheap}, not incurring the overhead typical of most BO methods.
    \item DEHB effectively takes advantage of parallel resources. 
\end{enumerate}


After discussing related work (\hyperref[sec:related-work]{Section 2}) and 
background on DE and Hyperband (\hyperref[sec:background]{Section 3}), \hyperref[sec:DEHB-details]{Section 4} describes our new DEHB method in detail. \hyperref[sec:experiments]{Section 5} then presents comprehensive experiments on artificial toy functions, surrogate benchmarks, Bayesian neural networks, reinforcement learning, and 13 different tabular neural architecture search benchmarks, demonstrating that DEHB is more effective and robust than a wide range of other HPO methods, and in particular up to $1000\times$ times faster than random search (Figure \ref{fig:sub-adult}) and up to $32\times$ times faster than BOHB (Figure \ref{fig:sub-201-imagenet}) on HPO problems; on toy functions, these speedup factors even reached $33\,440\times$ and $149\times$, respectively (Figure \ref{fig:sub32+32}).

\section{Related Work}
\label{sec:related-work}
HPO as a black-box optimization problem can be broadly tackled using two families of methods: model-free methods, such as evolutionary algorithms, and model-based Bayesian optimization methods.
\textit{Evolutionary Algorithms} (EAs) are model-free population-based methods which generally include a method of initializing a population; mutation, crossover, selection operations; and a notion of fitness. 
EAs are known for black-box optimization in a HPO setting since the 1980s~\citep{grefenstette-ga}.
They have also been popular for designing architectures of deep neural networks~\citep{angeline1994evolutionary,xie2017genetic,real2017large,liu2017hierarchical}; recently, Regularized Evolution~(RE)~\citep{real2019regularized} achieved state-of-the-art results on ImageNet. 



Bayesian optimization (BO) uses a probabilistic model based on the already observed data points to model the objective function and to trade off exploration and exploitation.
The most commonly used probabilistic model in BO are Gaussian processes (GP) since they obtain well-calibrated and smooth uncertainty estimates~\citep{snoek2012practical}. However, GP-based models have high complexity, do not natively scale well to high dimensions and do not apply to complex spaces without prior knowledge; alternatives include tree-based methods~\citep{bergstra-nips11a,hutter-lion11a} and Bayesian neural networks~\citep{springenberg2016bayesian}. 

Recent so-called \emph{multi-fidelity} methods exploit cheap approximations of the objective function to speed up the optimization~\citep{liu2016multi,wang2017generic}. Multi-fidelity optimization is also popular in BO, with Fabolas~\citep{klein2016fast} and Dragonfly~\citep{dragonfly2020} being GP-based examples. The popular method BOHB~\citep{falkner-icml18a}, which combines BO and the bandit-based approach Hyperband~\citep{li-iclr17a}, has been shown to be a strong off-the-shelf HPO method and to the best of our knowledge is the best previous off-the-shelf multi-fidelity optimizer.

\section{Background}
\label{sec:background}
\subsection{Differential Evolution (DE)}
In each generation $g$, DE uses an evolutionary search based on difference vectors to generate new candidate solutions. DE is a population-based EA which uses three basic iterative steps (mutation, crossover and selection). At the beginning of the search on a $D$-dimensional problem, we initialize a population of $N$ individuals $x_{i,g}=(x^{1}_{i,g},x^{2}_{i,g},...,x^{D}_{i,g})$ randomly within the search range of the problem being solved. 
Each individual $x_{i,g}$ is evaluated by computing its corresponding objective function value. Then the mutation operation generates a new offspring for each individual.
The canonical DE uses a mutation strategy called \textit{rand/1}, which selects three random parents $x_{r_1}, x_{r_2}, x_{r_3}$ to generate a new mutant vector $v_{i,g}$ for each $x_{i,g}$ in the population as shown in Eq. \ref{Eq.mutation} where $F$ is a scaling factor parameter and takes a value within the range (0,1]. 
\begin{equation}
 v_{i,g}= x_{r_1,g} + F\cdot(x_{r_2,g} - x_{r_3,g}).
\label{Eq.mutation}
\end{equation}%

The crossover operation then combines each individual $x_{i,g}$ and its corresponding mutant vector $v_{i,g}$ to generate the final offspring/child $u_{i,g}$. The canonical DE uses a simple binomial crossover to select values from $v_{i,g}$ with a probability $p$ (called crossover rate) and $x_{i,g}$ otherwise.
For the members $x_{i,g+1}$ of the next generations, DE then uses the better of $x_{i,g}$ and $u_{i,g}$.
More details on DE can be found in appendix \ref{More-detail-DE}.      

\subsection{Successive Halving (SH) and Hyperband (HB)}
Successive Halving (SH)~\citep{jamieson-aistats16a} is a simple yet effective multi-fidelity optimization method that exploits the fact that, for many problems, low-cost approximations of the expensive blackbox functions exist, which can be used to rule out poor parts of the search space at little computational cost. Higher-cost approximations are only used for a small fraction of the configurations to be evaluated. Specifically, an \emph{iteration} of SH starts by sampling $N$ configurations uniformly at random, evaluating them at the lowest-cost approximation (the so-called lowest \emph{budget}), and forwarding a fraction of the top $1/\eta$ of them to the next budget (function evaluations at which are expected to be roughly $\eta$ more expensive). This process is repeated until the highest budget, used by the expensive original blackbox function, is reached. Once the runs on the highest budget are complete, the current SH iteration ends, and the next iteration starts with the lowest budget.
We call each such fixed sequence of evaluations from lowest to highest budget a \textit{SH bracket}.
While SH is often very effective, it is not guaranteed to converge to the optimal configuration even with infinite resources, because it can drop poorly-performing configurations at low budgets that actually might be the best with the highest budget.

Hyperband (HB)~\citep{li-iclr17a} solves this problem by hedging its bets across different instantiations of SH with successively larger lowest budgets, thereby being provably at most a constant times slower than random search. 
In particular, this procedure also allows to find configurations that are strong for higher budgets but would have been eliminated for lower budgets.
Algorithm \ref{alg:HB} in Appendix \ref{More-detail-HB} shows the pseudocode for HB with the SH subroutine.
One iteration of HB (also called \emph{HB bracket}) can be viewed as a sequence of SH brackets with different starting budgets and different numbers of configurations for each SH bracket. The precise budgets and number of configurations per budget are determined by HB given its 3 parameters: \textit{minimum budget}, \textit{maximum budget}, and $\eta$. 

The main advantages of HB are its simplicity, theoretical guarantees, and strong anytime performance compared to optimization methods operating on the full budget. However, HB can perform worse than BO and DE for longer runs since it only selects configurations based on random sampling and does not learn from previously sampled configurations.

\section{DEHB}
\label{sec:DEHB-details}
We design DEHB 
to satisfy all the desiderata described in the introduction (Section \ref{sec:introduction}). DEHB inherits several advantages from HB to satisfy some of these desiderata, including its strong anytime performance, scalability and flexibility. From the DE component, it inherits robustness, simplicity, and computational efficiency. 
We explain DEHB in detail in the remainder of this section; full pseudocode can be found in Algorithm \ref{alg:dehb} in Appendix \ref{More-detail-DEHB}.

\begin{figure}
\centering
\includegraphics[width=0.9\columnwidth]{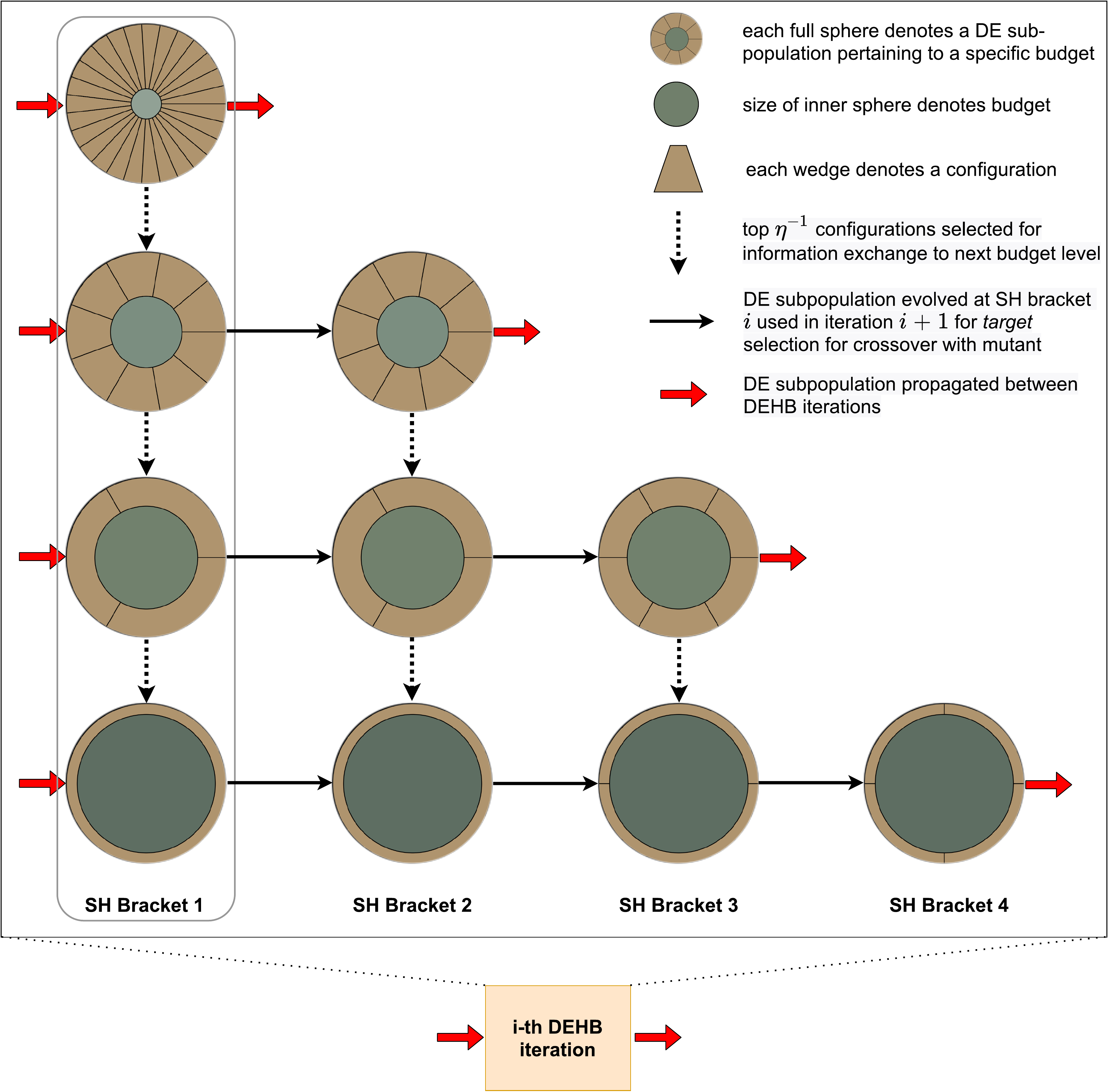}
\caption{Internals of a DEHB iteration showing information flow across fidelities (top-down), and how each subpopulation is updated in each DEHB iteration (left-right).}
\label{fig:DEHB-bracket}
\end{figure} 

\subsection{High-Level Overview} 
A key design principle of DEHB is to share information across the runs it executes at various budgets.
DEHB maintains a \textit{subpopulation} for each of the budget levels, where the \textit{population size} for each subpopulation is assigned as the maximum number of function evaluations HB allocates for the corresponding budget. 

We borrow nomenclature from HB and call the HB iterations that DEHB uses \textit{DEHB iterations}.
Figure \ref{fig:DEHB-bracket} illustrates one such iteration, where \textit{minimum budget}, \textit{maximum budget}, and $\eta$ are $1$, $27$, and $3$, respectively. The topmost sphere for \textit{SH Bracket 1}, is the first step, where $27$ configurations are sampled uniformly at random and evaluated at the lowest budget $1$. These evaluated configurations now form the DE subpopulation associated with budget $1$. The dotted arrow pointing downwards indicates that the top-$9$ configurations ($27/\eta$) are \textit{promoted} to be evaluated on the next higher budget $3$ to create the DE subpopulation associated with budget $3$, and so on until the highest budget. This progressive increase of the budget by $\eta$ and decrease of the number of configurations evaluated by $\eta$ is simply the vanilla SH. Indeed, each SH bracket for this first DEHB iteration is basically executing vanilla SH, starting from different minimum budgets, just like in HB. 

One difference from vanilla SH is that random sampling of configurations occurs only once: in the first step of the first SH bracket of the first DEHB iteration. Every subsequent SH bracket 
begins by reusing the subpopulation updated in the previous SH bracket, and carrying out a DE evolution (detailed in Section \ref{dehb-sh-de}). For example, for SH bracket 2 in Figure \ref{fig:DEHB-bracket}, the subpopulation of $9$ configurations for budget $3$ (topmost sphere) is propagated from SH bracket 1 and undergoes evolution. The top $3$ configurations ($9/\eta$) then affect the population for the next higher budget $9$ of SH bracket 2. Specifically, these  
will used as the so-called parent pool for that higher budget, using the modified DE evolution to be discussed in Section \ref{dehb-sh-de}.
The end of \textit{SH Bracket 4} marks the end of this DEHB iteration. 
We dub DEHB's first iteration its \emph{initialization iteration}.
At the end of this iteration, all DE subpopulations associated with the higher budgets are seeded with configurations that performed well in the lower budgets. In subsequent SH brackets, no random sampling occurs anymore, and the search runs separate DE evolutions at different budget levels, where information flows from the subpopulations at lower budgets to those at higher budgets through the modified DE mutation (Fig. \ref{fig:mutation-types}). 

\begin{figure}
\centering
\includegraphics[width=0.8\columnwidth]{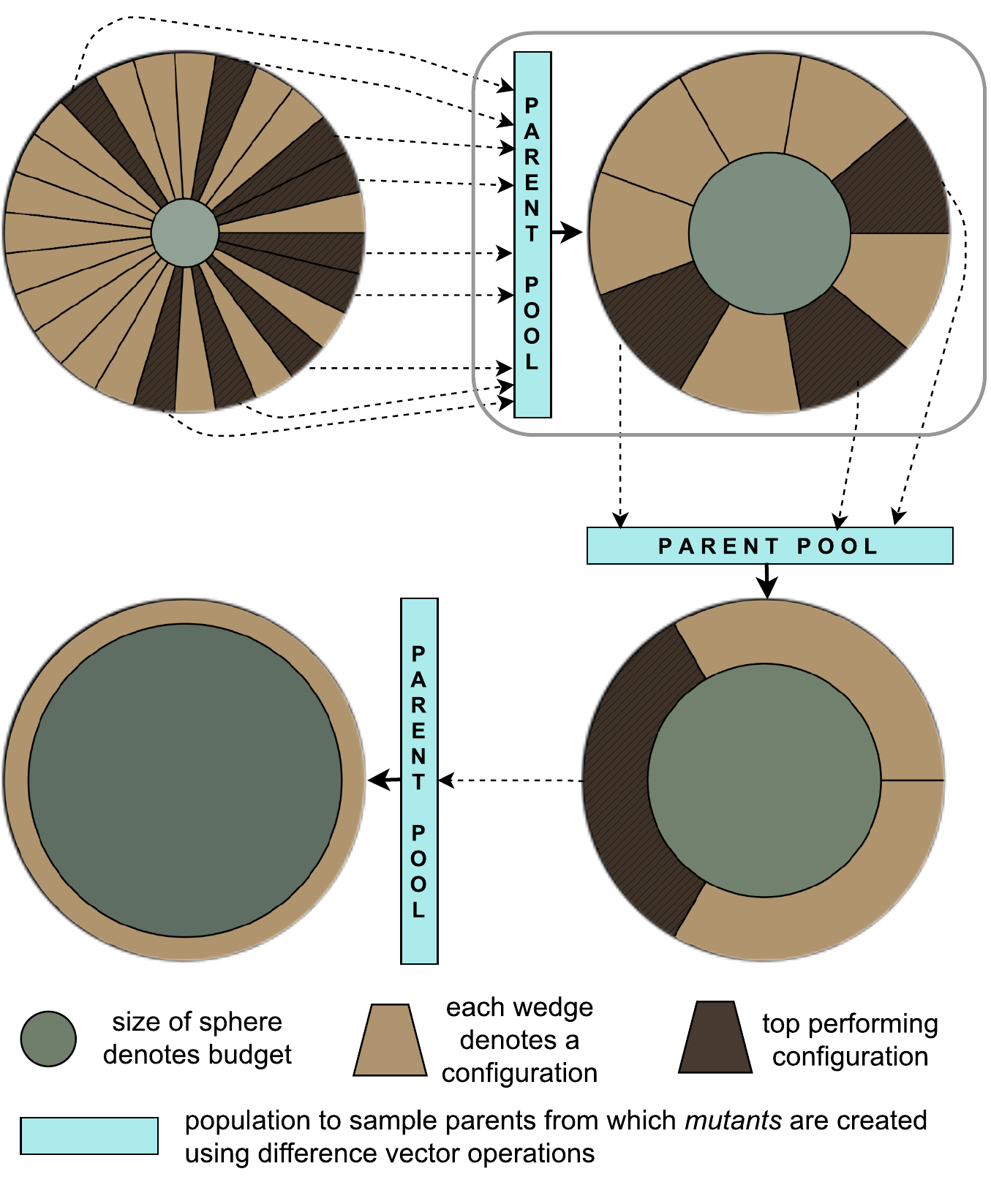}
\caption{\label{fig:SH-using-DE} Modified SH routine under DEHB}
\end{figure}

\subsection{Modified Successive Halving using DE Evolution}
\label{dehb-sh-de}
\begin{figure}
\centering
\includegraphics[width=0.85\columnwidth]{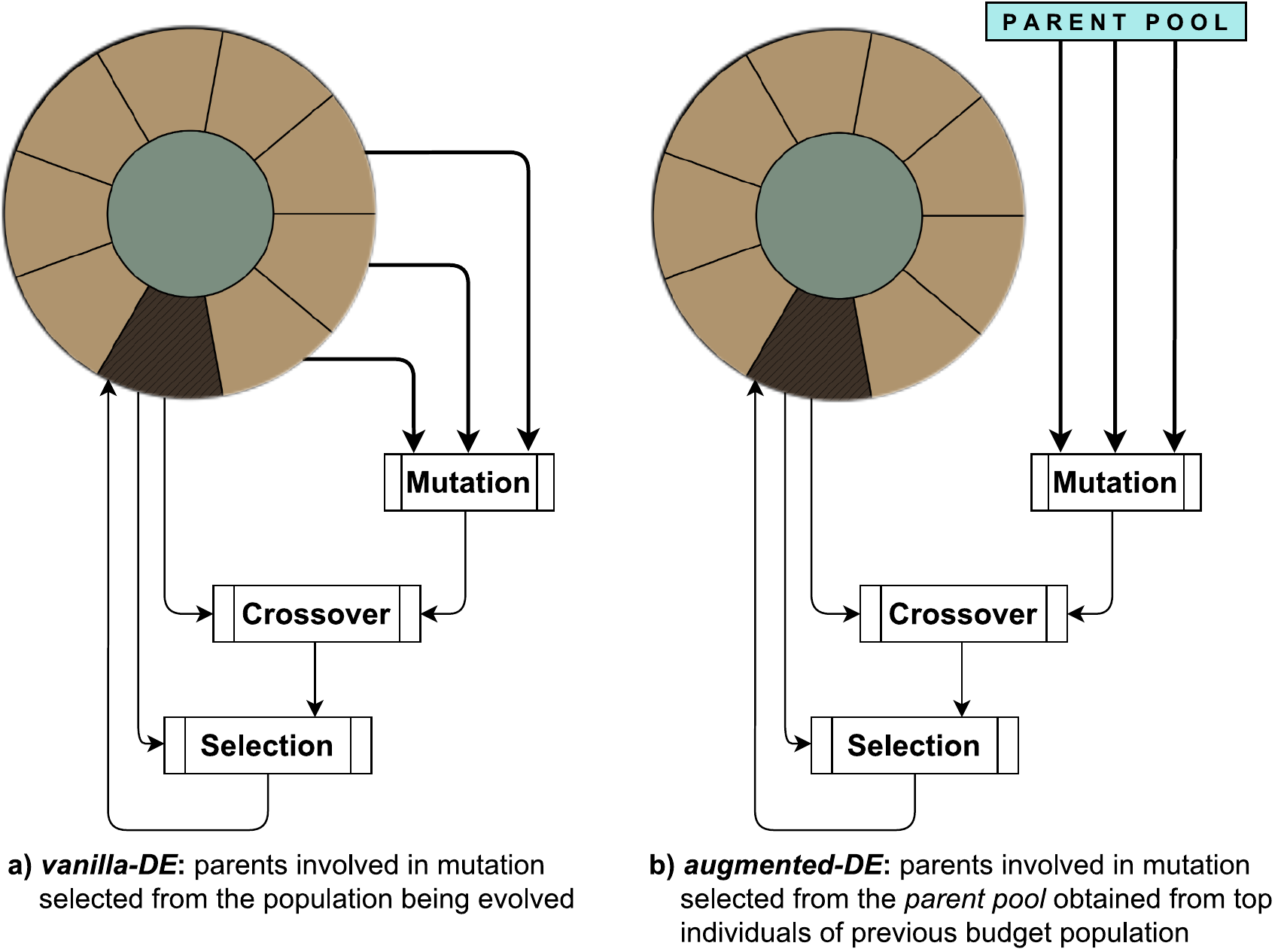}
\caption{\label{fig:mutation-types} Modified DE evolution under DEHB}
\end{figure}

We now discuss the deviations from vanilla SH by elaborating on the design of a SH bracket inside DEHB, highlighted with a box in Figure \ref{fig:DEHB-bracket} (SH Bracket 1). 
In DEHB, the top-performing configurations from a lower budget are not simply promoted and evaluated on a higher budget (except for the \textit{Initialization} SH bracket). 
Rather, in DEHB, the top-performing configurations are collected in a \textit{Parent Pool} (Figure \ref{fig:SH-using-DE}). This pool is responsible for transfer of information from a lower budget to the next higher budget, but not by directly suggesting best configurations from the lower budget for re-evaluation at a higher budget. Instead, the parent pool represents a good performing \textit{region} w.r.t.\ the lower budget, from which \textit{parents} can be sampled for mutation. 
Figure \ref{fig:mutation-types}b demonstrates how a parent pool contributes in a DE evolution in DEHB.
Unlike in vanilla DE (Figure \ref{fig:mutation-types}a), in DEHB, the mutants involved in DE evolution are extracted from the \textit{parent pool} instead of the population itself. This allows the evolution to incorporate and combine information from the current budget, and also from the decoupled search happening on the lower budget. The \textit{selection} step as shown in Figure \ref{fig:mutation-types} is responsible for updating the current subpopulation if the new suggested configuration is better. 
If not, the existing configuration is retained in the subpopulation. This guards against cases where performance across budget levels is not correlated and good configurations from lower budgets do not improve higher budget scores. However, search on the higher budget can still progress, as the first step of every SH bracket performs vanilla DE evolution (there is no parent pool to receive information from). Thereby, search at the required budget level progresses even if lower budgets are not informative.

Additionally, we also construct a \textit{global population pool} consisting of configurations from all the subpopulations. This pool does not undergo any evolution and serves as the parent pool in the edge case where the parent pool is smaller than the minimum number of individuals required for the mutation step. For the example in Figure \ref{fig:SH-using-DE}, under the \textit{rand1} mutation strategy (which requires three parents), we see that for the highest budget, only one configuration ($3/\eta$) is included from the previous budget. 
In such a scenario, the additional two required parents are sampled from the global population pool.

\subsection{DEHB efficiency and parallelization}

\begin{figure}
\centering
\floatbox[{\capbeside\thisfloatsetup{capbesideposition={right,top},capbesidewidth=4cm}}]{figure}[\FBwidth]
{\caption{\label{fig:cifar10-speed} Runtime comparison for DEHB and BOHB based on a single run on the Cifar-10 benchmark from NAS-Bench-201. The $x$-axis shows the actual cumulative wall-clock time spent by the algorithm (optimization time) in between the function evaluations.}}
{\includegraphics[width=4cm]{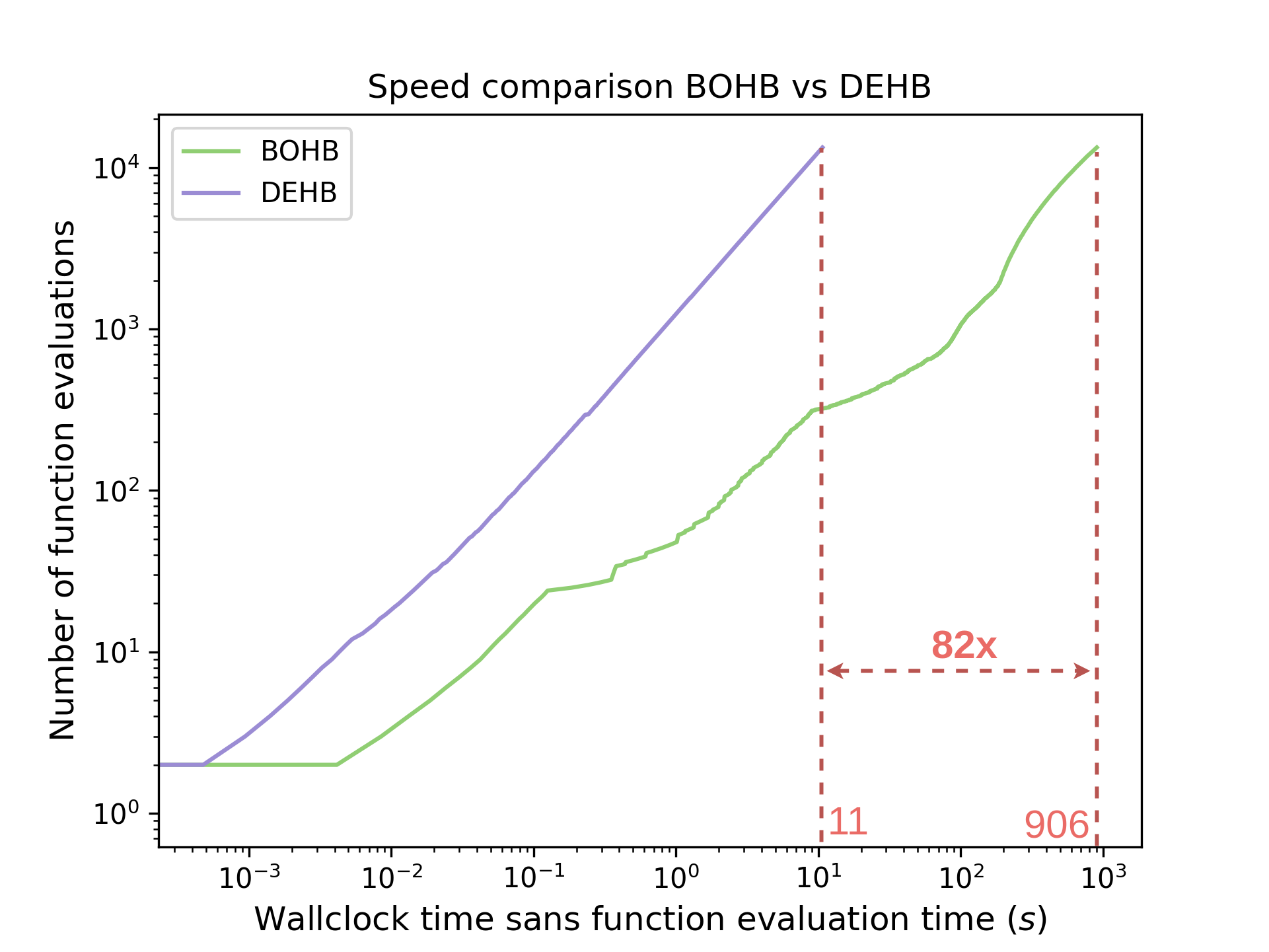}}
\end{figure}

As mentioned previously, DEHB carries out separate DE searches at each budget level.
Moreover, the DE operations involved in evolving a configuration are constant in operation and time. Therefore, DEHB's runtime overhead does not grow over time, even as the number of performed function evaluations increases; this is in stark contrast to model-based methods, whose time complexity is often cubic in the number of performed function evaluations. Indeed, Figure \ref{fig:cifar10-speed} demonstrates that, for a tabular benchmark with negligible cost for function evaluations, DEHB is almost 2 orders of magnitude faster than BOHB to perform 13336 function evaluations. GP-based Bayesian optimization tools would require approximations to even fit a single model with this number of function evaluations.

\begin{figure}
\centering
\floatbox[{\capbeside\thisfloatsetup{capbesideposition={right,top},capbesidewidth=4cm}}]{figure}[\FBwidth]
{\caption{\label{fig:parallel-DEHB} Results for the OpenML Letter surrogate benchmark where $n$ represents number of workers that were used for each DEHB run. Each trace is averaged over $10$ runs.}}
{\includegraphics[width=4cm]{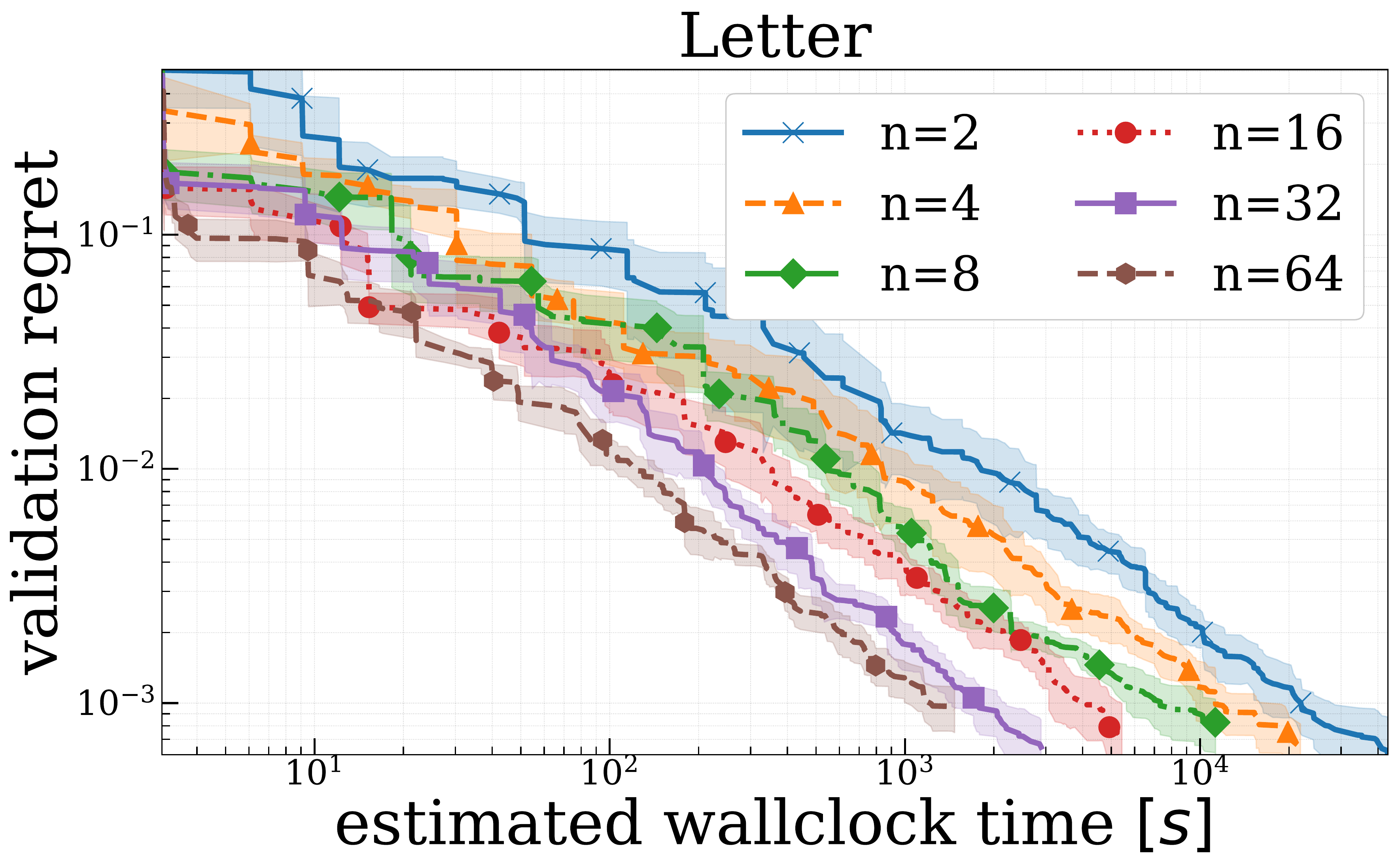}}
\end{figure}

We also briefly describe a parallel version of DEHB (see Appendix \ref{app:parallel} for details of its design). Since DEHB can be viewed as a sequence of predetermined SH brackets, the SH brackets can be asynchronously distributed over free workers. A central \textit{DEHB Orchestrator} keeps a single copy of all DE subpopulations, allowing for asynchronous, \textit{immediate} DE evolution updates. Figure \ref{fig:parallel-DEHB} illustrates that this parallel version achieves linear speedups for similar final performance.

\section{Experiments}
\label{sec:experiments}
We now comprehensively evaluate DEHB, illustrating that it is more robust and efficient than any other HPO method we are aware of.
To keep comparisons fair and reproducible, we use a broad collection of publicly-available HPO and NAS benchmarks:
all HPO benchmarks that were used to demonstrate the strength of BOHB~\citep{falkner-icml18a}\footnote{We leave out the $2$-dimensionsal SVM surrogate benchmarks since all multi-fidelity algorithms performed similarly for this easy task, without any discernible difference.} and also a broad collection of 13 recent tabular NAS benchmarks represented as HPO problems~\citep{awad2020iclr}.  

In this section, to avoid cluttered plots we present a focused comparison of DEHB with BOHB, the best previous off-the-shelf multi-fidelity HPO method we are aware of, which has in turn outperformed a broad range of competitors (GP-BO, TPE, SMAC, HB, Fabolas, MTBO, and HB-LCNet) on these benchmarks~\citep{falkner-icml18a}.
For reference, we also include the obligatory random search (RS) baseline in these plots, showing it to be clearly dominated, with up to 1000-fold speedups.
We also 
provide a comparison against a broader range of methods at the end of this section (see Figure \ref{fig:ranking} and Table \ref{table:rank-summary}), with a full comparison in Appendix \ref{sec:app-exp}. We also 
compare to the recent GP-based multi-fidelity BO tool Dragonfly in Appendix \ref{sec:app-bo-exps}.
Details for the hyperparameter values of the used algorithms can be found in Appendix \ref{baseline-algs}. 

We use the same parameter settings for mutation factor $F=0.5$ and crossover rate $p=0.5$ for both DE and DEHB. The population size for DEHB is not user-defined but set by its internal Hyperband component while we set it to 20 for DE following~\citep{awad2020iclr}.
Unless specified otherwise, we report results from \textit{$50$ runs} for all algorithms, plotting the validation regret\footnote{This is the difference of validation score from the global best.} over the cumulative cost incurred by the function evaluations, and ignoring the optimizers' overhead in order to not give DEHB what could be seen as an unfair advantage.\footnote{Shaded bands in plots represent the standard error of the mean.} We also show the speedups that DEHB achieves compared to RS and BOHB, where this is possible without adding clutter.



\subsection{Artificial Toy Function: Stochastic Counting Ones}\label{sec:exp-counting}
This toy benchmark by~\citet{falkner-icml18a} is useful to assess scaling behavior and ability to handle binary dimensions. 
The goal is to minimize the following objective function:
\begin{equation*}
     f(x) = - \left(\sum_{x \in X_{cat}} x + \sum_{x \in X_{cont}} \E_{b}[(B_{p=x})] \right),
\end{equation*}
where the sum of the categorical variables ($x_i \in \{0,1\}$) represents the standard discrete counting ones problem. The continuous variables ($x_j \in [0,1]$) represent the stochastic component, with the budget \textit{b} controlling the noise. The budget here represents the number of samples used to estimate the mean of the Bernoulli distribution ($B$) with parameters $x_j$. 
Following~\citet{falkner-icml18a}, we run 4 sets of experiments with $N_{cont}=N_{cat}=\{4, 8, 16, 32\}$, where $N_{cont}=|X_{cont}|$ and $N_{cat}=|X_{cat}|$, using the same budget spacing and plotting the normalized regret: $(f(x) + d)/d$, where $d = N_{cat} + N_{cont}$. Although this is a toy benchmark it can offer interesting insights since the search space has mixed binary/continuous dimensions which
DEHB handles well (refer to \ref{mixed-types} in Appendix for more details).
In Figure \ref{fig:sub32+32}, we consider the $64$-dimensional space $N_{cat}=N_{cont}=32$; results for the lower dimensions can be found in Appendix \ref{sec:app-counting}.
Both BOHB and DEHB begin with a set of randomly sampled individuals evaluated on the lowest budget. It is therefore unsurprising that in Figure \ref{fig:sub32+32} (and in other experiments too), these two algorithms follow a similar optimization trace at the beginning of the search. Given the high dimensionality, BOHB requires many more samples to switch to model-based search which slows its convergence in comparison to the lower dimensional cases ($N_{cont}=N_{cat}=\{4, 8, 16\}$). In contrast, DEHB's convergence rate is almost agnostic to the increase in dimensionality.
\begin{figure}
\centering
\floatbox[{\capbeside\thisfloatsetup{capbesideposition={right,top},capbesidewidth=4cm}}]{figure}[\FBwidth]
{\caption{\label{fig:sub32+32} Results for the Stochastic Counting Ones problem in $64$ dimensional space with $32$ categorical and $32$ continuous hyperparameters. All algorithms shown were run for $50$ runs.}}
{\includegraphics[width=4cm]{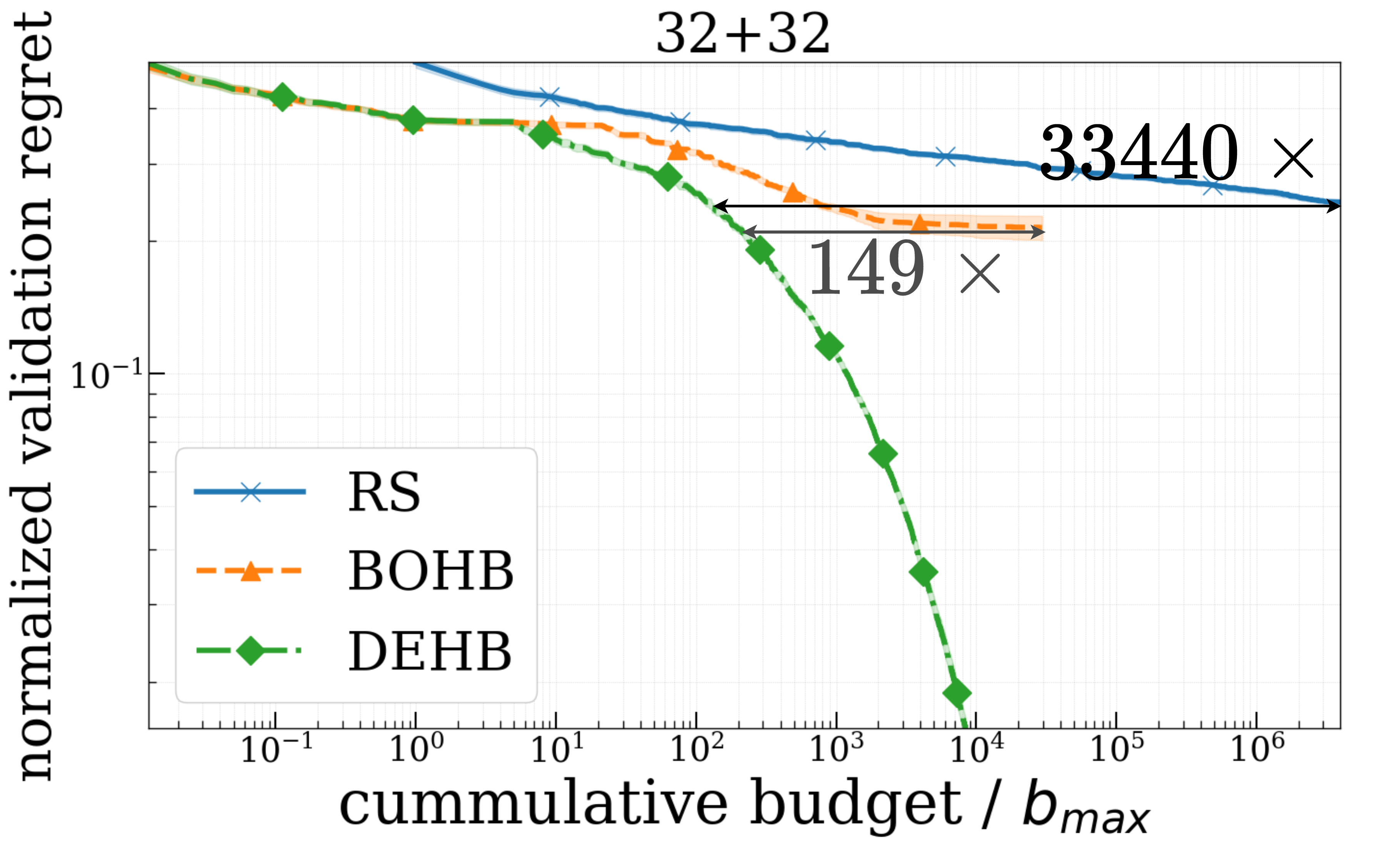}}
\end{figure}

\subsection{Surrogates for Feedforward Neural Networks}
\begin{figure}
\centering
\floatbox[{\capbeside\thisfloatsetup{capbesideposition={right,top},capbesidewidth=4cm}}]{figure}[\FBwidth]
{\caption{Results for the OpenML Adult surrogate benchmark for $6$ continuous hyperparameters for $50$ runs of each algorithm.}\label{fig:sub-adult}}
{\includegraphics[width=4cm]{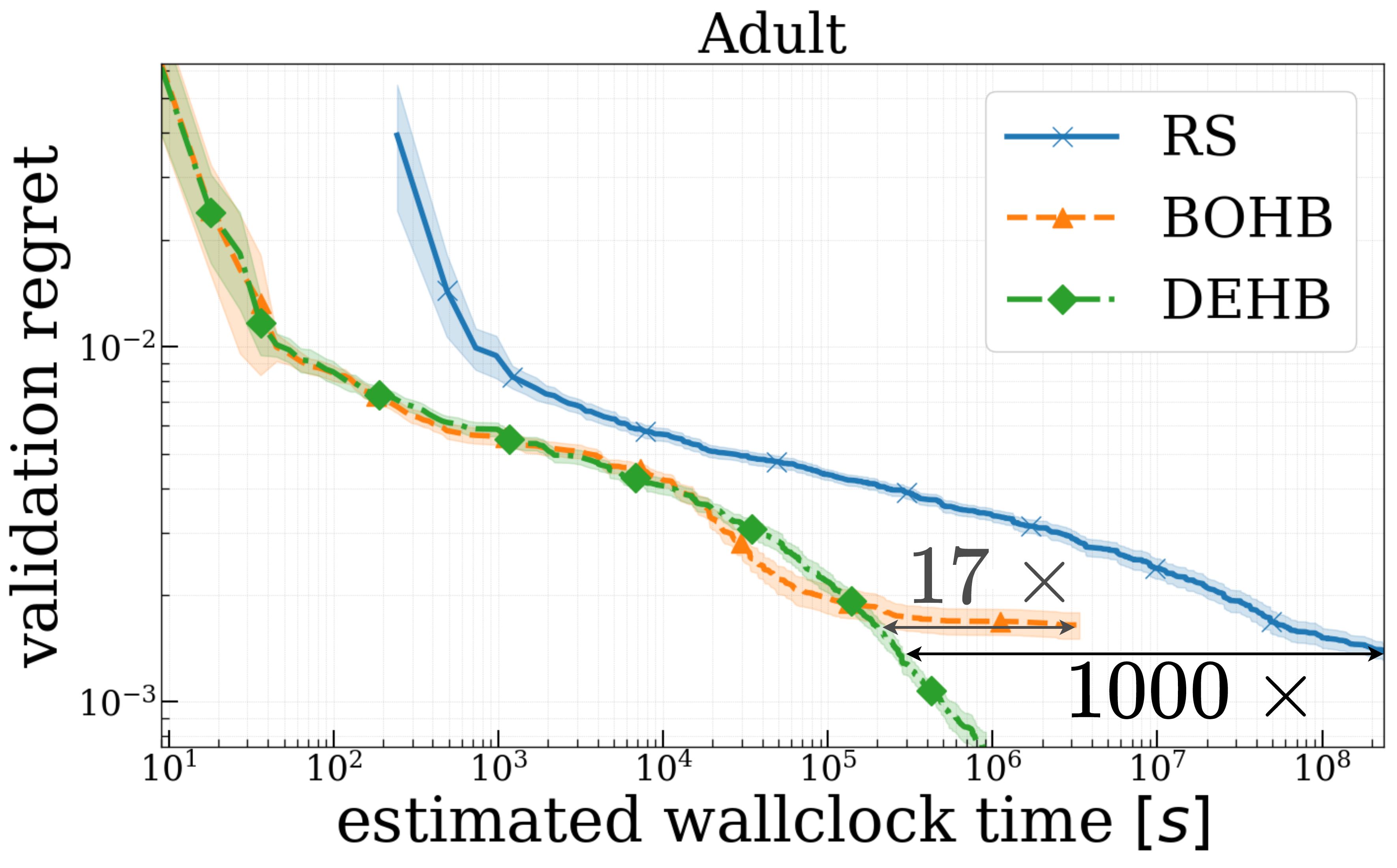}}
\end{figure}

In this experiment, we optimize six architectural and training hyperparameters of a feed-forward neural network on six different datasets from OpenML~\citep{vanschoren2014openml}, using a surrogate benchmark built by~\citet{falkner-icml18a}.
%
The budgets are the training epochs for the neural networks. 
For all six datasets, we observe a similar pattern of the search trajectory, with DEHB and BOHB having similar anytime performance and DEHB achieving the best final score. 
An example is given in 
Figure \ref{fig:sub-adult}, also showing a 1000-fold speedup over random search; qualitatitvely similar results for the other 5 datasets are in Appendix \ref{sec:app-openml}. 

\subsection{Bayesian Neural Networks}\label{sec:exp-bnn}
\begin{figure}
\centering
\floatbox[{\capbeside\thisfloatsetup{capbesideposition={right,top},capbesidewidth=4cm}}]{figure}[\FBwidth]
{\caption{Results for tuning $5$ hyperparameters of a Bayesian Neural Network on the Boston Housing regression dataset for $50$ runs each.}\label{fig:sub-bnnboston}}
{\includegraphics[width=4cm]{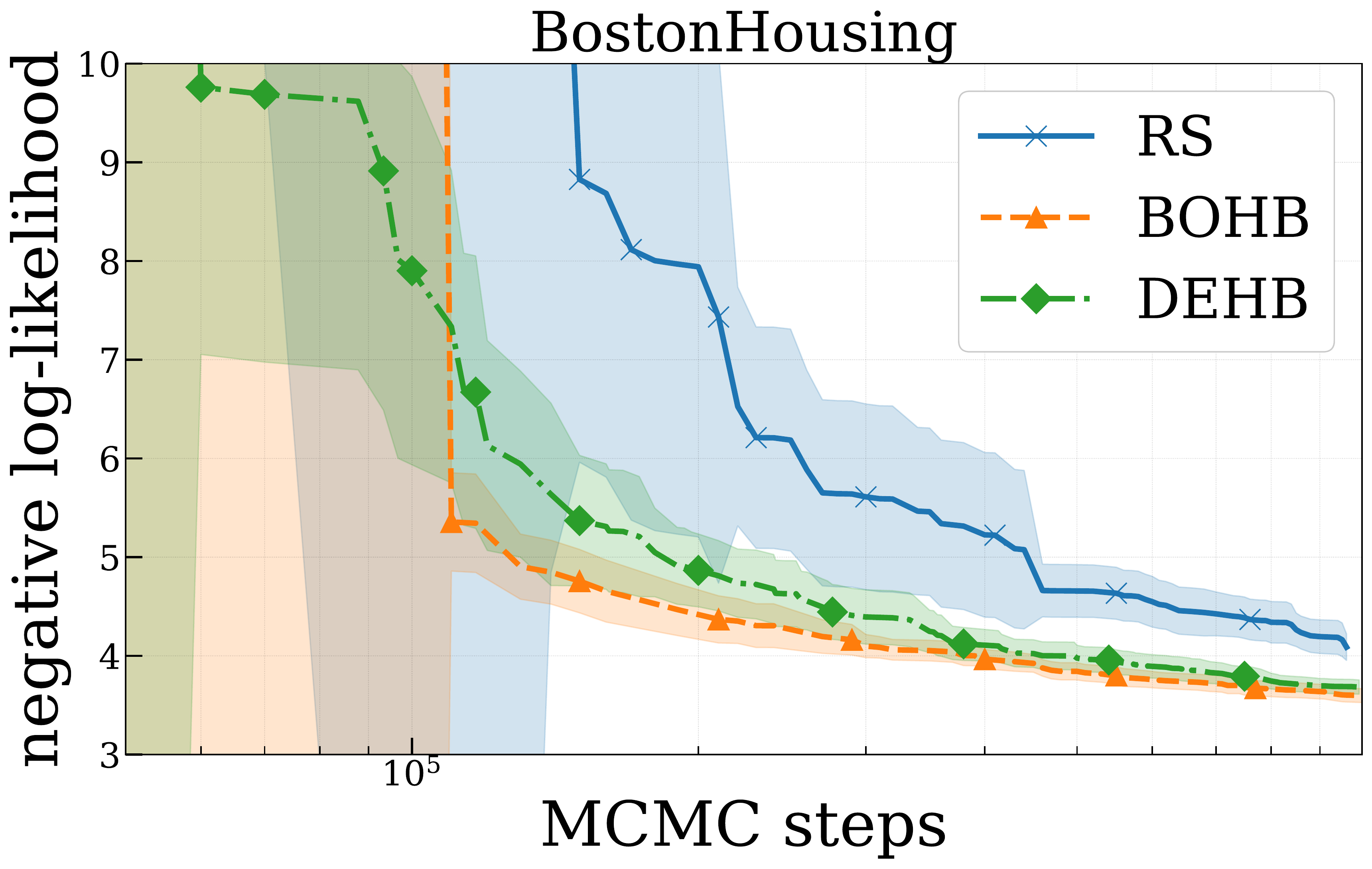}}
\end{figure}

In this benchmark, introduced by~\citet{falkner-icml18a}, a two-layer fully-connected Bayesian Neural Network is trained using stochastic gradient Hamiltonian Monte-Carlo sampling (SGHMC)~\citep{chen2014stochastic} with scale adaptation~\citep{springenberg2016bayesian}. The budgets were the number of MCMC steps ($500$ as minimum; $10000$ as maximum).
Two regression datasets from UCI~\citep{Dua:2019} were used for the experiments: \textit{Boston Housing} and \textit{Protein Structure}. 
Figure \ref{fig:sub-bnnboston} shows the results
(for \textit{Boston housing}; 
the results for \textit{Protein Structure} are in Appendix \ref{sec:app-bnn}). 
For this extremely noisy benchmark, BOHB and DEHB perform similarly, and both are about 2$\times$ faster than RS.




\subsection{Reinforcement Learning}\label{sec:exp-rl}
\begin{figure}
\centering
\floatbox[{\capbeside\thisfloatsetup{capbesideposition={right,top},capbesidewidth=4cm}}]{figure}[\FBwidth]
{\caption{Results for tuning PPO on OpenAI Gym cartpole environment with $7$ hyperparameters. Each algorithm was run for $50$ runs.}\label{fig:cartpole}}
{\includegraphics[width=4cm]{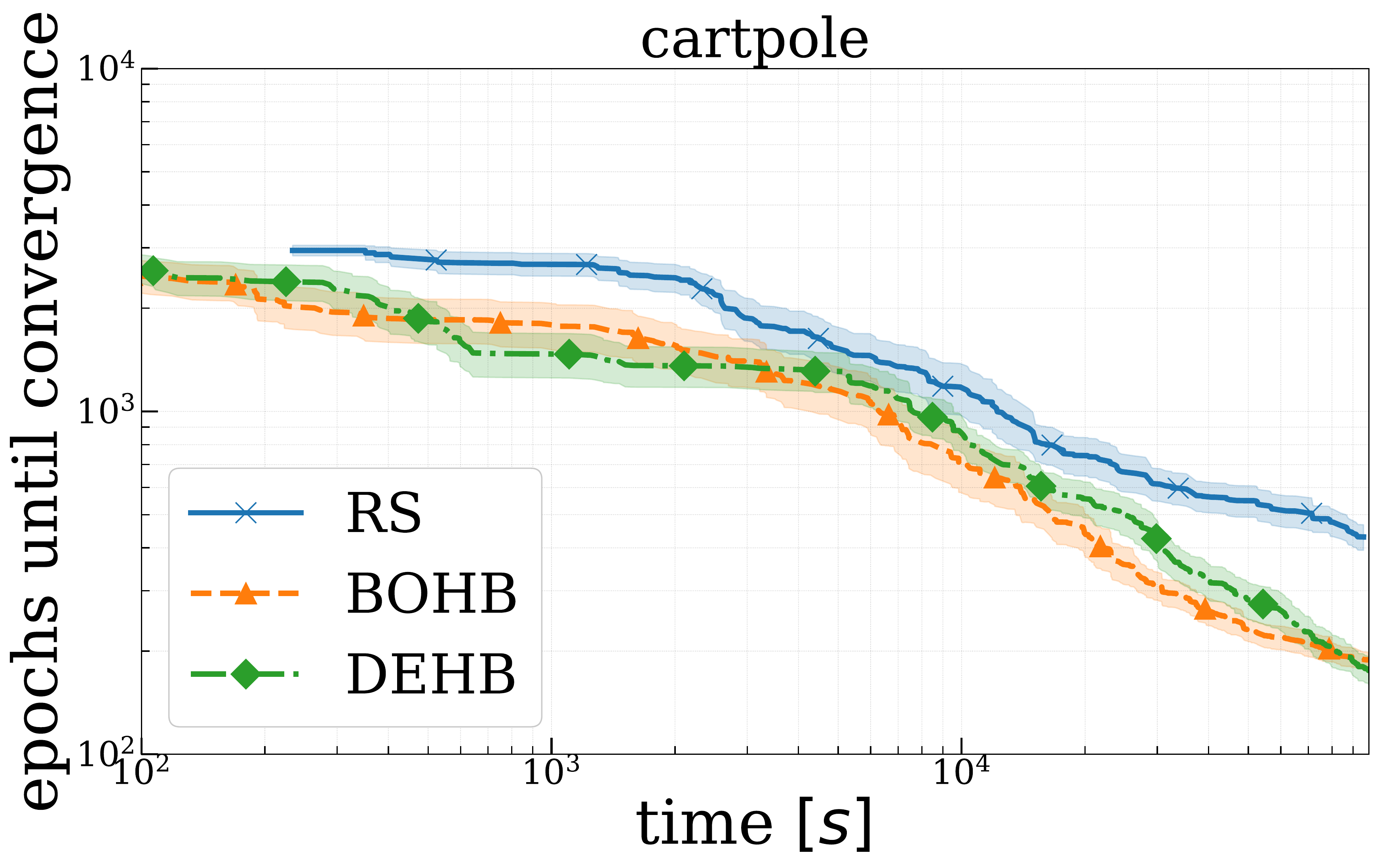}}
\end{figure}

For this benchmark used by~\citet{falkner-icml18a}), a proximal policy optimization (PPO) ~\citep{schulman2017proximal} implementation is parameterized with $7$ hyperparameters.
PPO is used to learn the \textit{cartpole swing-up} task from the OpenAI Gym ~\citep{brockman2016openai} environment. 
We plot the mean number of episodes needed until convergence for a configuration over actual cumulative wall-clock time in Figure \ref{fig:cartpole}. 
Despite the strong noise in this problem, BOHB and DEHB are able to improve continuously, showing similar performance, and speeding up over random search by roughly $2\times$.

\subsection{NAS Benchmarks}\label{sec:exp-nas}
\begin{figure}
\centering
\floatbox[{\capbeside\thisfloatsetup{capbesideposition={right,top},capbesidewidth=4cm}}]{figure}[\FBwidth]
{\caption{Results for Cifar C from NAS-Bench-101 for a $27$-dimensional space --- 22 continuous + 5 categorical hyperparameters)}\label{fig:sub-101-cifarc}}
{\includegraphics[width=4cm]{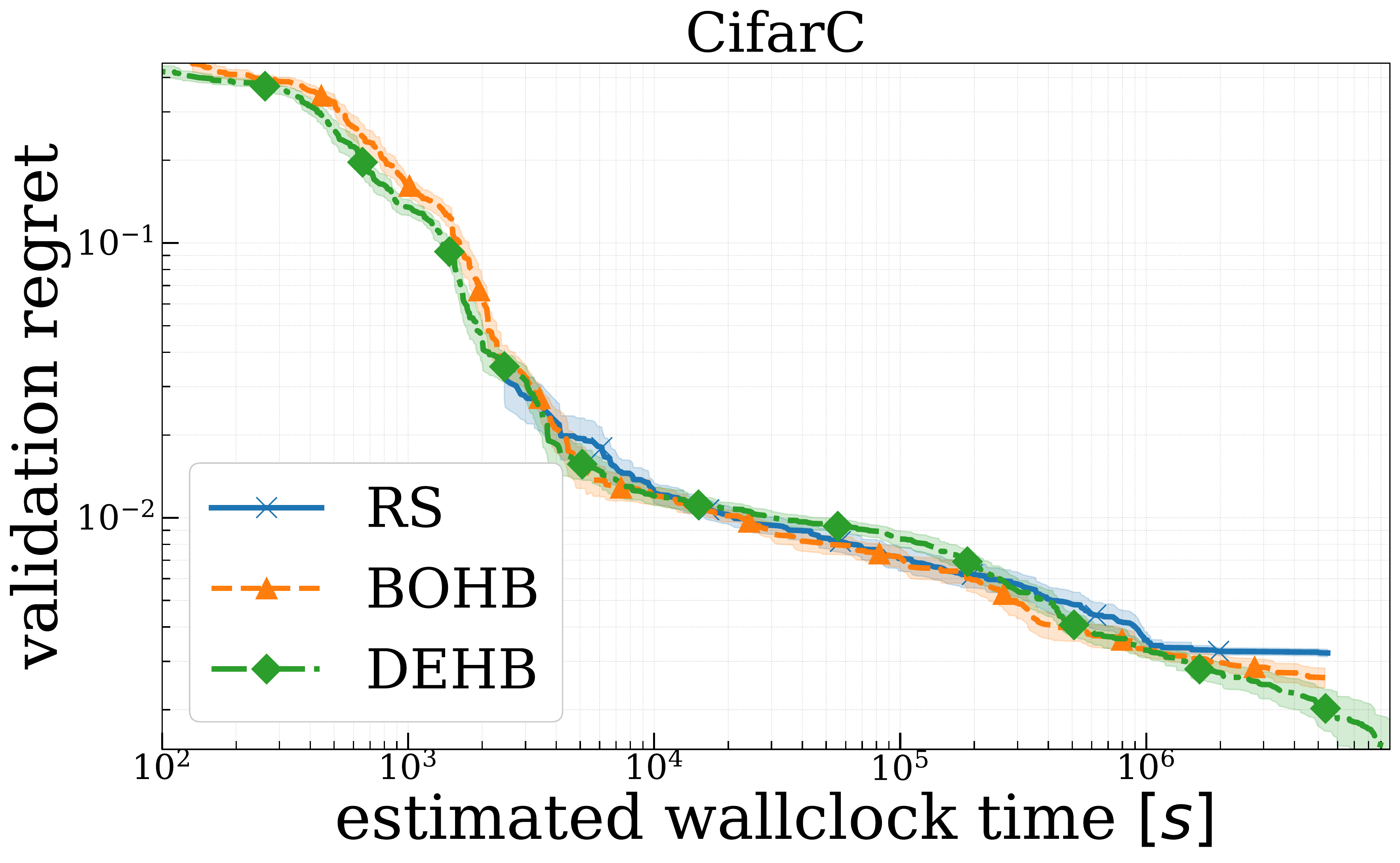}}
\end{figure}

\begin{figure}
\centering
\floatbox[{\capbeside\thisfloatsetup{capbesideposition={right,top},capbesidewidth=4cm}}]{figure}[\FBwidth]
{\caption{Results for ImageNet16-120 from NAS-Bench-201 for $50$ runs of each algorithm. The search space contains $6$ categorical parameters.}\label{fig:sub-201-imagenet}}
{\includegraphics[width=4cm]{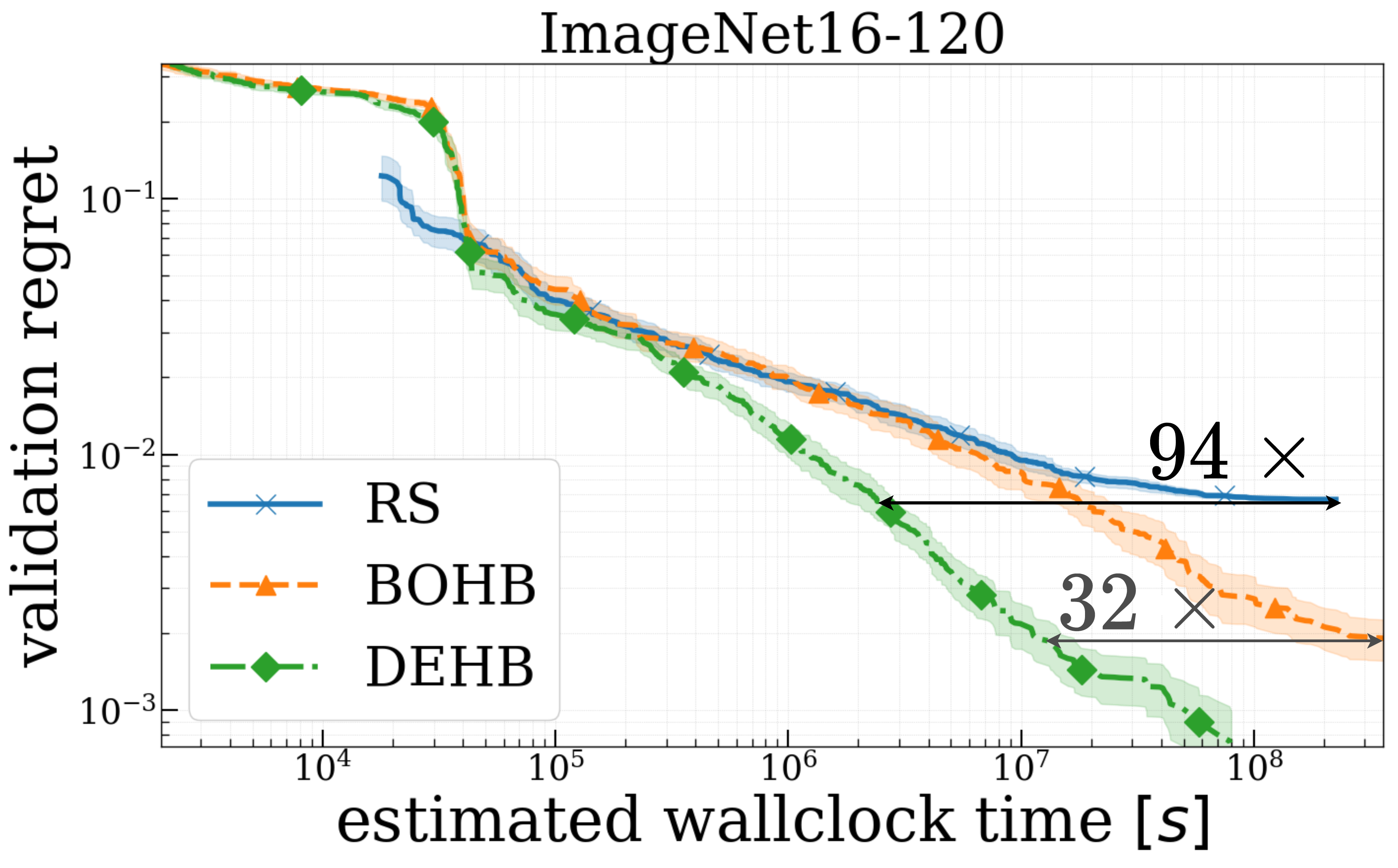}}
\end{figure}

In this series of experiments, we evaluate DEHB on a broad range of NAS benchmarks. We use a total of 13 tabular benchmarks from NAS-Bench-101 ~\citep{ying2019bench}, NAS-Bench-1shot1 ~\citep{zela2020bench}, NAS-Bench-201 ~\citep{dong2020bench} and NAS-HPO-Bench ~\citep{klein2019tabular}. 
For NAS-Bench-101, we show results on CifarC (a mixed data type encoding of the parameter space~\citep{awad2020iclr}) in Figure \ref{fig:sub-101-cifarc};
BOHB and DEHB initially perform similarly as RS for this dataset, since there is only little correlation between runs with few epochs (low budgets) and many epochs (high budgets) in NAS-Bench-101. In the end, RS stagnates, BOHB stagnates at a slightly better performance, and DEHB continues to improve.
In Figure \ref{fig:sub-201-imagenet}, we report results for ImageNet16-120 from NAS-201. In this case, DEHB is clearly the best of the methods, quickly converging to a strong solution. 

\begin{figure}
\centering
\floatbox[{\capbeside\thisfloatsetup{capbesideposition={right,top},capbesidewidth=4cm}}]{figure}[\FBwidth]
{\caption{Results for the Protein Structure dataset from NAS-HPO-Bench for $50$ runs of each algorithm. The search space contains $9$ hyperparameters.}\label{fig:main-nas-hpo-protein}}
{\includegraphics[width=4cm]{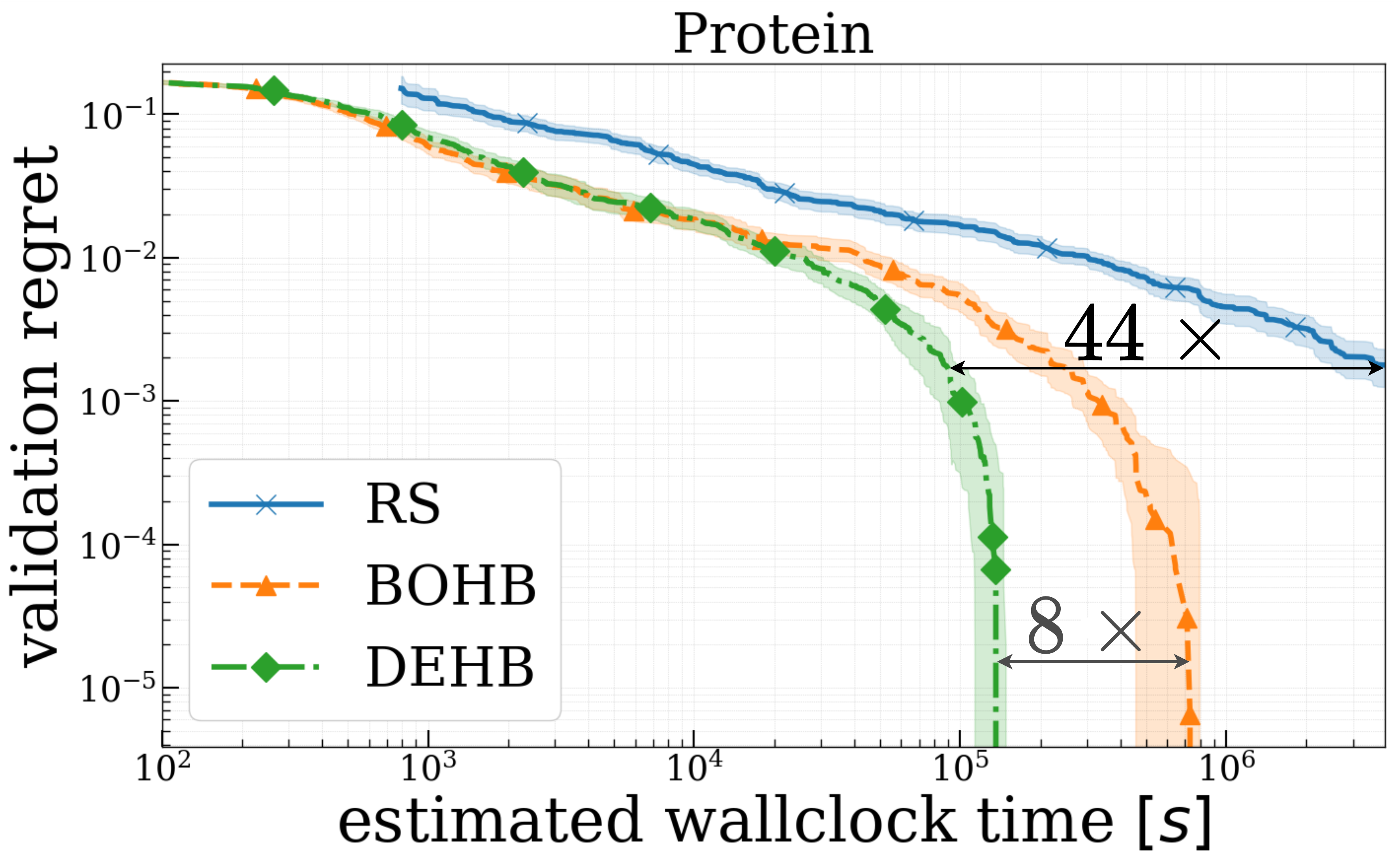}}
\end{figure}
Finally, Figure \ref{fig:main-nas-hpo-protein} reports results for the Protein Structure dataset provided in NAS-HPO-Bench.
DEHB makes progress faster than BOHB to reach the optimum.
The results on other NAS benchmarks are qualitatively similar to these 3 representative benchmarks, and are given in Appendix \ref{sec:app-nasbench}.

\subsection{Results summary}
\begin{figure}
\centering
  \includegraphics[width=0.85\columnwidth]{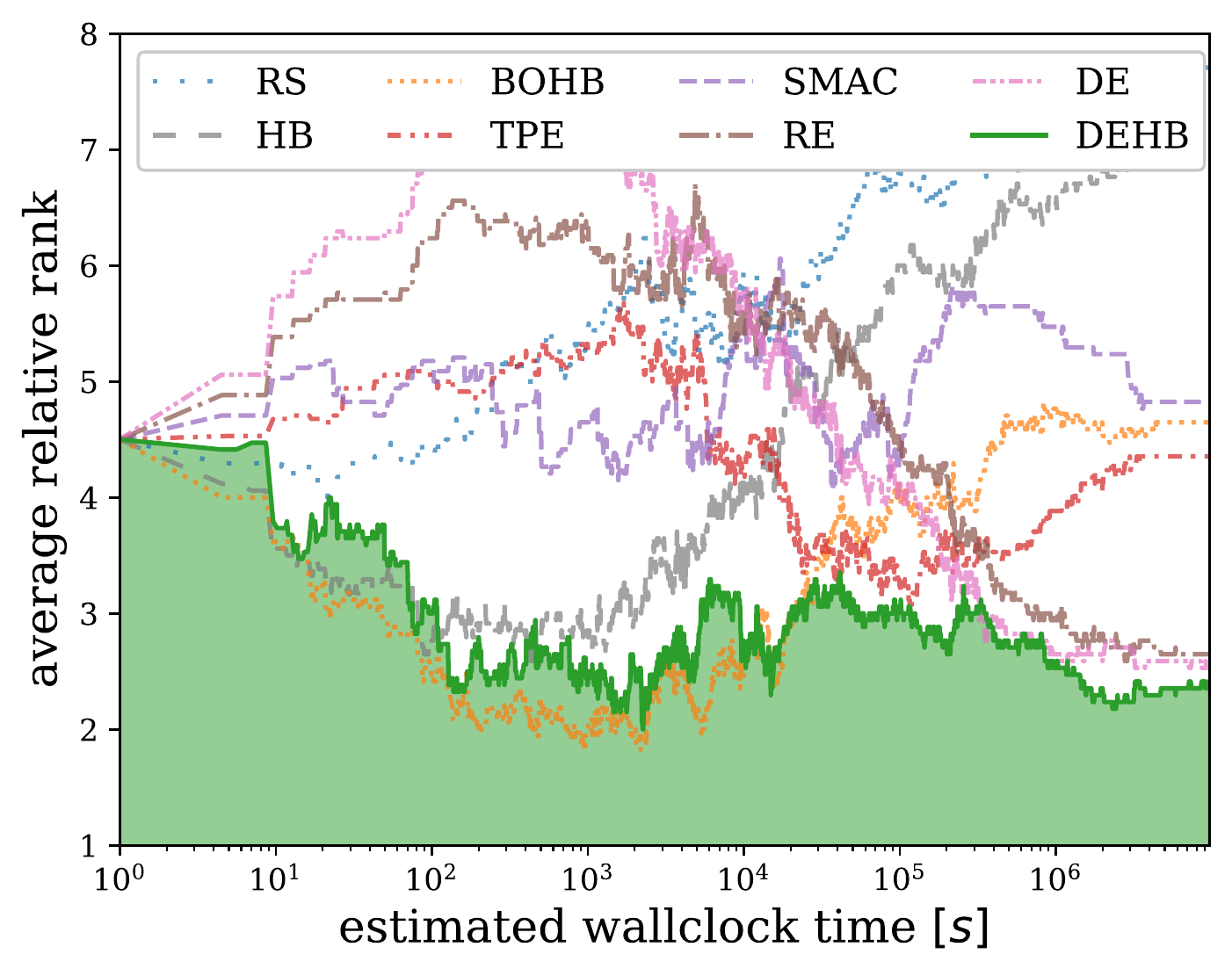}
  \caption{Average rank of the mean validation regret of $50$ runs of each algorithm, averaged over the NAS-Bench-101, NAS-Bench-1shot1, NAS-HPO-Bench, NAS-Bench-201, OpenML surrogates, and the Reinforcement Learning benchmarks.}
  \label{fig:ranking}
\end{figure}

We now compare DEHB to a broader range of baseline algorithms, also including HB, TPE~\citep{bergstra-nips11a}, SMAC~\citep{hutter-lion11a}, regularized evolution (RE)~\citep{real2019regularized}, and DE. 
Based on the mean validation regret, all algorithms can be ranked for each benchmark, for every second of the estimated wallclock time. Arranging the mean regret per timepoint across all benchmarks (except the Stochastic Counting Ones and the Bayesian Neural Network benchmarks, which do not have runtimes as budgets), we compute the \textit{average relative rank} over time for each algorithm in Figure \ref{fig:ranking}, where all $8$ algorithms were given the mean rank of $4.5$ at the beginning.
The shaded region clearly indicates that DEHB is the most robust algorithm for this set of benchmarks (discussed further in Appendix \ref{sec:app-res-summary}).
In the end, RE and DE are similarly good, but these blackbox optimization algorithms perform worst for small compute budgets, while DEHB's multi-fidelity aspect makes it robust across compute budgets.
In Table \ref{table:rank-summary}, we show the average rank of each algorithm based on the final validation regret achieved across all benchmarks (now also including Stochastic Counting Ones and Bayesian Neural Networks; data derived from Table \ref{table:summary} in Appendix \ref{sec:app-res-summary}). 
Next to its strong anytime performance, DEHB also yields the best final performance in this comparison, thus emerging as a strong general optimizer that works consistently across a diverse set of benchmarks. Result tables and figures for all benchmarks can be found in Appendix \ref{sec:app-exp}.

\begin{table}[h!]
\centering
\setlength{\tabcolsep}{2pt}
\scriptsize 
\begin{tabular}{||c|c c c c c c c c||} 
 \hline
 \  & RS & HB & BOHB & TPE & SMAC & RE & DE & DEHB \\ [0.5ex]  
 \hline
 \hline
 $Avg.\ rank$ & \textbf{7.46} & \textbf{6.54} & \textbf{4.42} & \textbf{4.35} & \textbf{4.73} & \textbf{3.16} & \textbf{2.96} & \textbf{2.39} \\ 
 \hline
 \hline
\end{tabular}
\vspace{0.1cm}
\caption{Mean ranks based on final mean validation regret for all algorithms tested for all benchmarks.}
\label{table:rank-summary}
\vspace*{-0.3cm}
\end{table}

\section{Conclusion}
\label{sec:conclusion}
We introduced DEHB, a new, general HPO solver, built to perform efficiently and robustly across many different problem domains. As discussed, DEHB satisfies the many requirements of such an HPO solver: strong performance with both short and long compute budgets, robust results, scalability to high dimensions, flexibility to handle mixed data types, parallelizability, and low computational overhead. Our experiments show that DEHB meets these requirements and in particular yields much more robust performance for discrete and high-dimensional problems than BOHB, the previous best overall HPO method we are aware of. Indeed, in our experiments, DEHB was up to $32\times$ faster than BOHB and up to $1000\times$ faster than random search.
DEHB does not require advanced software packages, is simple by design, and can easily be implemented across various platforms and languages, allowing for practical adoption. We thus hope that DEHB will become a new default HPO method. Our reference implementation of DEHB is available at \url{https://github.com/automl/DEHB}.


~\\\noindent\textbf{Acknowledgements.} The authors acknowledge funding by the Robert Bosch GmbH, by the German Federal Ministry of Education and Research (BMBF, grant RenormalizedFlows 01IS19077C), and support by the state of Baden-W\"{u}rttemberg through bwHPC and the German Research Foundation (DFG) through grant no INST 39/963-1 FUGG. 

\bibliographystyle{named}
\bibliography{bib/shortstrings,bib/lib,bib/local,bib/shortproc}

\begin{thebibliography}{}

\bibitem[\protect\citeauthoryear{Angeline \bgroup \em et al.\egroup
  }{1994}]{angeline1994evolutionary}
P.J. Angeline, G.M. Saunders, and J.B. Pollack.
\newblock An evolutionary algorithm that constructs recurrent neural networks.
\newblock {\em IEEE transactions on Neural Networks}, 5(1):54--65, 1994.

\bibitem[\protect\citeauthoryear{Awad \bgroup \em et al.\egroup
  }{2020}]{awad2020iclr}
N.~Awad, N.~Mallik, and F.~Hutter.
\newblock Differential evolution for neural architecture search.
\newblock In {\em First ICLR Workshop on Neural Architecture Search}, 2020.

\bibitem[\protect\citeauthoryear{Bergstra \bgroup \em et al.\egroup
  }{2011}]{bergstra-nips11a}
J.~Bergstra, R.~Bardenet, Y.~Bengio, and B.~K{\'e}gl.
\newblock Algorithms for hyper-parameter optimization.
\newblock In {\em Proc. of {N}eur{IPS}'11}, pages 2546--2554, 2011.

\bibitem[\protect\citeauthoryear{Brockman \bgroup \em et al.\egroup
  }{2016}]{brockman2016openai}
G.~Brockman, V.~Cheung, L.~Pettersson, J.~Schneider, J.~Schulman, J.~Tang, and
  W.~Zaremba.
\newblock Openai gym.
\newblock {\em arXiv preprint arXiv:1606.01540}, 2016.

\bibitem[\protect\citeauthoryear{Chakraborty}{2008}]{chakraborty2008advances}
U.~K. Chakraborty.
\newblock {\em Advances in differential evolution}, volume 143.
\newblock Springer, 2008.

\bibitem[\protect\citeauthoryear{Chen \bgroup \em et al.\egroup
  }{2014}]{chen2014stochastic}
T.~Chen, E.~Fox, and C.~Guestrin.
\newblock Stochastic gradient hamiltonian monte carlo.
\newblock In {\em International conference on machine learning}, pages
  1683--1691, 2014.

\bibitem[\protect\citeauthoryear{Das \bgroup \em et al.\egroup
  }{2016}]{das2016recent}
S.~Das, S.~S. Mullick, and P.~N. Suganthan.
\newblock Recent advances in differential evolution--an updated survey.
\newblock {\em Swarm and Evolutionary Computation}, 27:1--30, 2016.

\bibitem[\protect\citeauthoryear{Dong and Yang}{2020}]{dong2020bench}
X.~Dong and Y.~Yang.
\newblock Nas-bench-102: Extending the scope of reproducible neural
  architecture search.
\newblock {\em arXiv preprint arXiv:2001.00326}, 2020.

\bibitem[\protect\citeauthoryear{Dua and Graff}{2017}]{Dua:2019}
D.~Dua and C.~Graff.
\newblock Uci machine learning repository, 2017.

\bibitem[\protect\citeauthoryear{Falkner \bgroup \em et al.\egroup
  }{2018}]{falkner-icml18a}
S.~Falkner, A.~Klein, and F.~Hutter.
\newblock {BOHB}: Robust and efficient hyperparameter optimization at scale.
\newblock In {\em Proc. of {ICML}'18}, pages 1437--1446, 2018.

\bibitem[\protect\citeauthoryear{Grefenstette}{1986}]{grefenstette-ga}
J.~J. Grefenstette.
\newblock Optimization of control parameters for genetic algorithms.
\newblock {\em IEEE Transactions on Systems, Man, and Cybernetics},
  16:341--359, 1986.

\bibitem[\protect\citeauthoryear{Henderson \bgroup \em et al.\egroup
  }{2018}]{henderson-aaai18}
P.~Henderson, R.~Islam, P.~Bachman, J.~Pineau, D.~Precup, and D.~Meger.
\newblock Deep reinforcement learning that matters.
\newblock In {\em Proc. of {AAAI}'18}, 2018.

\bibitem[\protect\citeauthoryear{Hutter \bgroup \em et al.\egroup
  }{2011}]{hutter-lion11a}
F.~Hutter, H.~Hoos, and K.~Leyton-Brown.
\newblock Sequential model-based optimization for general algorithm
  configuration.
\newblock In {\em Proc. of {LION}'11}, pages 507--523, 2011.

\bibitem[\protect\citeauthoryear{Jamieson and
  Talwalkar}{2016}]{jamieson-aistats16a}
K.~Jamieson and A.~Talwalkar.
\newblock Non-stochastic best arm identification and hyperparameter
  optimization.
\newblock In {\em Proc. of {AISTATS}'16}, 2016.

\bibitem[\protect\citeauthoryear{K.~Price and
  Lampinen}{2006}]{price2006differential}
R.~M.~Storn K.~Price and J.~A. Lampinen.
\newblock {\em Differential evolution: a practical approach to global
  optimization}.
\newblock Springer Science \& Business Media, 2006.

\bibitem[\protect\citeauthoryear{Kandasamy \bgroup \em et al.\egroup
  }{2017}]{kandasamy2017multi}
K.~Kandasamy, G.~Dasarathy, J.~Schneider, and B.~P{\'o}czos.
\newblock Multi-fidelity bayesian optimisation with continuous approximations.
\newblock {\em arXiv:1703.06240 [stat.ML]}, 2017.

\bibitem[\protect\citeauthoryear{Kandasamy \bgroup \em et al.\egroup
  }{2020}]{dragonfly2020}
K.~Kandasamy, K.~R. Vysyaraju, W.~Neiswanger, B.~Paria, C.~R. Collins,
  J.~Schneider, B.~Poczos, and E.~P. Xing.
\newblock Tuning hyperparameters without grad students: Scalable and robust
  bayesian optimisation with dragonfly.
\newblock {\em Journal of Machine Learning Research}, 21(81):1--27, 2020.

\bibitem[\protect\citeauthoryear{Klein and Hutter}{2019}]{klein2019tabular}
A.~Klein and F.~Hutter.
\newblock Tabular benchmarks for joint architecture and hyperparameter
  optimization.
\newblock {\em arXiv preprint arXiv:1905.04970}, 2019.

\bibitem[\protect\citeauthoryear{Klein \bgroup \em et al.\egroup
  }{2016}]{klein2016fast}
A.~Klein, S.~Falkner, S.~Bartels, P.~Hennig, and F.~Hutter.
\newblock Fast bayesian optimization of machine learning hyperparameters on
  large datasets.
\newblock {\em arXiv:1605.07079 [cs.LG]}, 2016.

\bibitem[\protect\citeauthoryear{Li \bgroup \em et al.\egroup
  }{2017}]{li-iclr17a}
L.~Li, K.~Jamieson, G.~DeSalvo, A.~Rostamizadeh, and A.~Talwalkar.
\newblock Hyperband: Bandit-based configuration evaluation for hyperparameter
  optimization.
\newblock In {\em Proc. of {ICLR}'17}, 2017.

\bibitem[\protect\citeauthoryear{Liu \bgroup \em et al.\egroup
  }{2016}]{liu2016multi}
B.~Liu, S.~Koziel, and Q.~Zhang.
\newblock A multi-fidelity surrogate-model-assisted evolutionary algorithm for
  computationally expensive optimization problems.
\newblock {\em Journal of computational science}, 12:28--37, 2016.

\bibitem[\protect\citeauthoryear{Liu \bgroup \em et al.\egroup
  }{2017}]{liu2017hierarchical}
H.~Liu, K.~Simonyan, O.~Vinyals, C.~Fernando, and K.~Kavukcuoglu.
\newblock Hierarchical representations for efficient architecture search.
\newblock {\em arXiv preprint arXiv:1711.00436}, 2017.

\bibitem[\protect\citeauthoryear{Liu \bgroup \em et al.\egroup
  }{2018}]{liu2018darts}
H.~Liu, K.~Simonyan, and Y.~Yang.
\newblock Darts: Differentiable architecture search.
\newblock {\em arXiv preprint arXiv:1806.09055}, 2018.

\bibitem[\protect\citeauthoryear{Melis \bgroup \em et al.\egroup
  }{2018}]{melis-iclr18a}
G.~Melis, C.~Dyer, and P.~Blunsom.
\newblock On the state of the art of evaluation in neural language models.
\newblock In {\em Proc. of {ICLR}'18}, 2018.

\bibitem[\protect\citeauthoryear{Pham \bgroup \em et al.\egroup
  }{2018}]{pham2018efficient}
H.~Pham, M.~Y. Guan, B.~Zoph, Q.~V. Le, and J.~Dean.
\newblock Efficient neural architecture search via parameter sharing.
\newblock {\em arXiv preprint arXiv:1802.03268}, 2018.

\bibitem[\protect\citeauthoryear{Real \bgroup \em et al.\egroup
  }{2017}]{real2017large}
E.~Real, S.~Moore, A.~Selle, S.~Saxena, Y.~L. Suematsu, J.~Tan, Q.~V. Le, and
  A.~Kurakin.
\newblock Large-scale evolution of image classifiers.
\newblock In {\em Proc.~of~ICML}, pages 2902--2911. JMLR. org, 2017.

\bibitem[\protect\citeauthoryear{Real \bgroup \em et al.\egroup
  }{2019}]{real2019regularized}
E.~Real, A.~Aggarwal, Y.~Huang, and Q.~V. Le.
\newblock Regularized evolution for image classifier architecture search.
\newblock In {\em Proc.~of~AAAI}, volume~33, pages 4780--4789, 2019.

\bibitem[\protect\citeauthoryear{Schulman \bgroup \em et al.\egroup
  }{2017}]{schulman2017proximal}
J.~Schulman, F.~Wolski, P.~Dhariwal, A.~Radford, and O.~Klimov.
\newblock Proximal policy optimization algorithms.
\newblock {\em arXiv preprint arXiv:1707.06347}, 2017.

\bibitem[\protect\citeauthoryear{Snoek \bgroup \em et al.\egroup
  }{2012}]{snoek2012practical}
J.~Snoek, H.~Larochelle, and R.~P. Adams.
\newblock Practical {B}ayesian optimization of machine learning algorithms.
\newblock In {\em Proc. of {N}eur{IPS}'12}, pages 2951--2959, 2012.

\bibitem[\protect\citeauthoryear{Springenberg \bgroup \em et al.\egroup
  }{2016}]{springenberg2016bayesian}
J.~T. Springenberg, A.~Klein, S.~Falkner, and F.~Hutter.
\newblock Bayesian optimization with robust bayesian neural networks.
\newblock In {\em Proc.~of~NeurIPS}, pages 4134--4142, 2016.

\bibitem[\protect\citeauthoryear{Storn and Price}{1997}]{storn1997differential}
R.~Storn and K.~Price.
\newblock Differential evolution--a simple and efficient heuristic for global
  optimization over continuous spaces.
\newblock {\em Journal of global optimization}, 11(4):341--359, 1997.

\bibitem[\protect\citeauthoryear{Vallati \bgroup \em et al.\egroup
  }{2015}]{vallati2015effective}
M.~Vallati, F.~Hutter, Luk{\'a}s L.~Chrpa, and T.~L. McCluskey.
\newblock On the effective configuration of planning domain models.
\newblock In {\em Twenty-Fourth International Joint Conference on Artificial
  Intelligence}, 2015.

\bibitem[\protect\citeauthoryear{Vanschoren \bgroup \em et al.\egroup
  }{2014}]{vanschoren2014openml}
J.~Vanschoren, J.~N.~Van Rijn, B.~Bischl, and L.~Torgo.
\newblock Openml: networked science in machine learning.
\newblock {\em ACM SIGKDD Explorations Newsletter}, 15(2):49--60, 2014.

\bibitem[\protect\citeauthoryear{Wang \bgroup \em et al.\egroup
  }{2017}]{wang2017generic}
H.~Wang, Y.~Jin, and J.~Doherty.
\newblock A generic test suite for evolutionary multifidelity optimization.
\newblock {\em IEEE Transactions on Evolutionary Computation}, 22(6):836--850,
  2017.

\bibitem[\protect\citeauthoryear{Xie and Yuille}{2017}]{xie2017genetic}
L.~Xie and A.~Yuille.
\newblock Genetic cnn.
\newblock In {\em Proc.~of~ICCV}, pages 1379--1388, 2017.

\bibitem[\protect\citeauthoryear{Ying \bgroup \em et al.\egroup
  }{2019}]{ying2019bench}
C.~Ying, A.~Klein, E.~Real, E.~Christiansen, K.~Murphy, and F.~Hutter.
\newblock Nas-bench-101: Towards reproducible neural architecture search.
\newblock {\em arXiv preprint arXiv:1902.09635}, 2019.

\bibitem[\protect\citeauthoryear{Zela \bgroup \em et al.\egroup
  }{2020}]{zela2020bench}
A.~Zela, J.~Siems, and F.~Hutter.
\newblock Nas-bench-1shot1: Benchmarking and dissecting one-shot neural
  architecture search.
\newblock {\em arXiv preprint arXiv:2001.10422}, 2020.

\end{thebibliography}

\clearpage




\appendix

\section{More details on DE}
\label{More-detail-DE} 
Differential Evolution (DE) is a simple, well-performing evolutionary algorithm to solve a variety of optimization problems \citep{price2006differential} \citep{das2016recent}. This algorithm was originally introduced in 1995 by Storn and Price \citep{storn1997differential}, and later attracted the attention of many researchers to propose new improved state-of-the-art algorithms \citep{chakraborty2008advances}. DE is based on four steps: initialization, mutation, crossover and selection. \noindent{}Algorithm \ref{alg:DE} presents the DE pseudo-code.

\textbf{Initialization.} DE is a population-based meta-heuristic algorithm which consists of a population of $N$ individuals. Each individual is considered a solution and expressed as a vector of $D$-dimensional decision variables as follows: 

\begin{equation}
pop_{g}=(x^{1}_{i,g},x^{2}_{i,g},...,x^{D}_{i,g}), i=1,2,...,N,
\label{Eq.1}
\end{equation}~

\noindent{}where $g$ is the generation number, $D$ is the dimension of the problem being solved and $N$ is the population size. The algorithm starts initially with randomly distributed individuals within the search space. The function value for the problem being solved is then computed for each individual, $f(x)$.

\begin{figure}
\centering
\includegraphics[width=0.9\columnwidth]{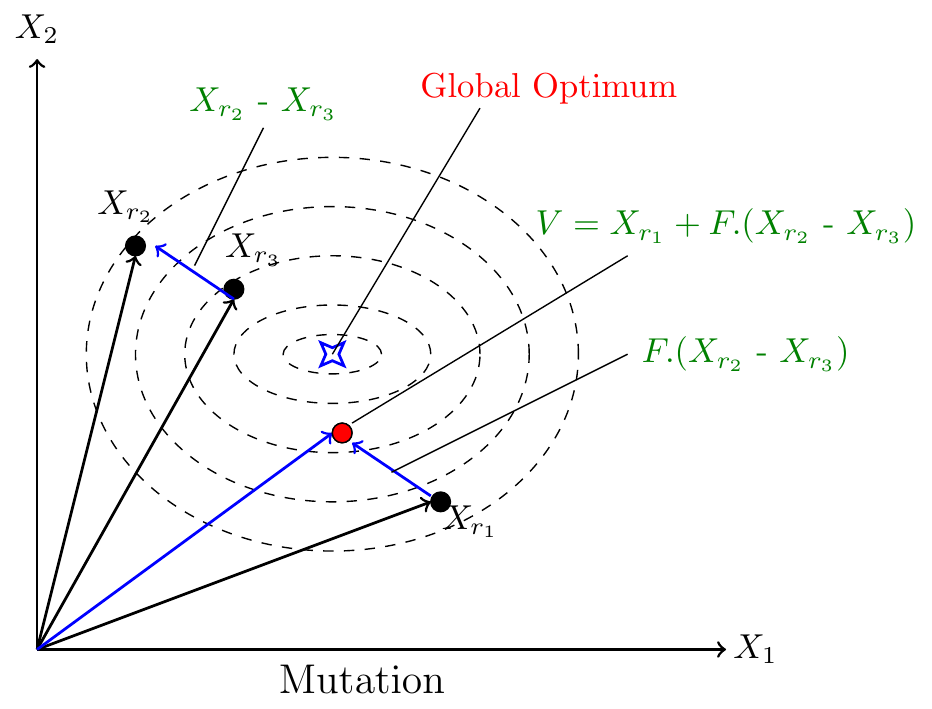}
\caption{Illustration of DE Mutation operation for a $2$-dimensional case using the \textit{rand/1} mutation strategy. The scaled difference vector ($F.(x_{r_2}-x_{r_3})$) is used to determine the neighbourhood of search from $x_{r_1}$. Depending on the diversity of the population, DE mutation's search will be explorative or exploitative}
\label{fig:de-mutation}
\end{figure}

\textbf{Mutation.} A new child/offspring is generated using the mutation operation for each individual in the population by a so called mutation strategy. Figure \ref{fig:de-mutation} illustrates this operation for a $2$-dimensional case. The classical DE uses the mutation operator $rand/1$, in which three random individuals/parents denoted as $x_{r_{1}}, x_{r_{2}}, x_{r_{3}}$ are chosen to generate a new vector $v_i$ as follows:

\begin{equation}
v_{i,g}= x_{r_1,g} + F\cdot(x_{r_2,g} - x_{r_3,g}),
\label{Eq.2}
\end{equation}%

\noindent{}where $v_{i,g}$ is the mutant vector generated for each individual $x_{i,g}$ in the population. $F$ is the scaling factor that usually takes values within the range (0, 1] and $r_1, r_2, r_3$ are the indices of different randomly-selected individuals. Eq.\ref{Eq.2} allows some parameters to be outside the search range, therefore, each parameter in $v_{i,g}$ is checked and reset\footnote{a random value from $[0,1]$ is chosen uniformly in this work} if it happens to be outside the boundaries. 

\textbf{Crossover.} When the mutation phase is completed, the crossover operation is applied to each target vector $x_{i,g}$ and its corresponding mutant vector $v_{i,g}$ to generate a trial vector $u_{i,g}$. Classical DE uses the following uniform (binomial) crossover:

\begin{equation}
    u^j_{i,g} =
    \begin{cases*}
      v^j_{i,g} & if $(rand \leq p)$ or $(j=j_{rand})$ \\
      x^j_{i,g} & otherwise
    \end{cases*}
    \label{Eq.3}
\end{equation}

\begin{figure}
\centering
\includegraphics[width=0.9\columnwidth]{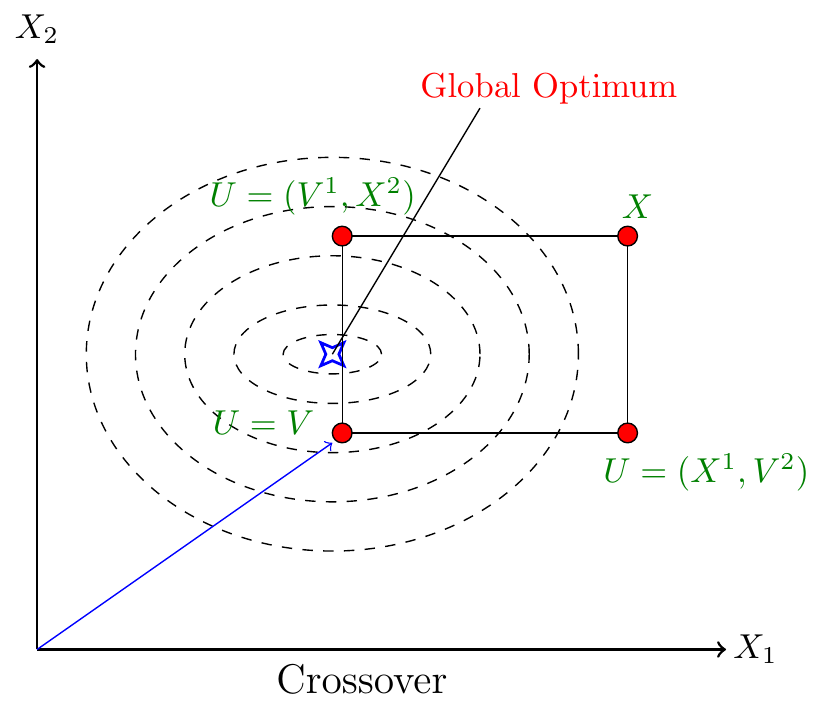}
\caption{Illustration of DE Crossover operation for a $2$-dimensional case using the \textit{binomial} crossover. The vertex of the rectangle shows the possible solutions of between a parent $x$ and mutant $v$. Based on the choice of $p$, the resultant individual will either be a copy of the parent, or the mutant, or incorporate either component from parent and mutant}
\label{fig:de-crx}
\end{figure}

The crossover rate $p$ is real-valued and is usually specified in the range [0, 1]. This variable controls the portion of parameter values that are copied from the mutant vector. The $j$th parameter value is copied from the mutant vector $v_{i,g}$ to the corresponding position in the trial vector $u_{i,g}$ if a random number is less than or equal to $p$. If the condition is not satisfied, then the $j$th position is copied from the target vector $x_{i,g}$. $j_{rand}$ is a random integer in the range [1, $D$] to ensure that at least one dimension is copied from the mutant, in case the random number generated for all dimensions is \textgreater $p$. Figure \ref{fig:de-crx} shows an illustration of the crossover operations.


\textbf{Selection.} After the final offspring is generated, the selection operation takes place to determine whether the target (the parent, $x_{i,g}$) or the trial (the offspring, $u_{i,g}$) vector survives to the next generation by comparing the function values. The offspring replaces its parents if it has a better\footnote{DE is a minimizer} function value as shown in Equation \ref{eq:selection}. Otherwise, the new offspring is discarded, and the target vector remains in the population for the next generation. 

\begin{equation}\label{eq:selection}
    x_{i,g} =
    \begin{cases*}
      u_{i,g} & if $(f(u_{i,g}) \leq f(x_{i,g}))$ \\
      x_{i,g} & otherwise
    \end{cases*}
\end{equation}%


\begin{algorithm}[!ht]
\KwIn{{}\\
  {$f$ - black-box problem}\\
  {$F$ - scaling factor (default $F=0.5$)}\\
  {$p$ - crossover rate (default $p=0.5$)}\\
  {$N$ - population size}\\
  }
  \KwOut{{}Return best found individual in $pop$}
	\BlankLine
	$g$ = 0, $FE$ = 0;\\
	$pop_g$ $\leftarrow$ initial\_population($N$, $D$); \\
	$fitness_g$ $\leftarrow$ evaluate\_population($pop_g$); \\
	$FE$ = $N$; \\
	\While{($FE$ $<$ $FE_{max}$)}{
		mutate($pop_g$);\\
		$offspring_g$ $\leftarrow$ crossover($pop_g$);\\
		$fitness_g$ $\leftarrow$ evaluate\_population($offspring_g$); \\
		$pop_{g+1}$,$fitness_{g+1}$ $\leftarrow$ select($pop_g$,$offspring_g$); \\
		$FE$ = $FE$ + $N$; \\
	    $g$ = $g$+1;\\
	}
	\Return Individual with highest fitness seen
	\caption{DE\_Optimizer}\label{alg:DE}
\end{algorithm}

\section{More details on Hyberband}
\label{More-detail-HB}
The Hyperband \citep{li-iclr17a} (HB) algorithm was designed to perform random sampling with early stopping based on pre-determined geometrically spaced resource allocation. For DEHB we replace the random sampling with DE search. However, DEHB uses HB at its core to solve the ``$n$ versus $B/n$" tradeoff that HB was designed to address. Algorithm \ref{alg:HB} shows how DEHB interfaces HB to query the sequence of how many configurations of each budget to run at each iteration. This view treats the DEHB algorithm as a sequence of predetermined (by HB), repeating Successive Halving brackets where, \textit{iteration} number refers to the index of SH brackets run by DEHB.

\begin{algorithm}[!ht]
\caption{A SH bracket under Hyperband}\label{alg:HB}
  \KwIn{{}\\
  {$b_{min}$, $b_{max}$ - min and max budgets}\\
  {$\eta$ - fraction of configurations promoted}\\
  {$iteration$ - iteration number}
  }
  \KwOut{List of no. of configurations and budgets}
  $s_{max} = \lfloor log_{\eta} \frac{b_{max}}{b_{min}} \rfloor$\\
  $s$ = $s_{max} - (iteration \bmod (s_{max}+1))$ \\
  $N \ {=}\ \lceil{ \frac{s_{max}+1}{s+1}\cdot\eta^s\rceil}$ \\
  $b_0 \ {=}\ \frac{b_{max}}{b_{min}}\cdot\eta^{-s}$ \\
  $budgets$ = $n\_configs$ = [] \\
  \For{$i\in \{0,...,s\}$}{
    $N_{i} \ {=}\ \lfloor N\cdot \eta^{-i} \rfloor$\\
    $b \ {=}\ b_0\cdot\eta^i$\\
    $n\_configs$.append($N_i$)\\
    $budgets$.append($b$)
  }
  \Return $n\_configs$, $budgets$
\end{algorithm}



\section{More details on DEHB}
\label{More-detail-DEHB}

\subsection{DEHB algorithm} \label{dehb-pseudo}
Algorithm \ref{alg:dehb} gives the pseudo code describing DEHB. DEHB takes as input the parameters for HB ($b_{min}$, $b_{max}$, $\eta$) and the parameters for DE ($F$, $p$). For the experiments in this paper, the \textit{termination\_condition} was chosen as the total number of DEHB brackets to run. However, in our implementation it can also be specified as the total absolute number of function evaluations, or a cumulative wallclock time as budget. L6 is the call to Algorithm \ref{alg:HB} which gives a list of \textit{budgets} which represent the sequence of increasing budgets to be used for that SH bracket. The nomenclature DE[\textit{budgets[i]}], used in L9 and L12, indicates the DE subpopulation associated with the \textit{budgets[i]} fidelity level. The \textit{if...else} block from L11-15 differentiates the first DEHB bracket from the rest. During the first DEHB bracket (\textit{bracket\_counter$==0$}) and its second SH bracket onwards ($i \textgreater 0$), the top configurations from the lower fidelity are \textit{promoted}\footnote{only \textit{evaluate} on higher budget and not evolve using mutation-crossover-selection} for evaluation in the next higher fidelity.
The function \textit{DE\_trial\_generation} on L14, i.e, the sequence of mutation-crossover operations, generates a candidate configuration (\textit{config}) to be evaluated for all other scenarios. 
L17 carries out the DE selection procedure by comparing the fitness score of \textit{config} and the selected \textit{target} for that DE evolution step. The \textit{target} ($x_{i,g}$ from Equation \ref{Eq.3}) is selected on L9 by a rolling pointer over the subpopulation list. That is, for every iteration (every increment of $j$) a pointer moves forward by one index position in the subpopulation selecting an individual to be a \textit{target}. When this pointer reaches the maximal index, it resets to point back to the starting index of the subpopulation. L18 compares the score of the last evaluated \textit{config} with the best found score so far. If the new \textit{config} has a better fitness score, the best found score is updated and the new \textit{config} is marked as the incumbent, $config_{inc}$. This stores the best found configuration as an \textit{anytime} best performing configuration.

\begin{algorithm}[!ht]
\caption{DEHB}\label{alg:dehb}
  \KwIn{
    {}\\
    {$b_{min}$, $b_{max}$ - min and max budgets}\\
    {$\eta$ - (default $\eta$=3})\\
    {$F$ - scaling factor (default $F=0.5$)}\\
    {$p$ - crossover rate (default $p=0.5$)}\\
  }
  \KwOut{Best found configuration, $config_{inc}$}
  $s_{max} = \lfloor log_{\eta} \frac{b_{max}}{b_{min}} \rfloor$\\
  $Initialize$ $(s_{max}+1)$ DE subpopulations randomly \\
  $bracket\_counter$ = 0 \\
  \While {termination\_condition}{
    \For { $iteration \in \{ 0, 1, ..., s_{max} \}$ } {
        $budgets$, $n\_configs$ = SH\_bracket\_under\_HB($b_{min}$, $b_{max}$, $\eta$, $iteration$) \\
        \For { $i \in \{0, 1, ..., s_{max} - iterations\}$ } {
            \For { $j \in \{1, 2, ..., n\_configs[i]\}$ } {
                target = rolling pointer for DE[$budgets[i]$] \\
                $mutation\_types$ = ``vanilla" \textbf{if} {$i$ is $0$} \textbf{else} ``altered" \\
                \eIf { $bracket\_counter$ is $0$ \text{and} i \textgreater 0 }{
                    $config$ = $j$-th best config from DE[$budgets[i-1]$] \\
                }{
                    $config$ = DE\_trial\_generation($target$, $mutation\_type$) \\
                }
                $result$ = Evaluate $config$ on $budgets[i]$ \\
                DE selection using $result$, $config$ vs. $target$ \\
                Update incumbent, $config_{inc}$\\
            }
        }
    } 
    $bracket\_counter$ += 1
  }
  \Return $config_{inc}$
\end{algorithm}

\subsection{Handling Mixed Data Types} \label{mixed-types}
When dealing with discrete or categorical search spaces, such as the NAS problem, the best way to apply DE with such parameters is to keep the population continuous and perform mutation and crossover normally (Eq. \ref{Eq.2}, \ref{Eq.3}); then, to evaluate a configuration we evaluate a copy of it in the original discrete space as we explain below. If we instead dealt with a discrete population, then the diversity of population would drop dramatically, leading to many individuals having the same parameter values; the resulting population would then have many duplicates, lowering the diversity of the difference distribution and making it hard for DE to explore effectively. We designed DEHB to scale all parameters of a configuration in a population to a unit hypercube $[0, 1]$, for the two broad types of parameters normally encountered:

\begin{itemize}
    \item \textit{Integer} and \textit{float} parameters, $X^i \in [a_i, b_i]$ are retrieved as: $a_i + (b_i - a_i) \cdot U_{i,g}$, where the integer parameters are additionally rounded.
    \item \textit{Ordinal} and \textit{categorical} parameters, $X^i \in \{x_1, ..., x_n\}$, are treated equivalently s.t. the range [0, 1] is divided uniformly into $n$ bins.
\end{itemize}

\begin{figure}
\centering
\floatbox[{\capbeside\thisfloatsetup{capbesideposition={right,center},capbesidewidth=4cm}}]{figure}[\FBwidth]
{\caption{Comparing DEHB encodings for the Stochastic Counting Ones problem in $64$ dimensional space with $32$ categorical and $32$ continuous hyperparameters. Results for all algorithms on 50 runs.}\label{fig:sub-32+32-encoding}}
{\includegraphics[width=4cm]{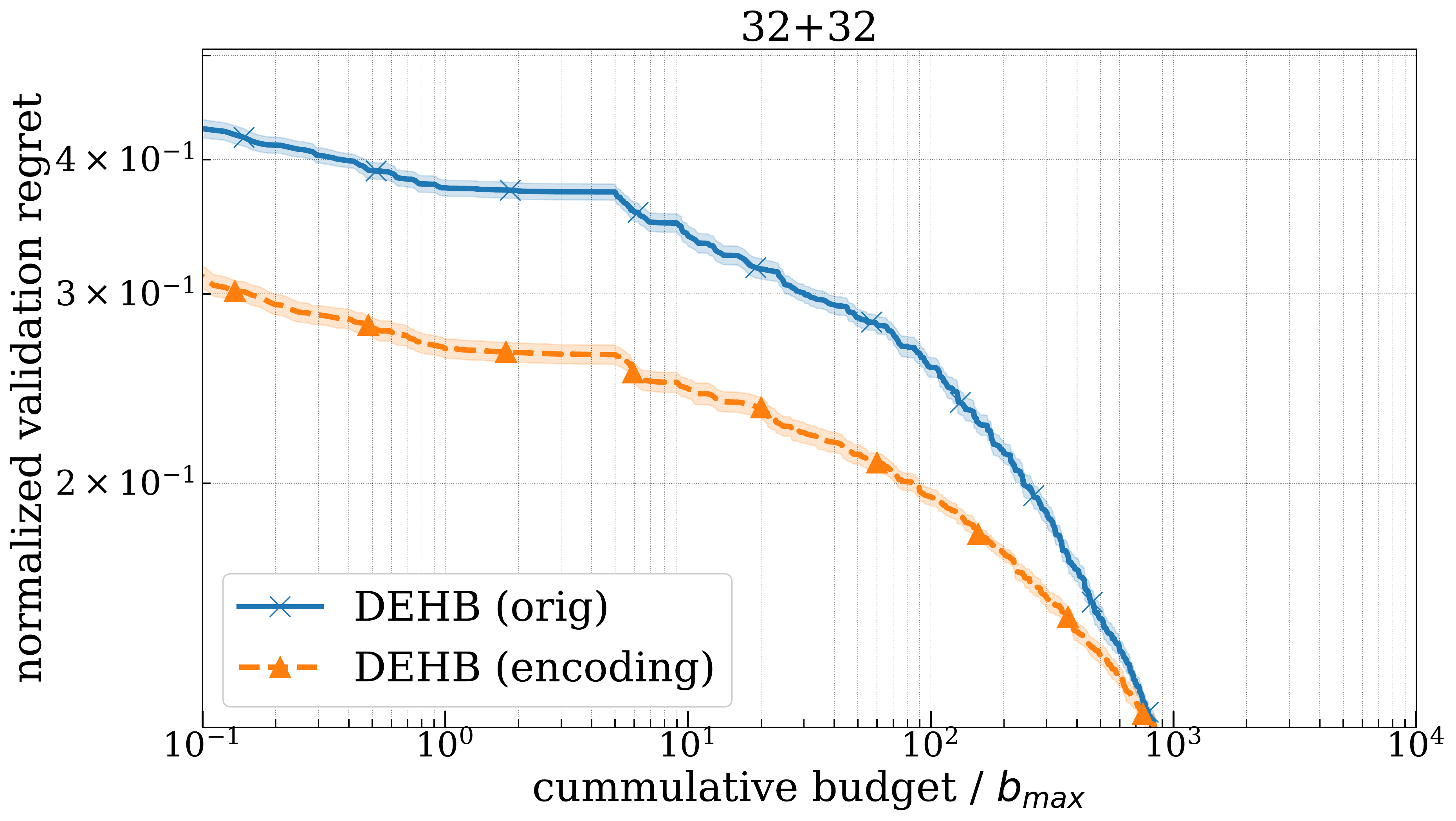}}
\end{figure}

\begin{figure}
\centering
\floatbox[{\capbeside\thisfloatsetup{capbesideposition={right,center},capbesidewidth=4cm}}]{figure}[\FBwidth]
{\includegraphics[width=4cm]{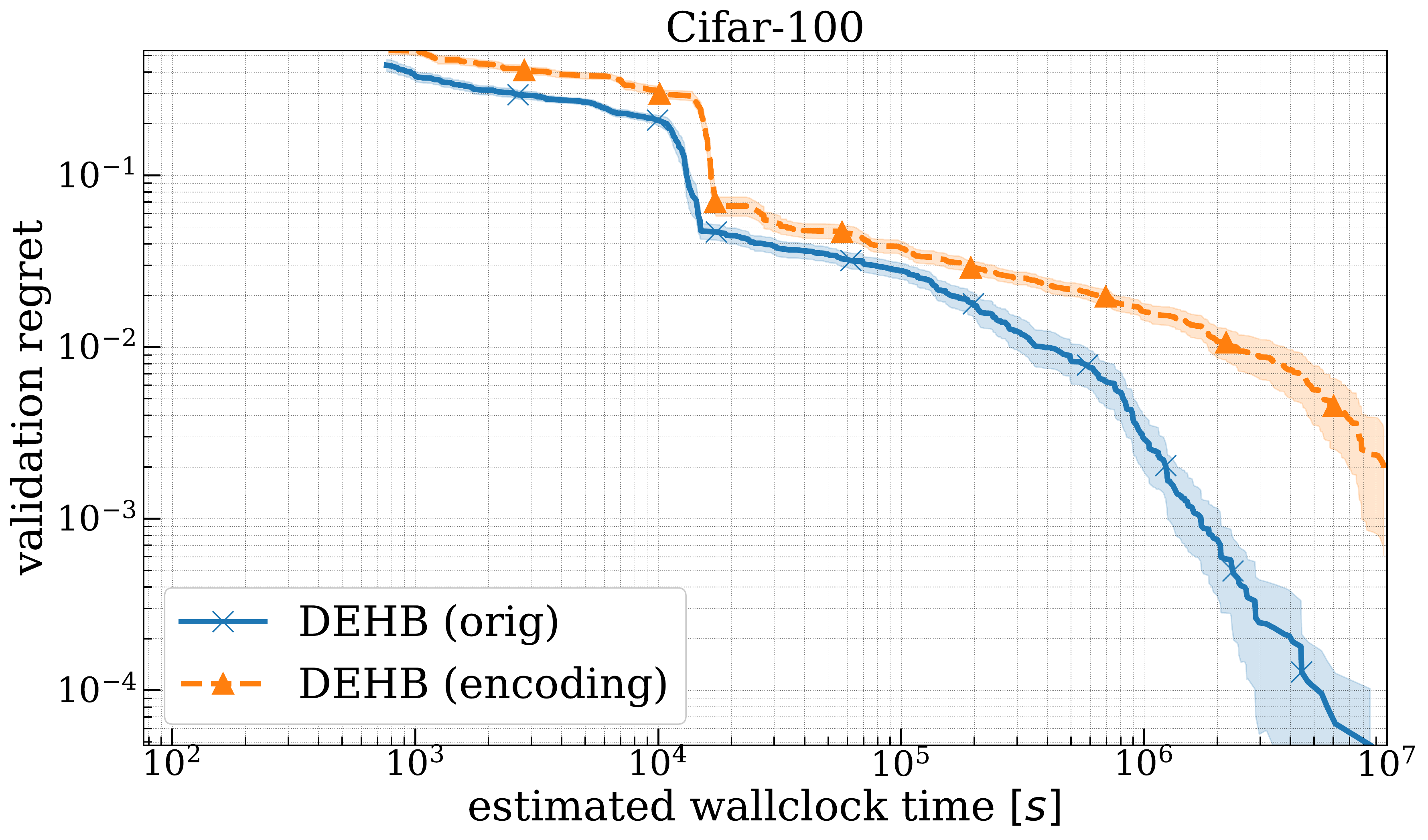}}
{\caption{Comparing DEHB encodings for the Cifar-100 dataset from NAS-Bench-201's $6$-dimensional space. Results for all algorithms on 50 runs.}\label{fig:sub-cifar100-encoding}}
\end{figure}

We also experimented with another encoding design where each category in each of the categorical variables are represented as a continuous variables $[0, 1]$ and the variable with the \textit{max} over the continuous variables is chosen as the category \citep{vallati2015effective}. For example, in Figure \ref{fig:sub-32+32-encoding}, the effective dimensionality of the search space will become $96$-dimensional --- $32$ continuous variables + $64$ continuous variables derived from $32$ binary variables. We choose a representative set of benchmarks (NAS-Bench-201 and Counting Ones) to compare DEHB with the two encodings mentioned, in Figures \ref{fig:sub-32+32-encoding}, \ref{fig:sub-cifar100-encoding}. It is enough to see one example which performs much worse than the DE-NAS \citep{awad2020iclr} encoding we chose for DEHB. The encoding from \citep{vallati2015effective} did not achieve a better final performance than DEHB in any of our experiments.


\subsection{Parallel Implementation} \label{app:parallel}
The DEHB algorithm is a sequence of DEHB Brackets, which in turn are a fixed sequence of SH brackets. This feature, along with the asynchronous nature of DE allows a parallel execution of DEHB. We dub the main process as the \textit{DEHB Orchestrator} which maintains a single copy of all DE subpopulations. An \textit{HB bracket manager} determines which budget to run from which SH bracket. Based on this input from the bracket manager, the orchestrator can fetch a configuration\footnote{DE mutation and crossover to generate configuration} from the current subpopulations and make an asynchronous call for its evaluation on the assigned budget. The rest of the orchestrator continues synchronously to check for free workers, and query the HB bracket manager for the next budget and SH bracket. Once a worker finishes computation, the orchestrator collects the result, performs DE selection and updates the relevant subpopulation accordingly. This form of an update is referred to as \textit{immediate, asynchronous} DE. 

DEHB uses a \textit{synchronous} SH routine. Though each of the function evaluations at a particular budget can be distributed, a higher budget needs to wait on all the lower budget evaluations to be finished. A higher budget evaluation can begin only once the lower budget evaluations are over and the top 1/$\eta$ can be selected. However, the \textit{asynchronous} nature of DE allows a new bracket to begin if a worker is available while existing SH brackets have pending jobs or are waiting for results. The new bracket can continue using the current state of DE subpopulations maintained by the \textit{DEHB Orchestrator}. Once the pending jobs from previous brackets are over, the DE selection updates the \textit{DEHB Orchestrator's} subpopulations. Thus, the utilisation of available computational resources is maximized while the central copy of subpopulations maintained by the \textit{Orchestrator} ensures that each new SH bracket spawned works with the latest updated subpopulation.


\section{More details on Experiments} \label{sec:app-exp}
\subsection{Baseline Algorithms}
\label{baseline-algs}
In all our experiments we keep the configuration of all the algorithms the same. These settings are well-performing setting that have been benchmarked in previous works --- \citep{falkner-icml18a}, \citep{ying2019bench}, \citep{awad2020iclr}.

\textbf{Random Search (RS)} We sample random architectures in the configuration space from a uniform distribution in each generation.   

\textbf{BOHB} We used the implementation from \url{https://github.com/automl/HpBandSter}. In \citep{ying2019bench}, they identified the settings of key hyperparameters as: $\eta$ is set to $3$, the minimum bandwidth for the kernel density estimator is set to 0.3 and bandwidth factor is set to 3. In our experiments, we deploy the same settings.   

\textbf{Hyperband (HB)} We used the implementation from \url{https://github.com/automl/HpBandSter}. We set $\eta=3$ and this parameter is not free to change since there is no other different budgets included in the NAS benchmarks.

\textbf{Tree-structured Parzen estimator (TPE)} We used the open-source implementation from \url{https://github.com/hyperopt/hyperopt}. We kept the settings of hyperparameters to their default.  

\textbf{Sequential Model-based Algorithm Configuration (SMAC)} We used the implementation from \url{https://github.com/automl/SMAC3} under its default parameter setting. Only for the Counting Ones problem with $64$-dimensions, the \textit{initial design} had to be changed to a Latin Hypercube design, instead of a Sobol design.

\textbf{Regularized Evolution (RE)} We used the implementation from \citep{real2019regularized}. We initially sample an edge or operator uniformly at random, then we perform the mutation. After reaching the population size, RE kills the oldest member at each iteration. As recommended by \citep{ying2019bench}, the population size (PS) and sample size (TS) are set to 100 and 10 respectively. 

\textbf{Differential Evolution (DE)} We used the implementation from \citep{awad2020iclr}, keeping the \textit{rand1} strategy for mutation and \textit{binomial crossover} as the crossover strategy. We also use the same population size of $20$ as \citep{awad2020iclr}.

All plots for all baselines were plotted for the incumbent validation regret over the estimated wallclock time, ignoring the optimization time. The $x$-axis therefore accounts for only the cumulative cost incurred by function evaluations for each algorithm. All algorithms were run for similar \textit{actual} wallclock time budget. Certain algorithms under certain benchmarks may not appear to have equivalent total \textit{estimated} wallclock time. That is an artefact of ignoring optimization time. Model-based algorithms such as SMAC, BOHB, TPE have a computational cost dependent on the observation history. They might undertake lesser number of function evaluations for the same actual wallclock time.

\subsection{Artificial Toy Function: Stochastic Counting Ones}\label{sec:app-counting}

\begin{figure}[ht]
\centering
\begin{tabular}{ll}
    \includegraphics[width=0.45\columnwidth]{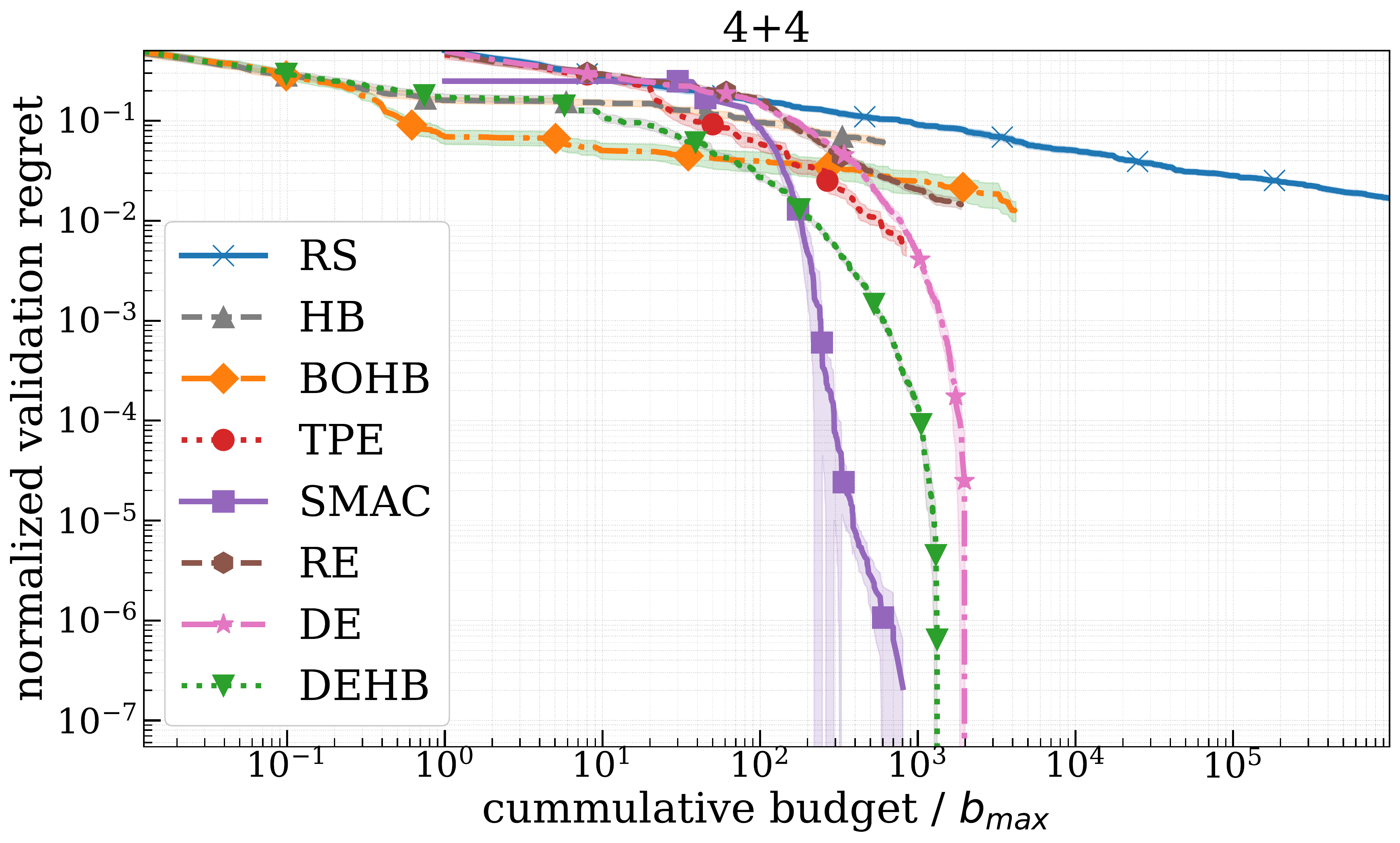} &
    \includegraphics[width=0.45\columnwidth]{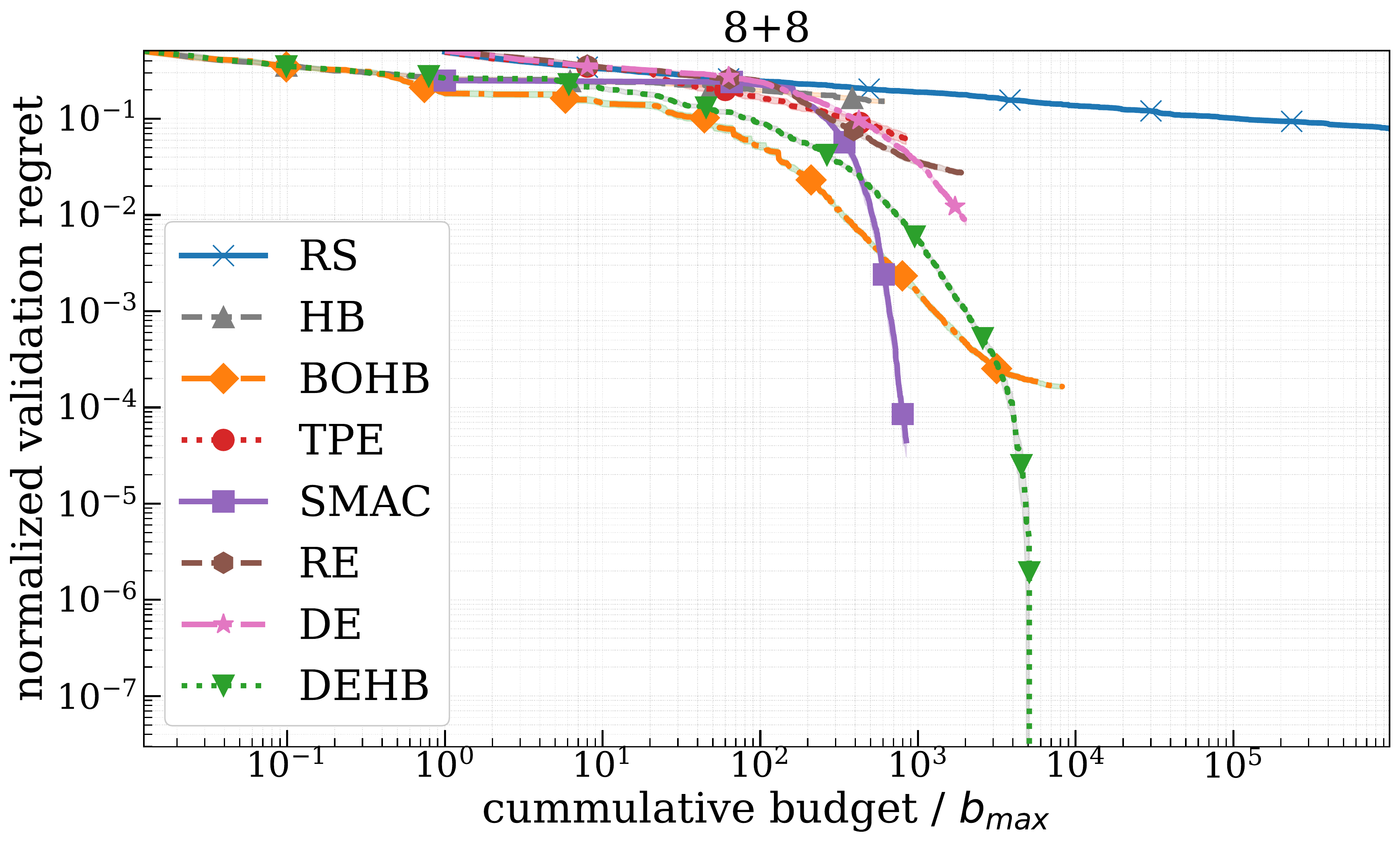} \\
    \includegraphics[width=0.45\columnwidth]{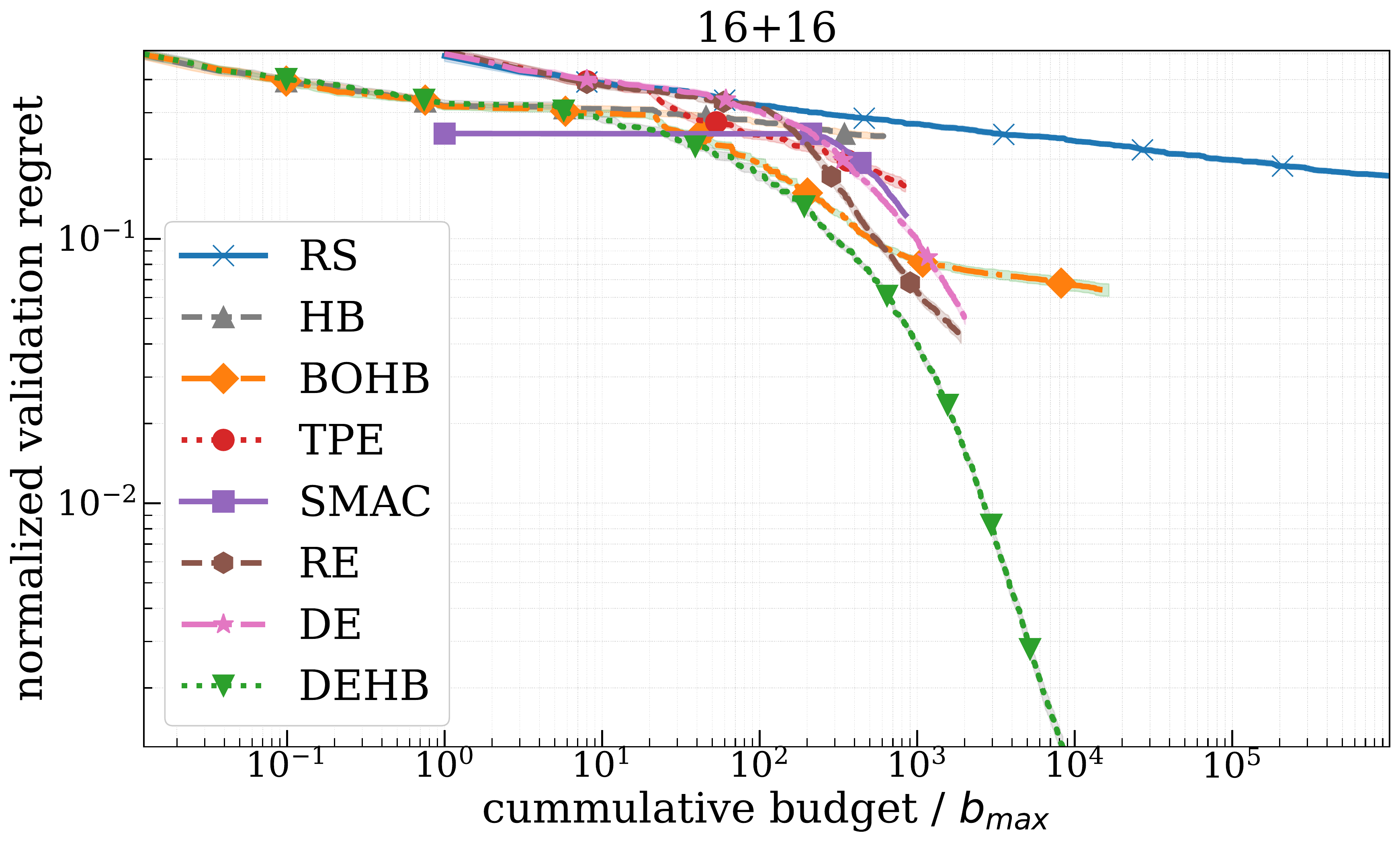} &
    \includegraphics[width=0.45\columnwidth]{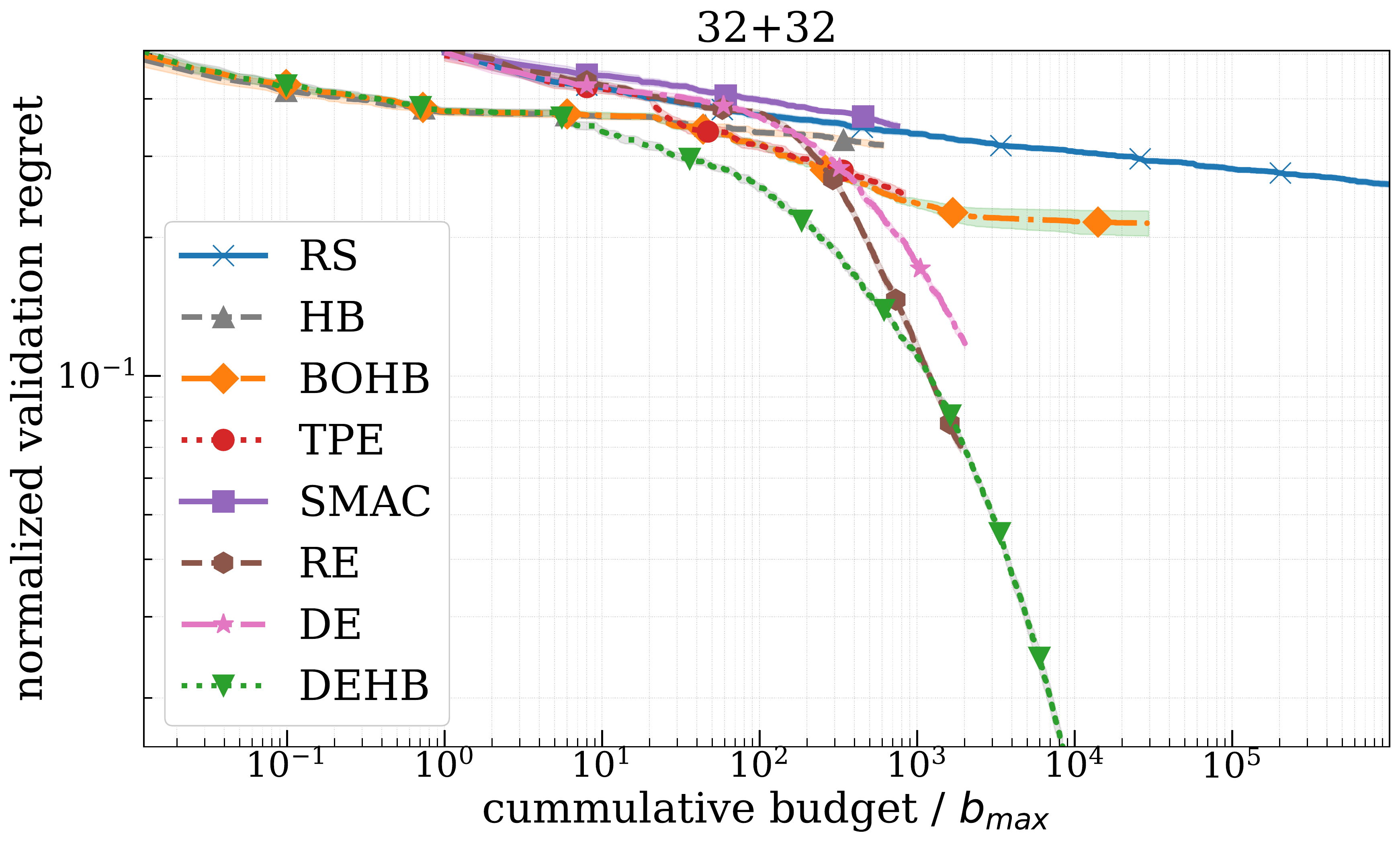}
\end{tabular}
\caption{Results for the Stochastic Counting Ones problem for $N=\{4, 8, 16, 32\}$ respectively indicating $N$ categorical and $N$ continuous hyperparameters for each case. All algorithms shown were run for 50 runs.}
\label{fig:sub-countingones}
\end{figure}



The Counting Ones benchmark was designed to minimize the following objective function:
\begin{equation*}
    f(x) = - \left(\sum_{x_i \in X_{cat}} x_i + \sum_{x_j \in X_{cont}} \E_{b}[(B_{p=x_j})] \right),
\end{equation*}
where the sum of the categorical variables ($x_i \in \{0,1\}$) represents the standard discrete counting ones problem. The continuous variables ($x_j \in [0,1]$) represent the stochastic component with the budget \textit{b} controlling the noise. The budget here represents the number of samples used to estimate the mean of the Bernoulli distribution ($B$) with parameters $x_j$. 



The experiments on the Stochastic Counting Ones benchmark used $N=\{4, 8, 16, 32\}$, all of which are shown in Figure \ref{fig:sub-countingones}. For the low dimensional cases, BOHB and SMAC's models are able to give them an early advantage. For this toy benchmark the global optima is located at the corner of a unit hypercube. Random samples can span the lower dimensional space adequately for a model to improve the search rapidly. DEHB on the other hand may require a few extra function evaluations to reach similar convergence. However, this conservative approach aids DEHB for the high-dimensional cases where it is able to converge much more rapidly in comparison to other algorithms. Especially where SMAC and BOHB's convergence worsens significantly. DEHB thus showcases its robust performance even when the dimensionality of the problem increases exponentially.

\subsection{Feed-forward networks on OpenML datasets}\label{sec:app-openml}

\begin{figure}[htbp]
\centering
\begin{tabular}{ll}
    \includegraphics[width=0.45\columnwidth]{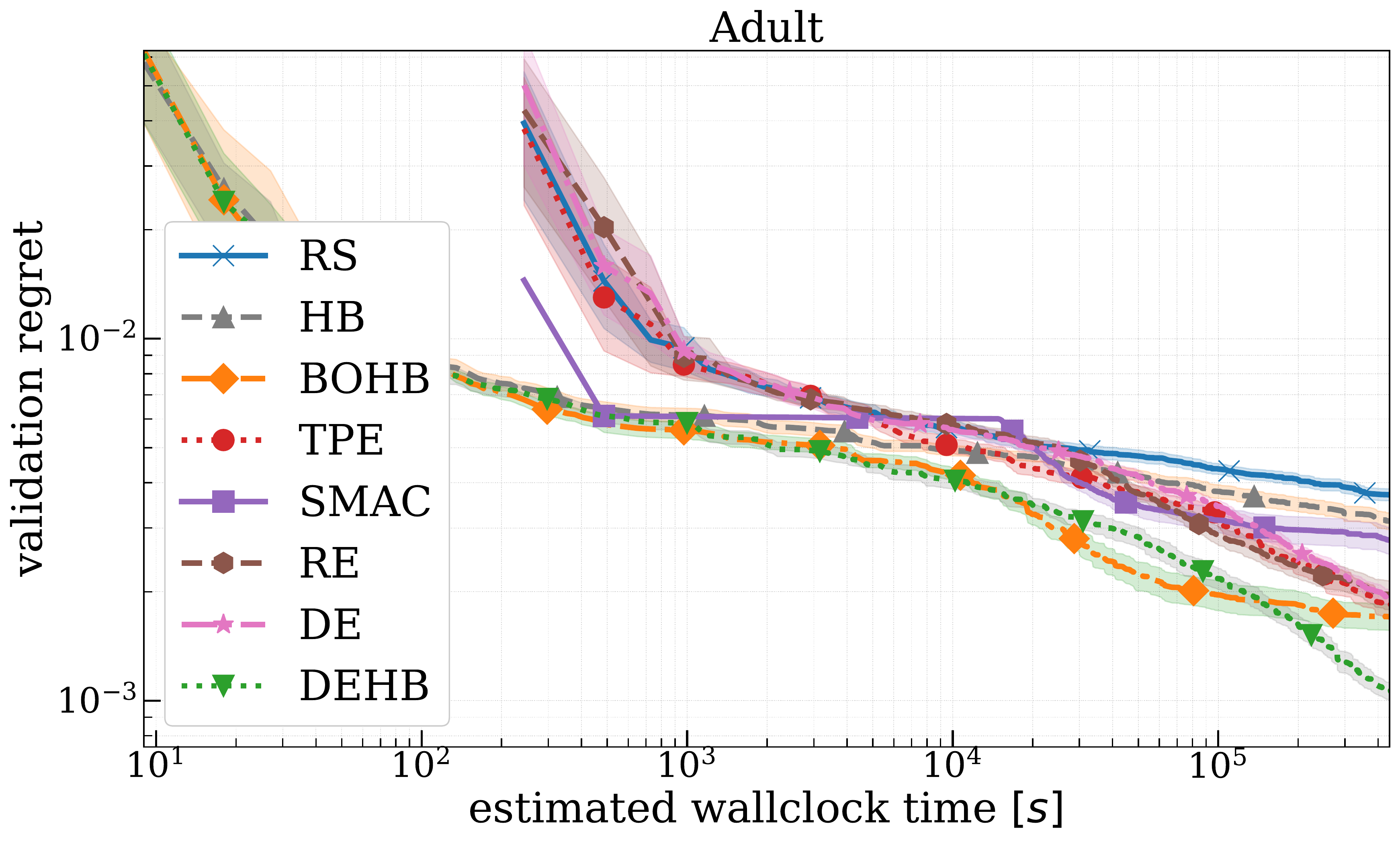} &
    \includegraphics[width=0.45\columnwidth]{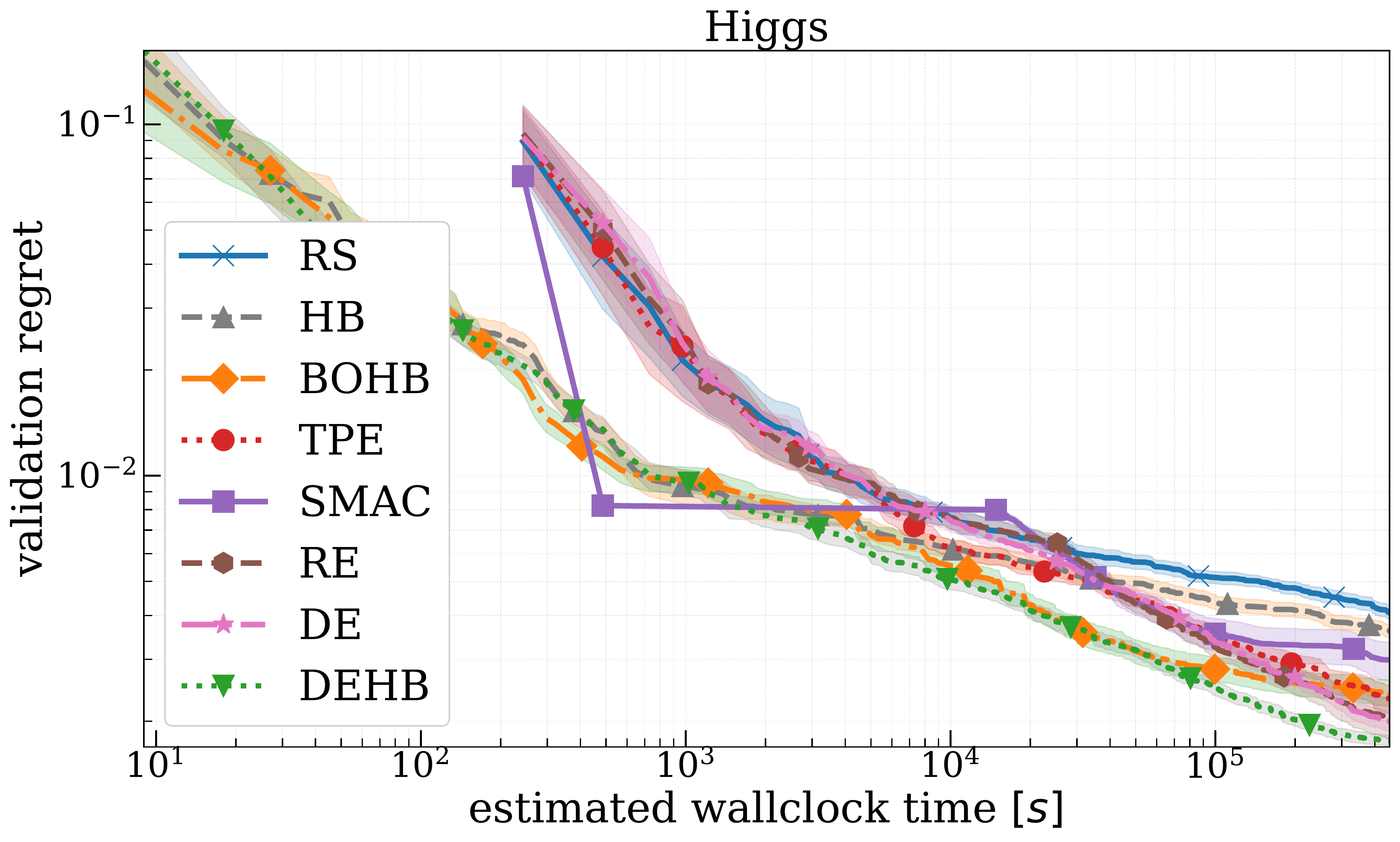} \\
    \includegraphics[width=0.45\columnwidth]{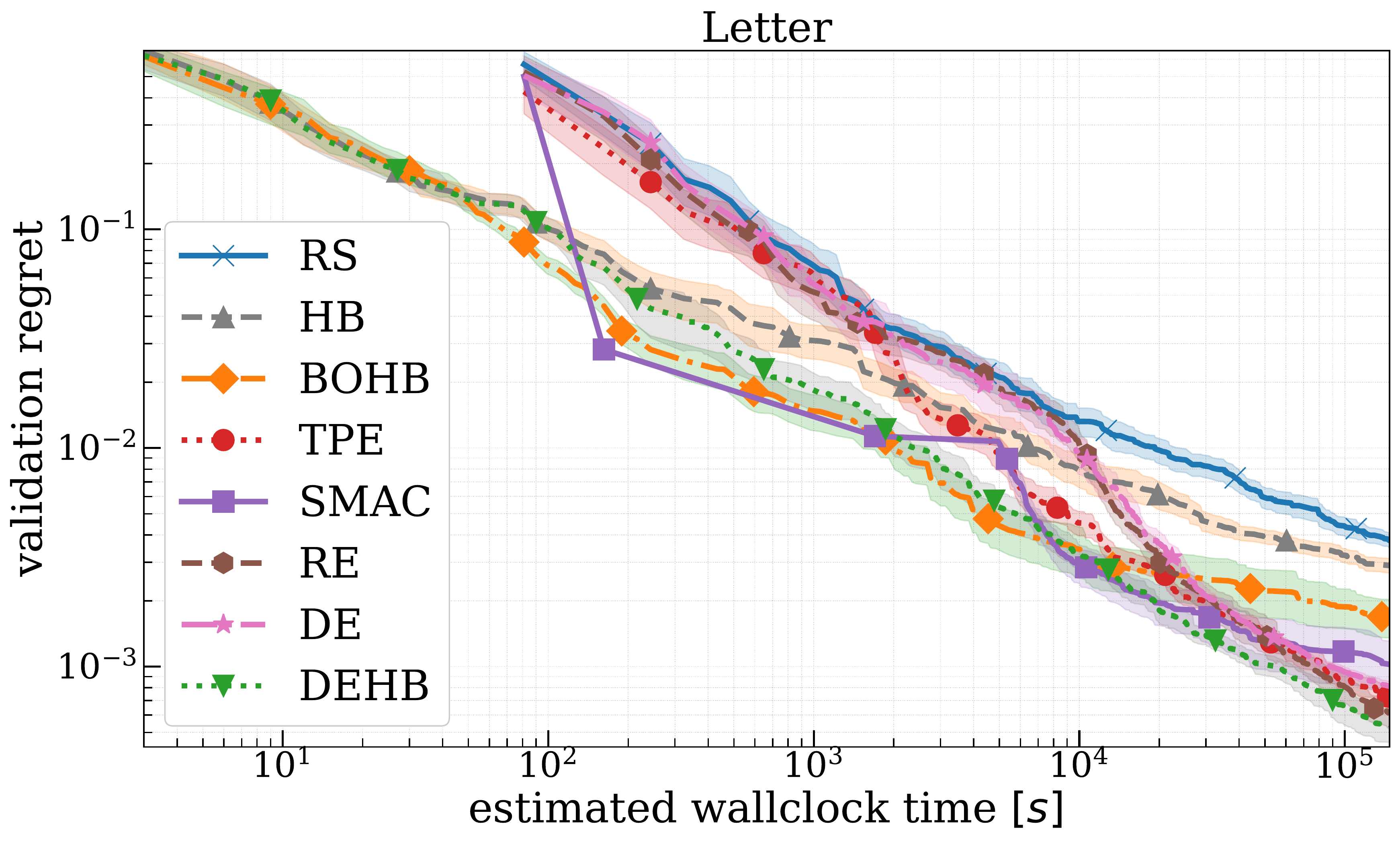} &
    \includegraphics[width=0.45\columnwidth]{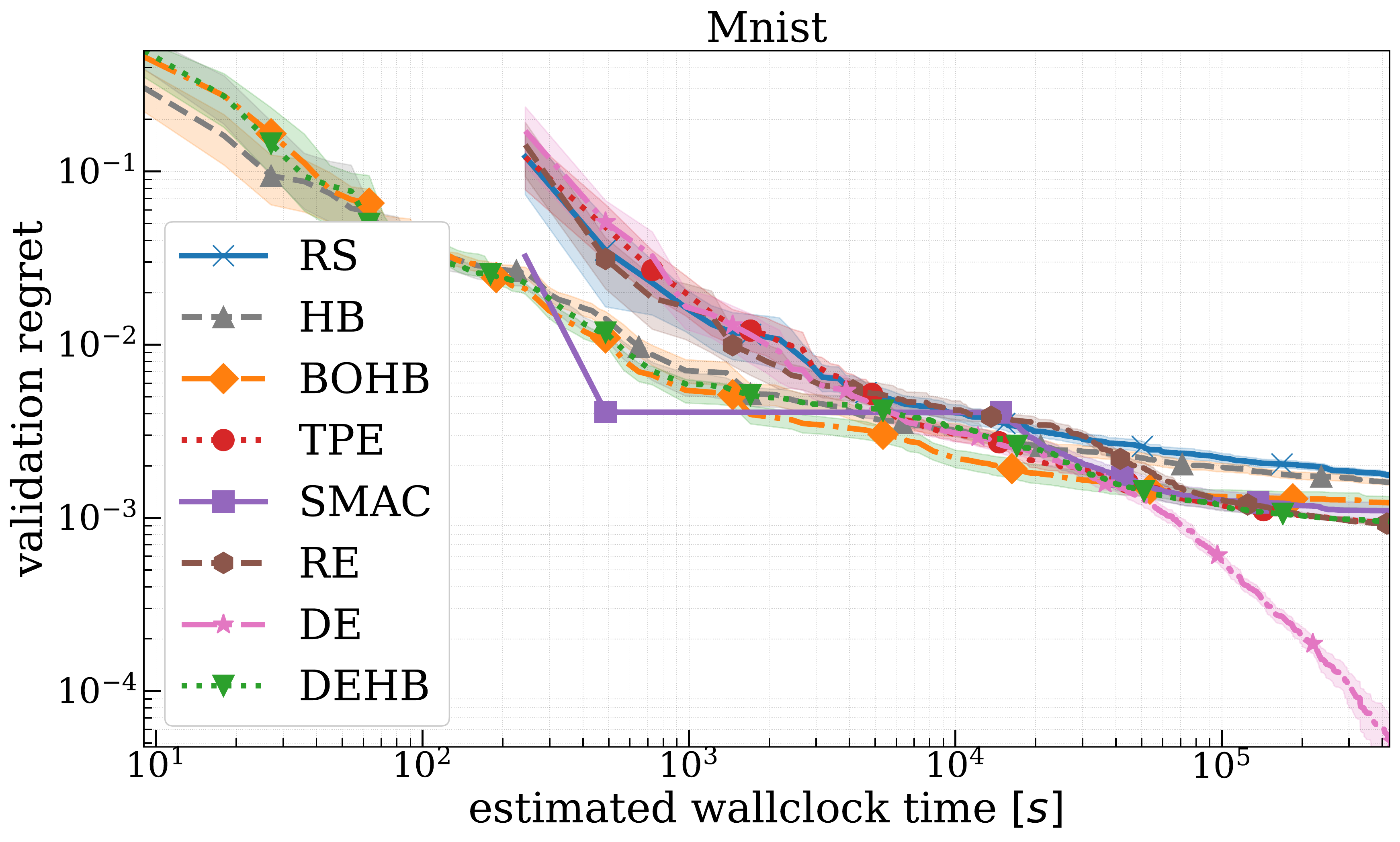} \\
    \includegraphics[width=0.45\columnwidth]{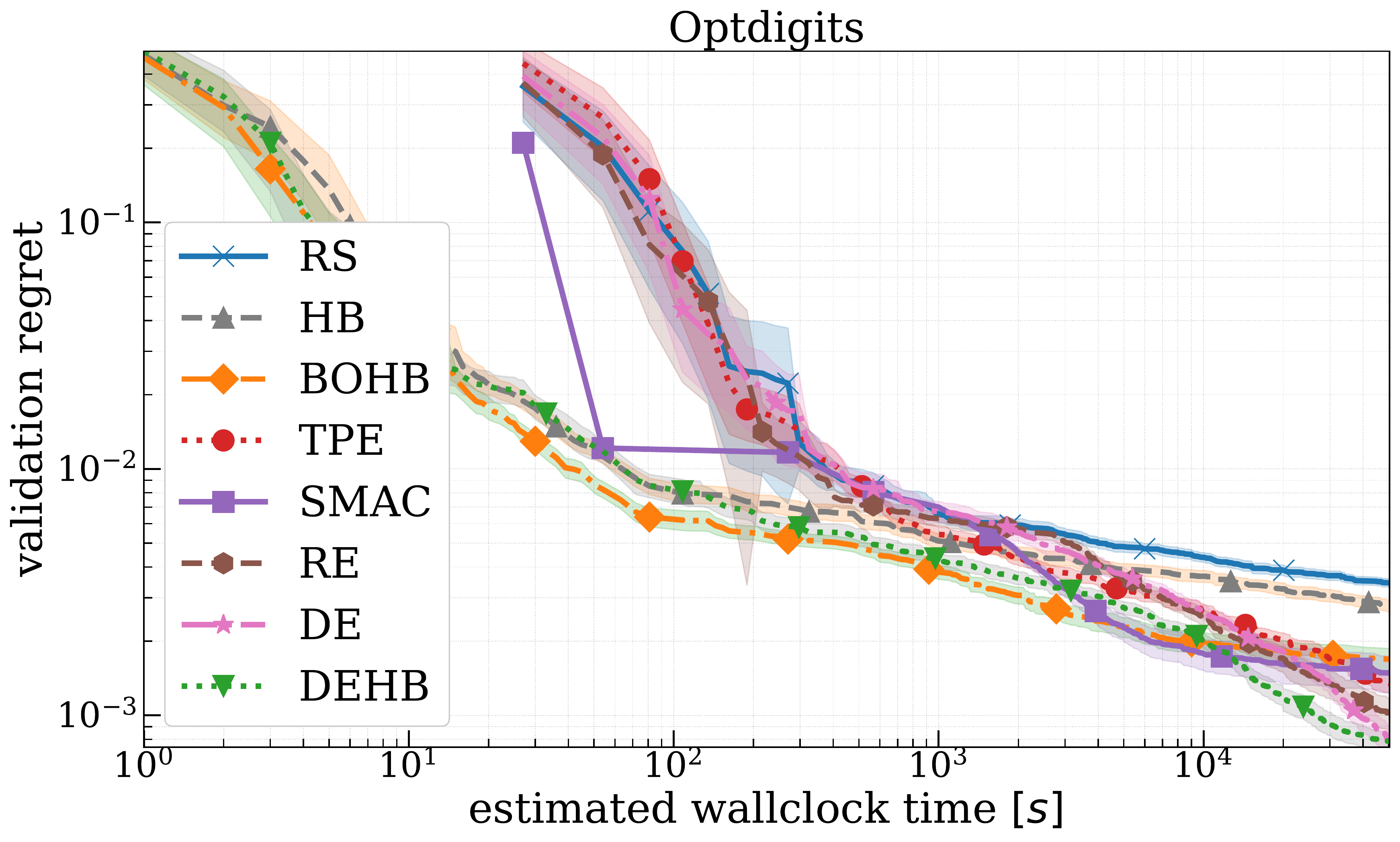} &
    \includegraphics[width=0.45\columnwidth]{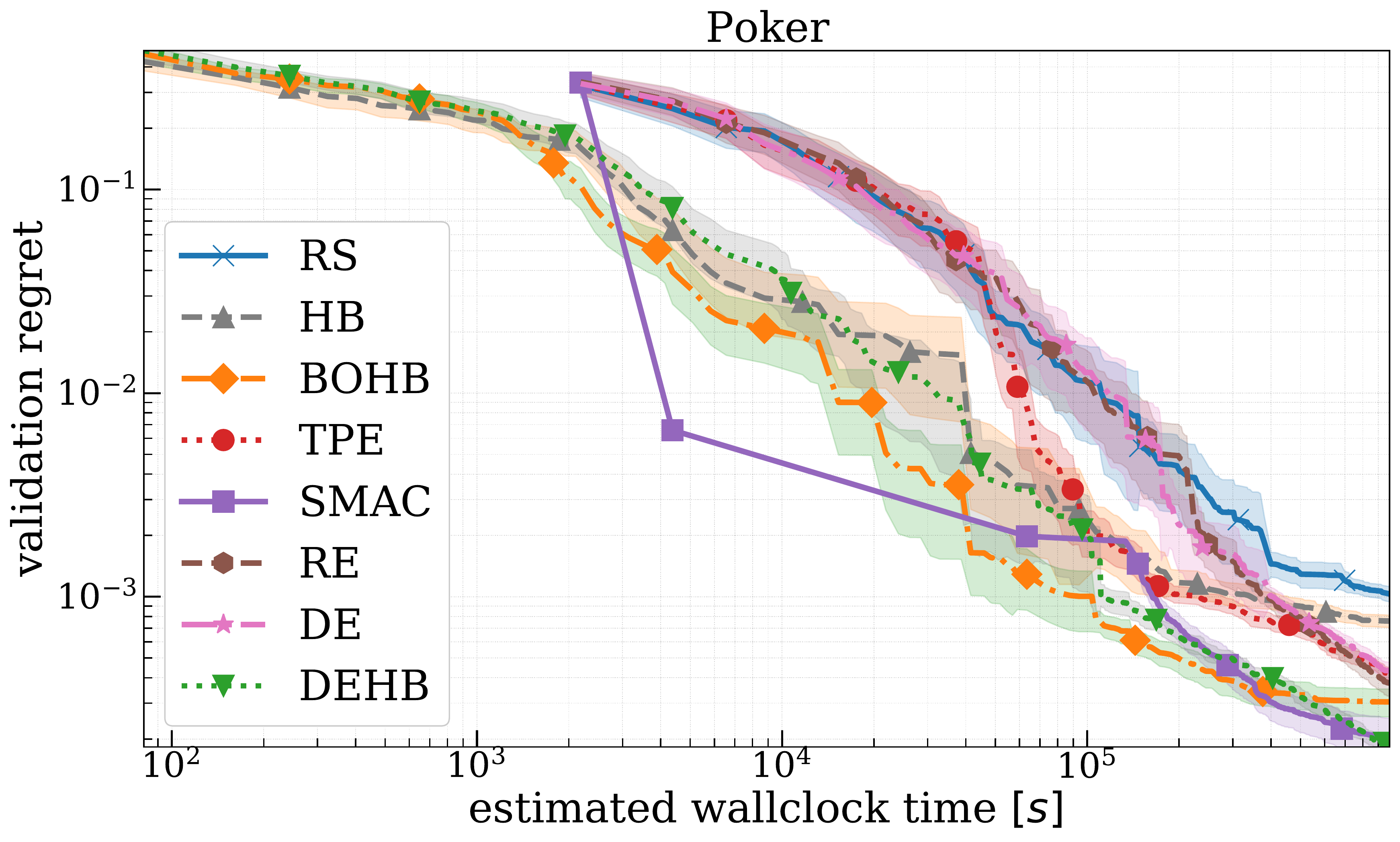} \\
\end{tabular}
\caption{Results for the OpenML surrogate benchmark for the $6$ datasets: Adult, Higgs, Letter, MNIST, Optdigits, Poker. The search space had $6$ continuous hyperparameters. All plots shown were averaged over $50$ runs of each algorithm.}
\label{fig:sub-OpenML}
\end{figure}




Figure \ref{fig:sub-OpenML}, show the results on all $6$ datasets from OpenML surrogates benchmark --- Adult, Letter, Higgs, MNIST, Optdigits, Poker. The surrogate model space is just 6-dimensional, allowing BOHB and TPE to build more confident models and be well-performing algorithms in this space, especially early in the optimization. However, DE and DEHB are able to remain competitive and consistently achieve an improved final performance than TPE and BOHB respectively. While even TPE achieves a better final performance than BOHB. Overall, DEHB is a competetive \textit{anytime} performer for this benchmark with the most robust final performances.




\subsection{Bayesian Neural Networks}\label{sec:app-bnn}
\begin{figure}[htbp]
\centering
\begin{tabular}{ll}
    \includegraphics[width=0.45\columnwidth]{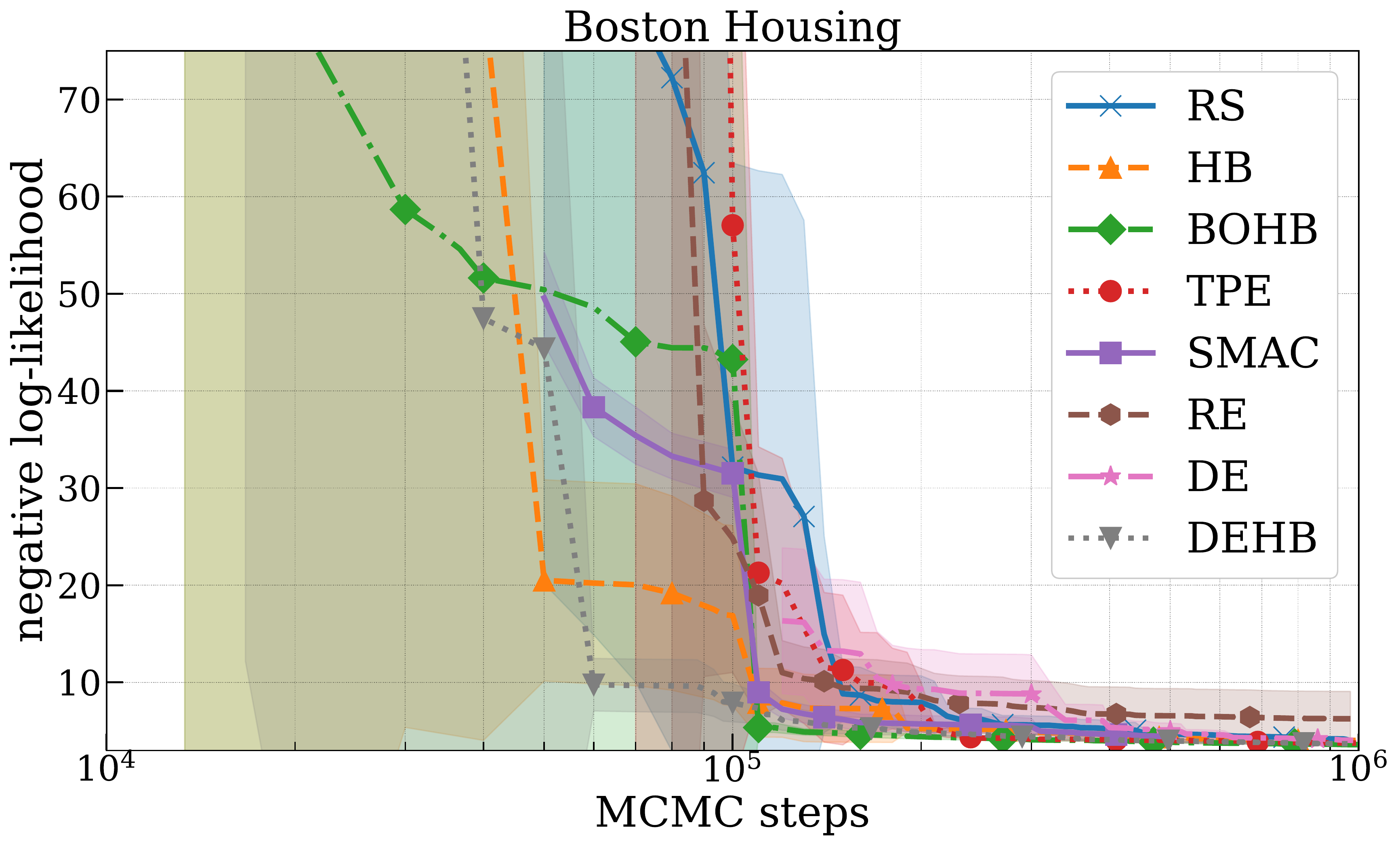} &
    \includegraphics[width=0.45\columnwidth]{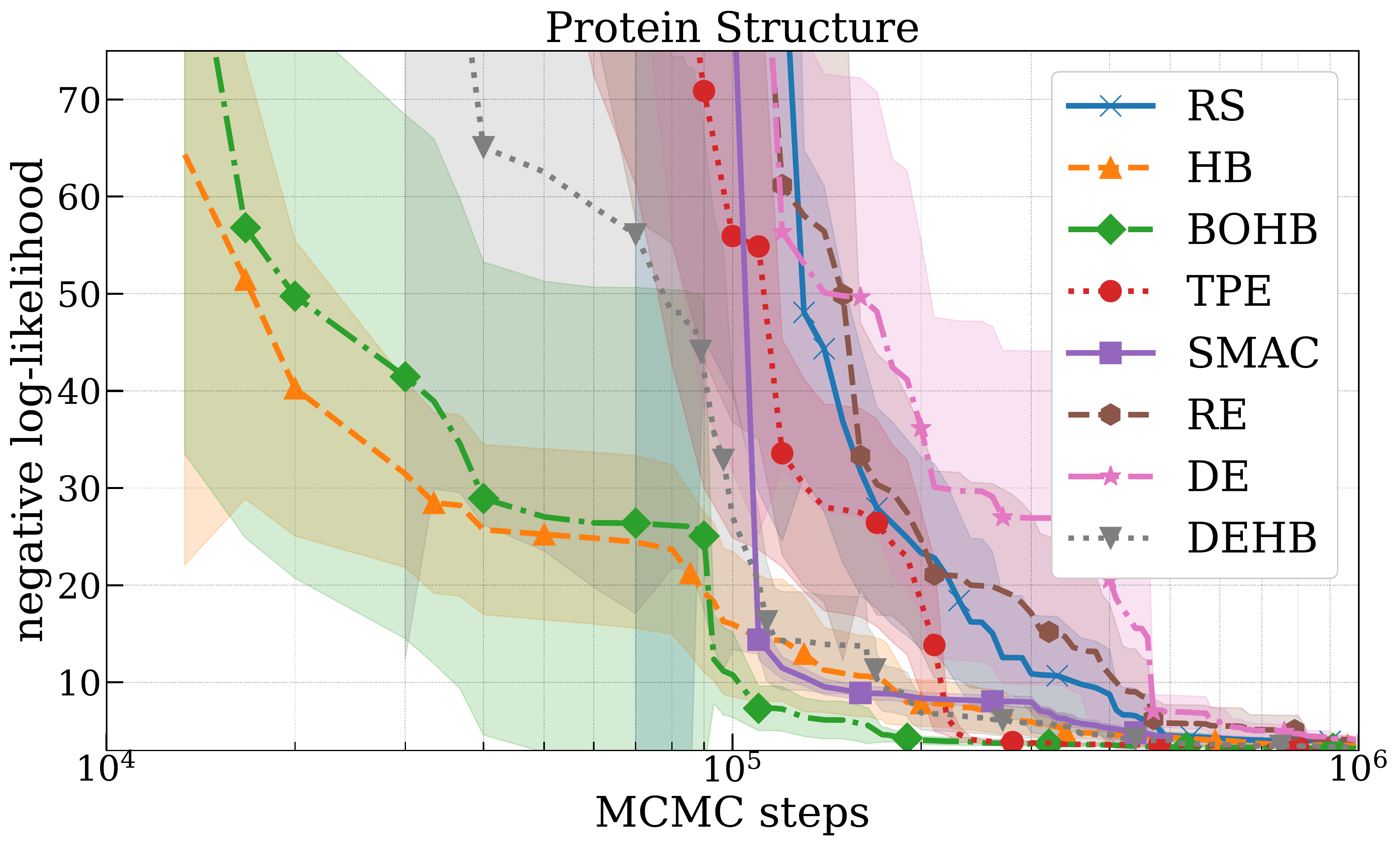} \\
\end{tabular}
\caption{Results for tuning $5$ hyperparameters of a Bayesian Neural Network on the the Boston Housing and Protein Structure datasets respectively, for $50$ runs of each algorithm.}
\label{fig:sub-bnn}
\end{figure}

The search space for the two-layer fully-connected Bayesian Neural Network is defined by $5$ hyperparameters which are: the step length, the length of the burn-in period, the number of units in each layer, and the decay parameter of the momentum variable.
In Figure \ref{fig:sub-bnn}, we show the results for the tuning of Bayesian Neural Networks on both the Boston Housing and Protein Structure datasets for the $6$-dimensional Bayesian Neural Networks benchmark. We observe that SMAC, TPE and BOHB are able to build models and reach similar regions of performance with high confidence. DEHB is slower to match in such a low-dimensional noisy space. However, given the same cumulative budget, DEHB achieves a competitive final score.

\subsection{Reinforcement Learning}\label{sec:app-rl}
\begin{figure}[htbp]
\centering
\begin{tabular}{ll}
    \includegraphics[width=0.45\columnwidth]{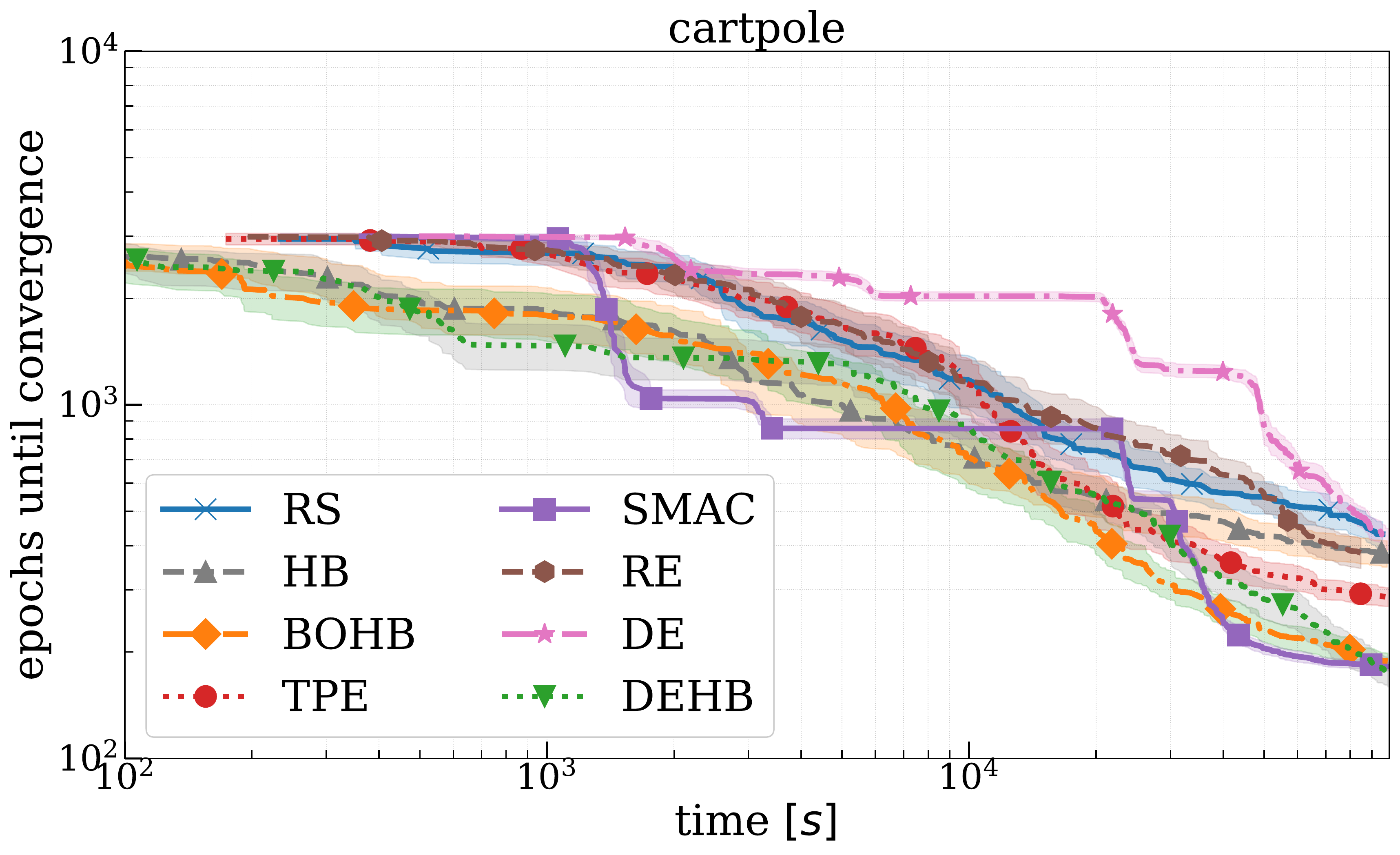} &
    \includegraphics[width=0.45\columnwidth]{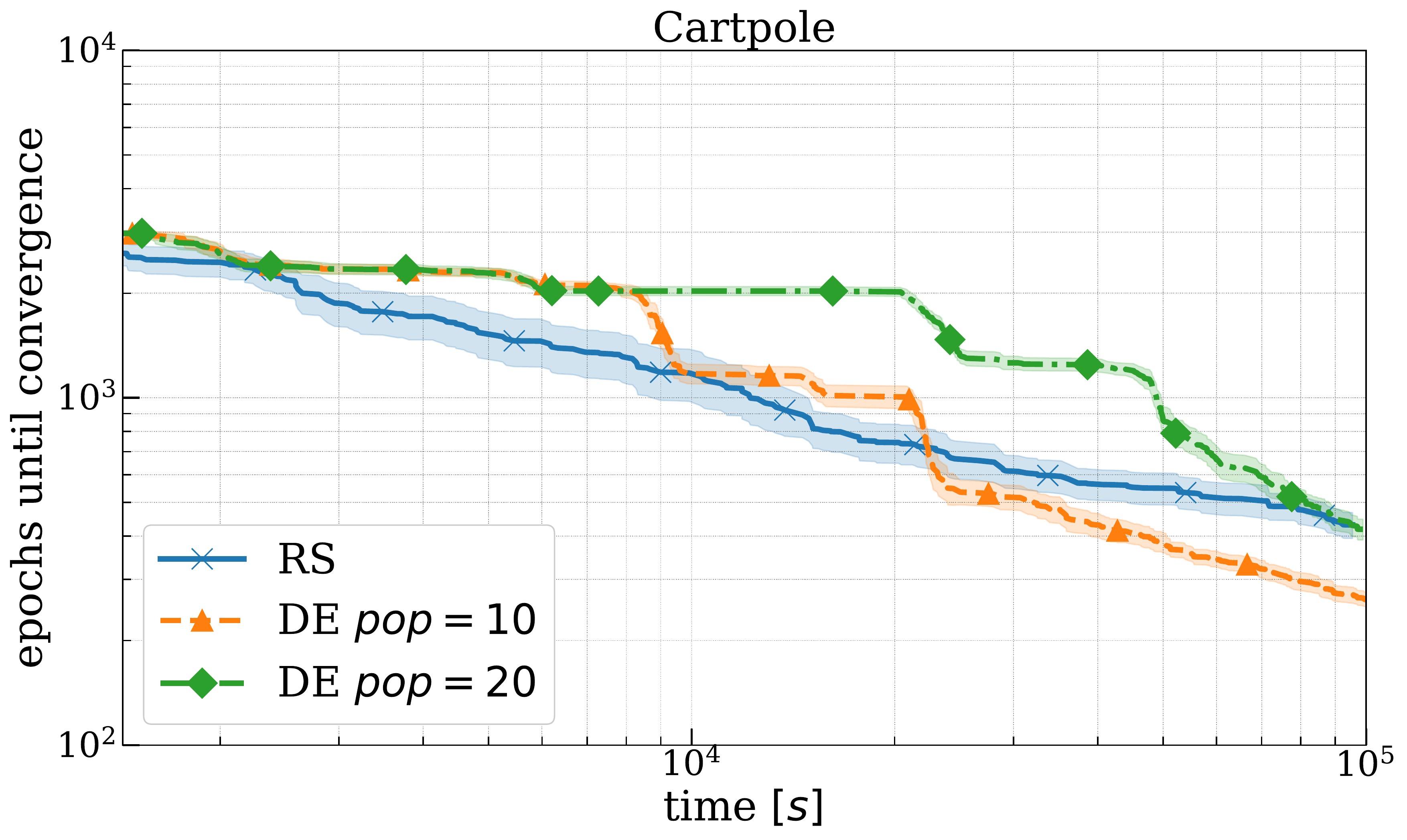} \\
\end{tabular}
\caption{(left) Results for tuning PPO on OpenAI Gym cartpole environment with $7$ hyperparameters. Each algorithm shown was run for $50$ runs. (right) Same experiment to compare DE with a population size of $10$ and $20$.}
\label{fig:sub-cartpole}
\end{figure}

For this benchamrk, the proximal policy optimization (PPO) ~\citep{schulman2017proximal} implementation is parameterized with $7$ hyperparameters: \# units layer 1, \# units layer 2, batch size, learning rate, discount, likelihood ratio clipping and entropy regularization. 
Figure \ref{fig:sub-cartpole} summarises the performance of all algorithms on the RL problem for the Cartpole benchmark. SMAC uses a SOBOL grid as its initial design and both its benefit and drawback can be seen as SMAC rapidly improves, stalls, and then improves again once model-based search begins. However, BOHB and DEHB both remain competitive and BOHB, DEHB, SMAC emerge as the top-$3$ for this benchmark, achieving similar final scores. We notice that the DE trace stands out as worse than RS and will explain the reason behind this. Given the late improvement for DE $pop=20$, we posit that this is a result of the \textit{deferred} updates of DE based on the classical DE \citep{awad2020iclr} update design and also the design of the benchmark. 

For classical-DE, the updates are \textit{deferred}, that is the results of the \textit{selection} process are incorporated into the population for consideration in the next evolution step, only after \textit{all} the individuals of the population have undergone evolution. In terms of computation, the wall-clock time for \textit{population size} number of function evaluations are accumulated, before the population is updated. In Figure \ref{fig:sub-cartpole} we illustrate why given how this benchmark is designed, this minor detail for DE slows down convergence. Along with a DE of population size $20$ as used in the experiments, we compare a DE of population size $10$ in Figure \ref{fig:sub-cartpole}. For the Reinforcement Learning benchmark from \citep{falkner-icml18a}, each full budget function evaluation consists of 9 trials of a maximum of 3000 episodes. With a population of $20$, DE will not inject a new individual into a population unless all $20$ individuals have participated as a parent in the crossover operation. This accumulates wallclock time equivalent to $20$ individuals $times$ $9$ trials $times$ time taken for a maximum of 3000 episodes. Which can explain the flat trajectories in the optimization trace for DE $pop=20$ in Figure \ref{fig:sub-cartpole} (right). DE $pop=10$ slashes this accumulated wallclock time in half and is able to inject newer configurations into the population \textit{faster} and is able to search faster. Given enough runtime, we expect DE $pop=20$ to converge to similar final scores. DEHB uses the $immediate$ update design for DE, wherein it updates the population immediately after a DE selection, and not wait for the entire population to evolve. We posit that this feature, along with lower fidelity search, and performing grid search over population sizes with Hyperband, enables DEHB to be more practical than classical-DE.

\subsection{NAS benchmarks} \label{sec:app-nasbench}

\subsubsection{NAS-Bench-101} \label{sec:app-101}

This benchmark was the first NAS benchmark relying on tabular lookup that was introduced to encourage research and reproducibility ~\citep{ying2019bench}. 
Each architecture from the search space is represented as a stack of cells. Each cell is treated as a directed acyclic graph (DAG) and the nodes and edges of these DAGs are parameterized which serve as the hyperparameters specifying a neural network architecture. NAS-Bench-101 offers a large search space of nearly 423k unique architectures that are trained on Cifar-10. The benchmark also offers a fidelity level --- training epoch length --- which allows HB, BOHB, and DEHB, to run on this benchmark. We run experiments on all 3 variants provided by NAS-101: Cifar \textit{A}, Cifar \textit{B}, Cifar \textit{C}. The primary search space discussed by ~\citep{ying2019bench} is Cifar \textit{A}; Cifar \textit{B} and Cifar \textit{C} are variants of the same search space with alternative encodings that deal with the hyperparameters defined on the edges of the DAG as categorical or continuous. 

In the NAS-Bench-101 benchmark, the correlation between the performance scores and the different budgets are small ~\citep{ying2019bench}, and therefore BOHB and DEHB do not yield better performance than the methods using full function evaluations only. All 3 evolutionary algorithms tested are able to exploit the discrete high-dimensional space much better than model-based methods such as BOHB and TPE, as seen by the performances of DE, DEHB and RE. While DEHB appears to be the algorithm with the best \textit{anytime} performance in the high-dimensional discrete NAS space. DE yields the final best performance, closely followed by DEHB.


\begin{figure}[htbp]
\centering
\begin{tabular}{ll}
    \includegraphics[width=0.45\columnwidth]{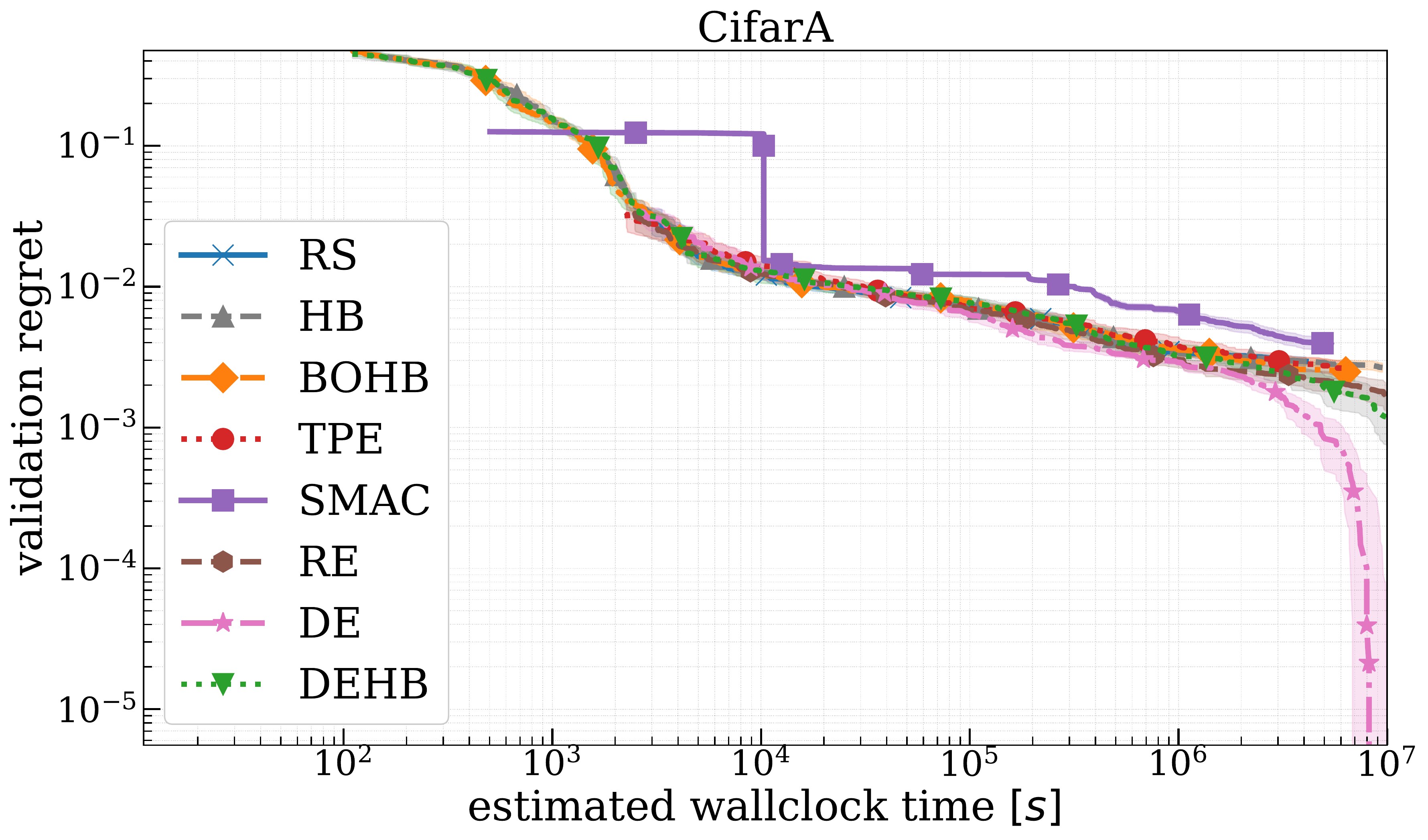} &
    \includegraphics[width=0.45\columnwidth]{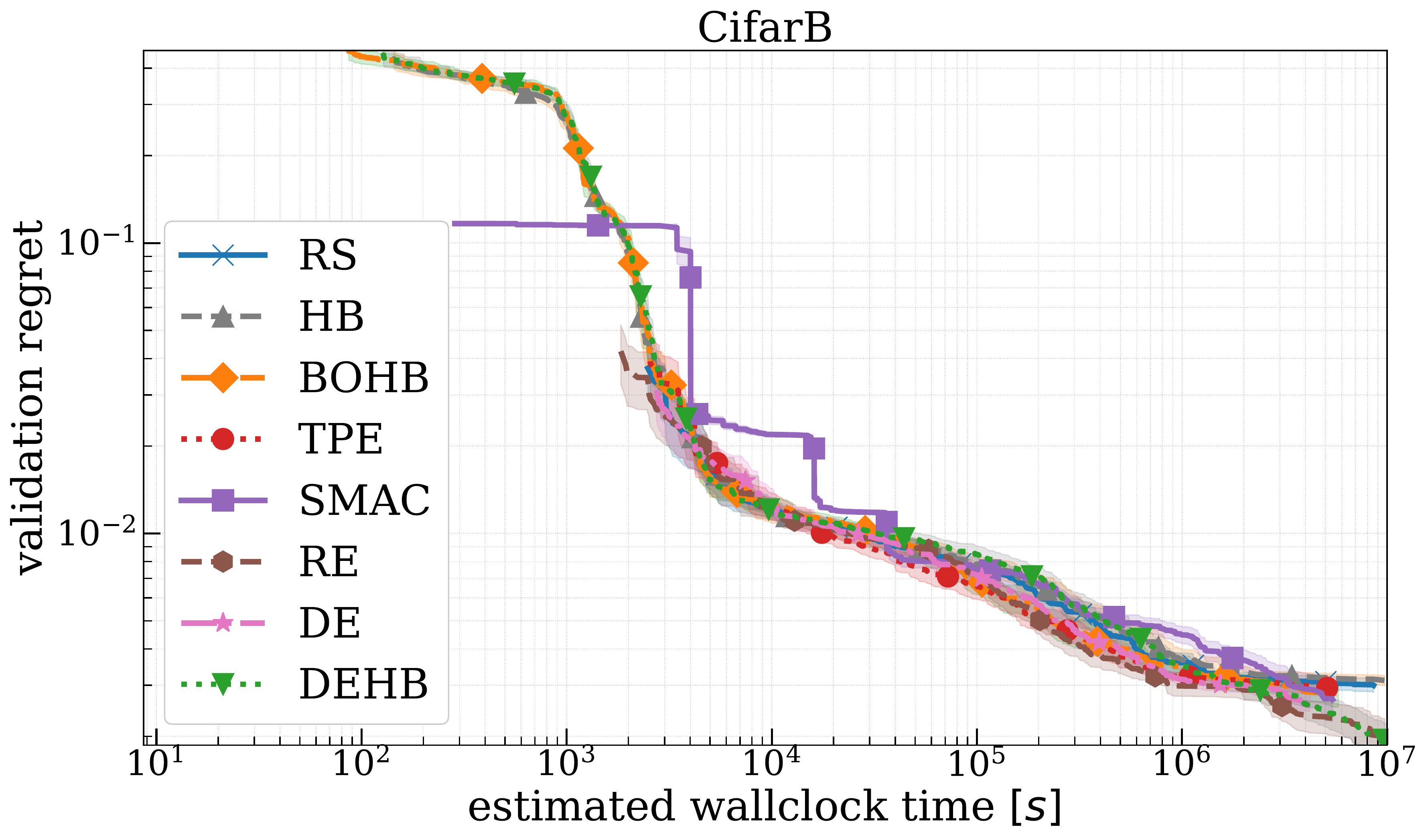} \\
\end{tabular}
\begin{tabular}{l}
    \includegraphics[width=0.45\columnwidth]{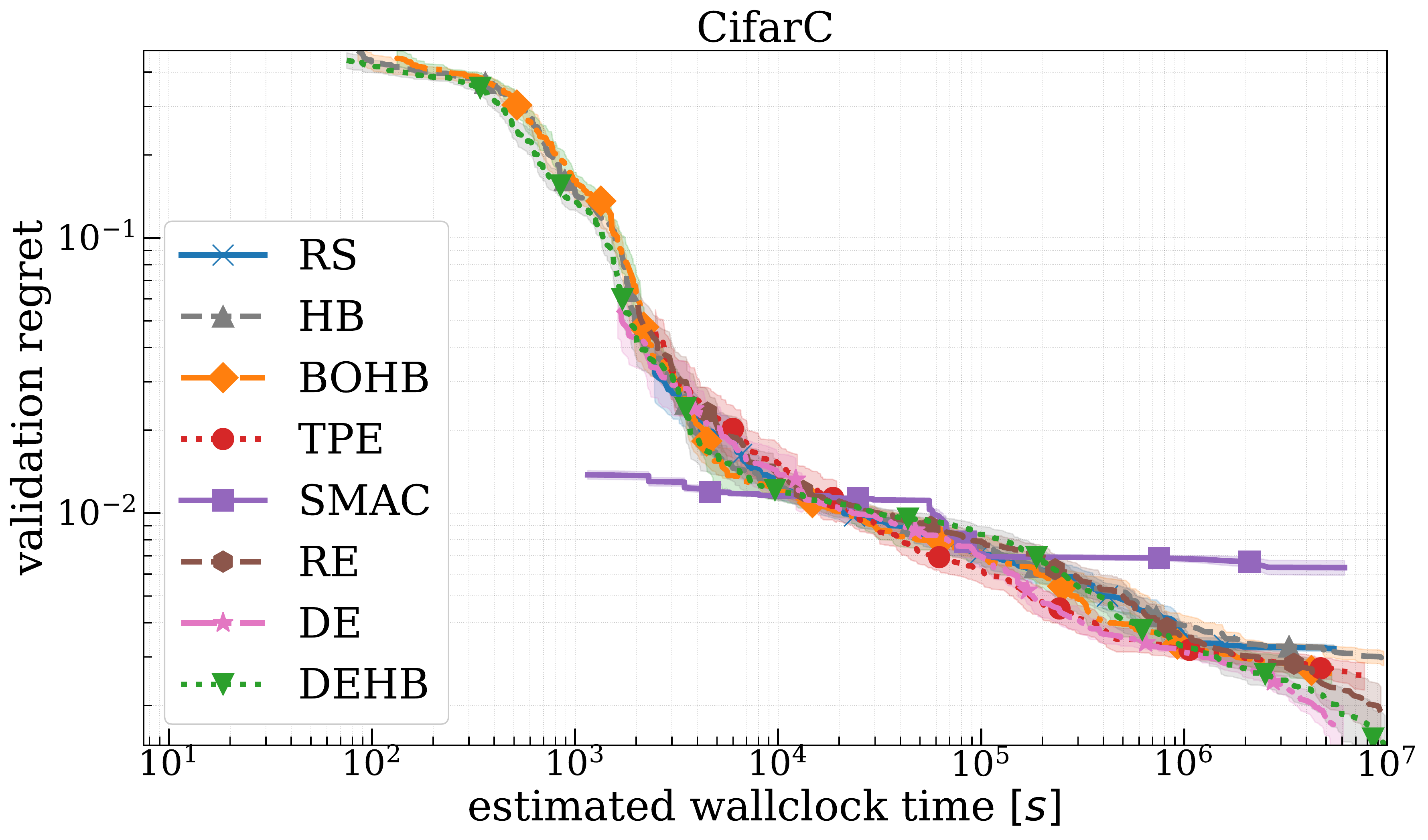} \\
\end{tabular}
\caption{Results for Cifar A, B and C from NAS-Bench-101 for $26, 14, 27 $-dimensional spaces respectively. All algorithms reported for $50$ runs.}
\label{fig:sub-nas101}
\end{figure}




\subsubsection{NAS-Bench-1shot1}\label{sec:app-1shot1}

\begin{figure}[htbp]
\centering
\begin{tabular}{ll}
    \includegraphics[width=0.45\columnwidth]{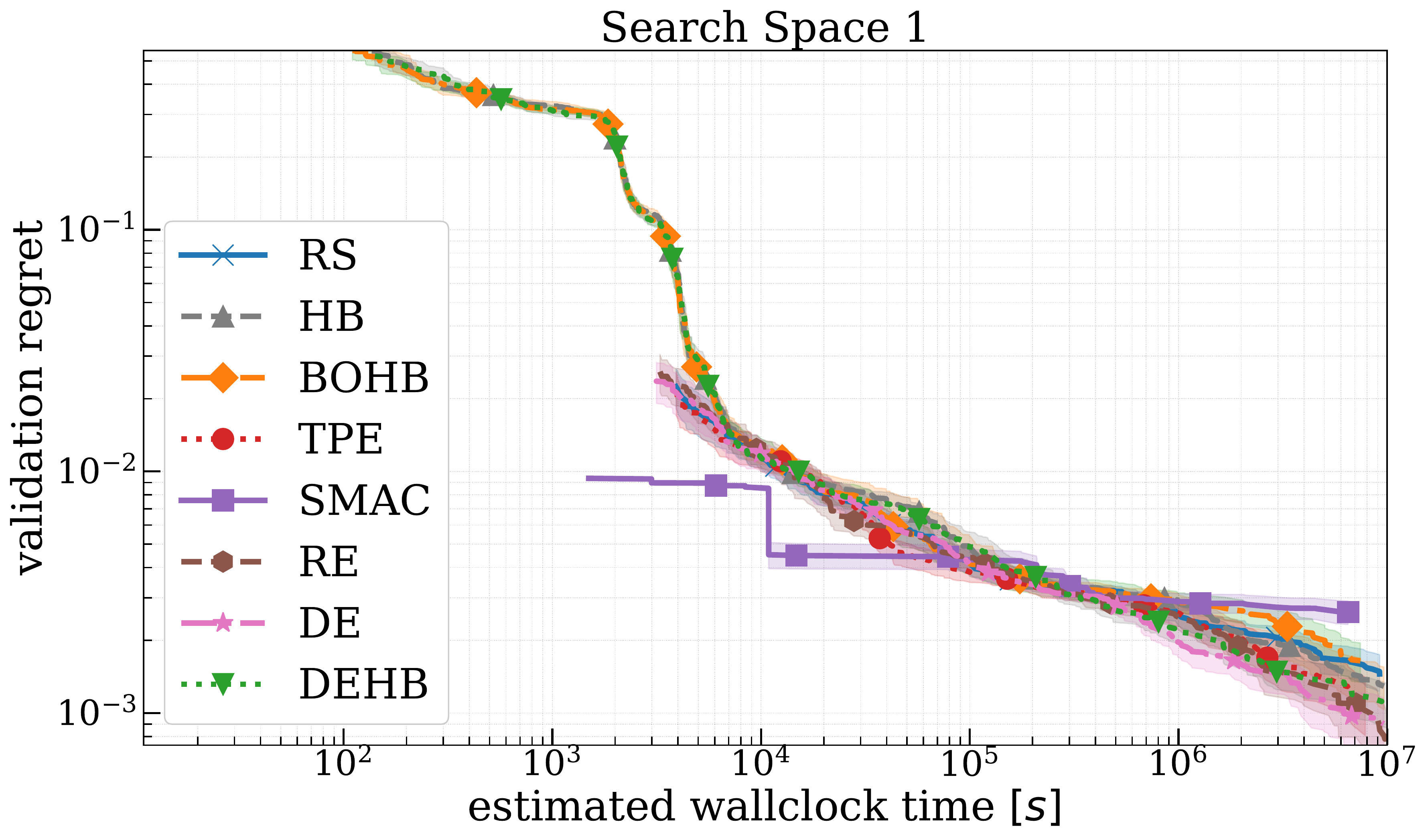} &
    \includegraphics[width=0.45\columnwidth]{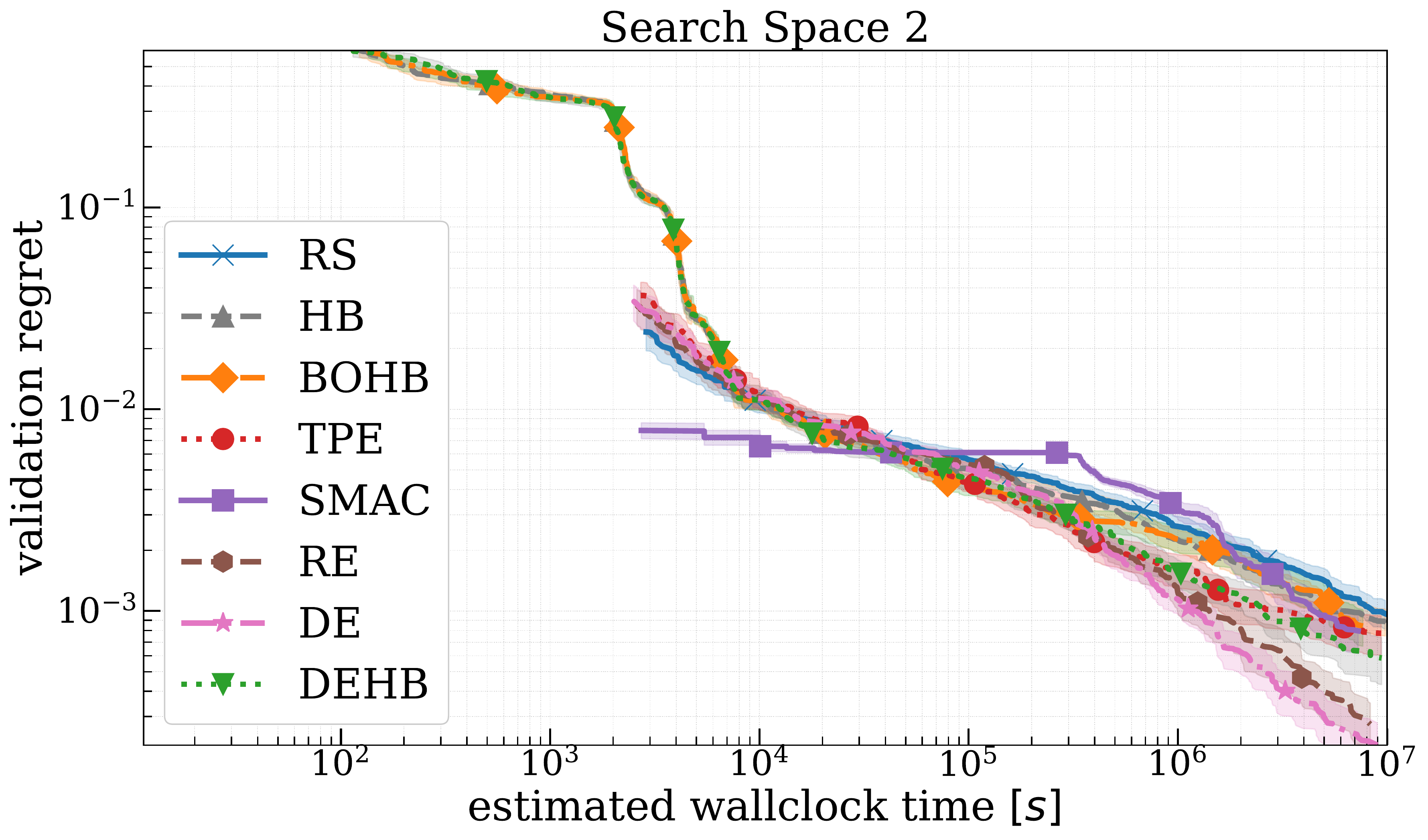} \\
\end{tabular}
\begin{tabular}{l}
    \includegraphics[width=0.45\columnwidth]{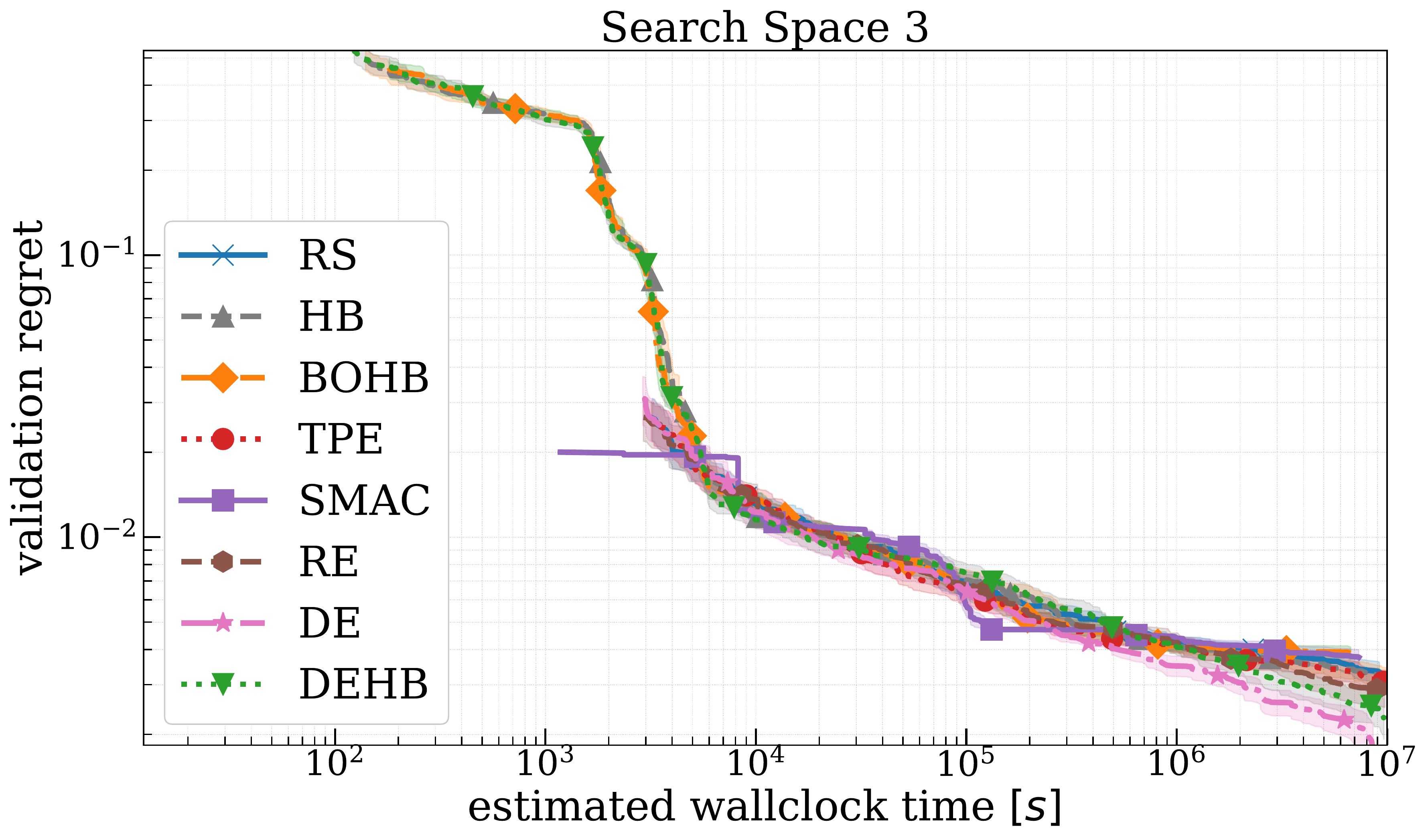} \\
\end{tabular}
\caption{Results for the $3$ search spaces from NAS-Bench-1shot1 for $50$ runs of each algorithm. The $3$ search spaces contains $9, 9, 11$ categorical parameters respectively.}
\label{fig:sub-nas1shot1}
\end{figure}

NAS-Bench-1shot1 was introduced by~\citep{zela2020bench}, as a benchmark derived from the large space of architectures offered by NAS-Bench-101. This benchmark allows the use of modern \textit{one-shot}\footnote{training a single large architecture that contains all possible architectures in the search space} NAS methods with weight sharing (~\citep{pham2018efficient}, ~\citep{liu2018darts}). The search space in NAS-Bench-1shot1 was modified to accommodate one-shot methods by keeping the macro network-level topology of the architectures similar and offering a different encoding design for the cell-level topology. This resulted in three search spaces: search space 1, search space 2 and search space 3 with 6240, 29160, and 363648 architectures respectively. In Figure \ref{fig:sub-nas1shot1}, we show the results on all 3 search spaces. We exclude weight sharing methods from the algorithms compared, in order to maintain parity across all experiments, while focusing on the objective of comparing black-box solvers.

From among RS, TPE, SMAC, RE and DE --- the full budget algorithms --- only DE is able to improve significantly as optimization proceeds. For the multi-fidelity algorithms --- HB, BOHB and DEHB --- only DEHB is able to improve and diverge away from HB by the end of optimization. BOHB, HB, RS, TPE and RE all appear to follow a similar trace showing the difficulty of finding good architectures in this benchmark. Nevertheless, the DE-based family of algorithms is able to further exploit the search space and show better performance than the other algorithms. Though DE performs the best, RE remains competitive, again suggesting the power of evolutionary methods on discrete spaces. Among model-based methods, only TPE competes with DEHB. 



\subsubsection{NAS-Bench-201} \label{sec:app-201}

\begin{figure}[htbp]
\centering
\begin{tabular}{ll}
    \includegraphics[width=0.45\columnwidth]{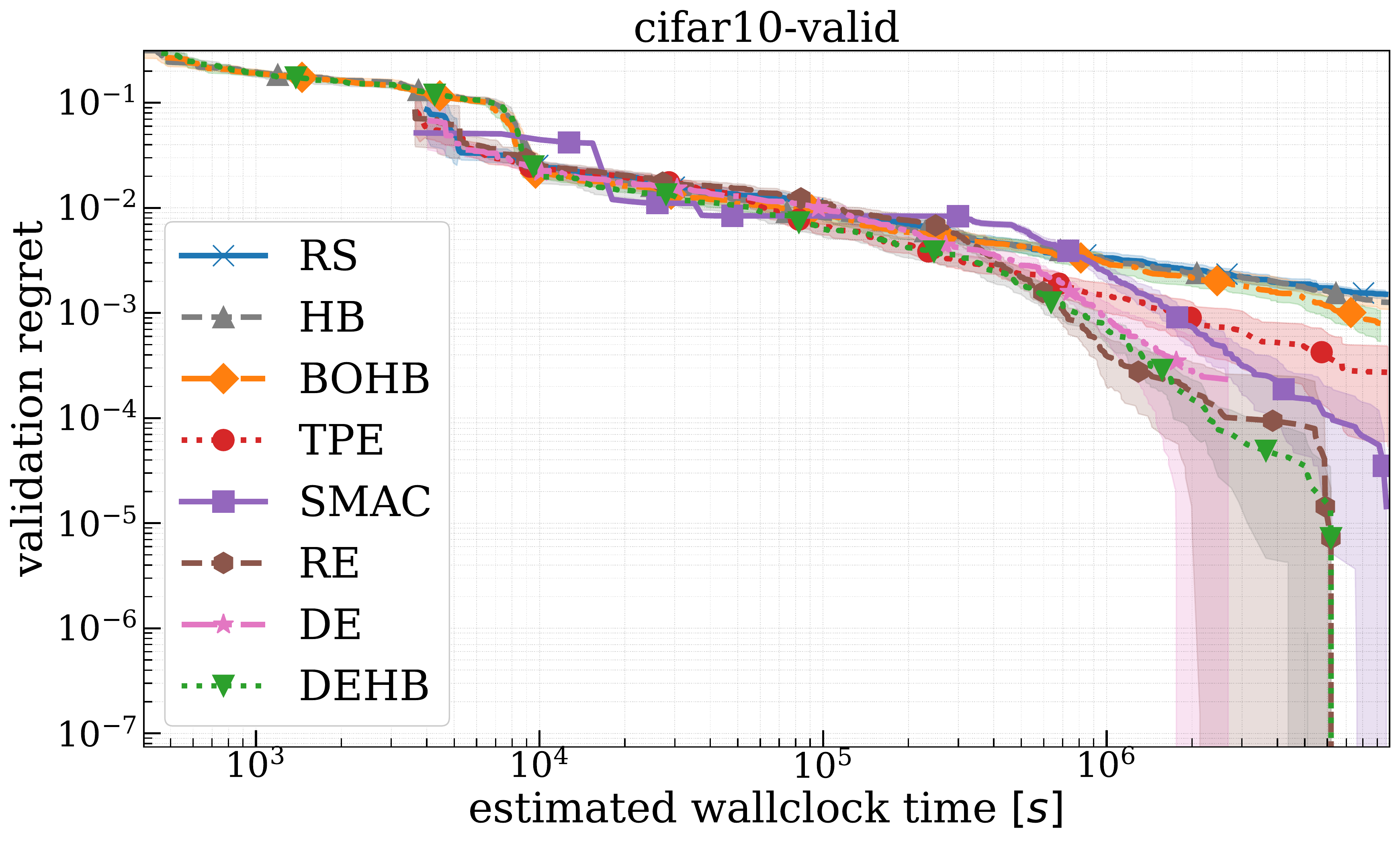} &
    \includegraphics[width=0.45\columnwidth]{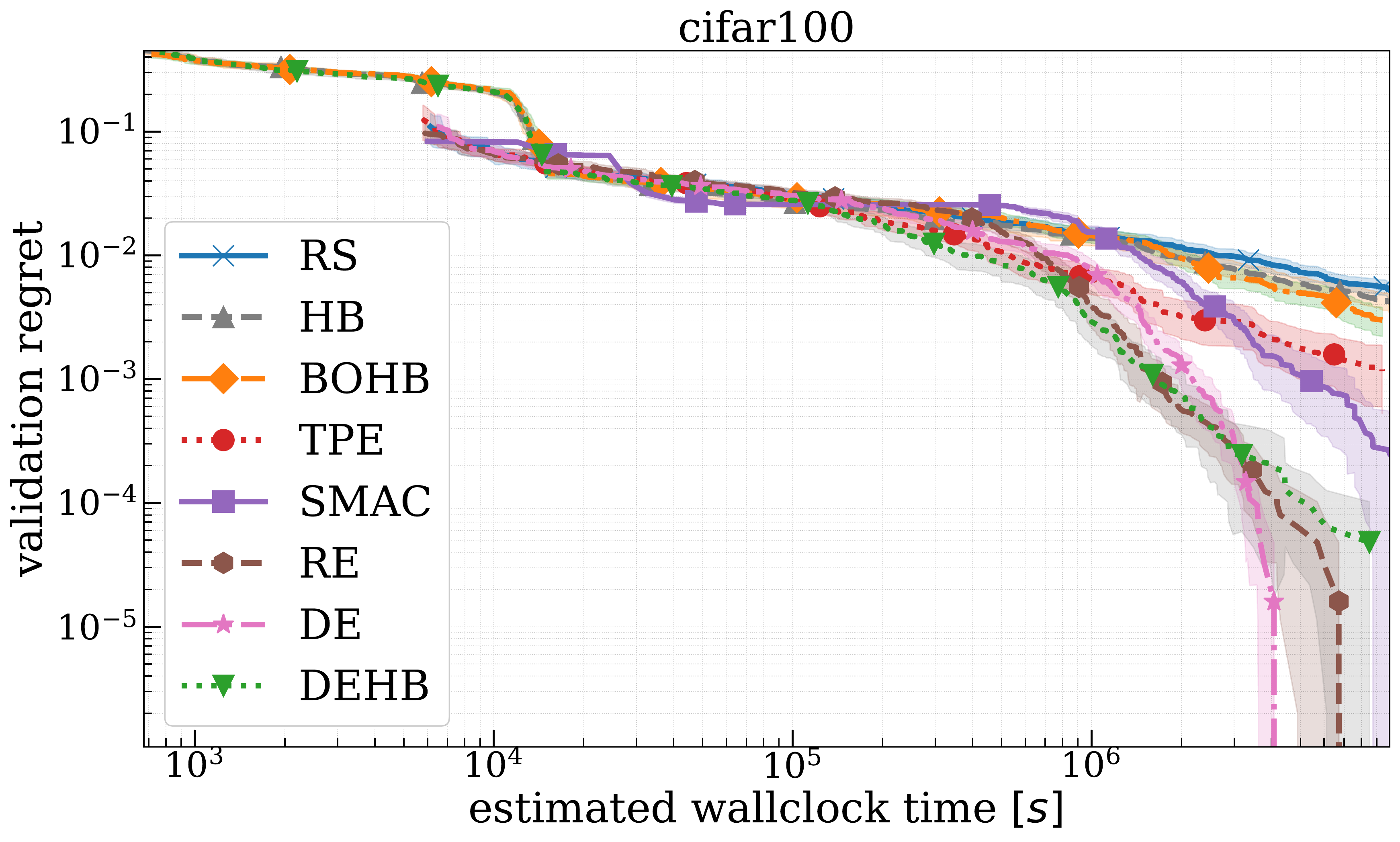} \\
\end{tabular}
\begin{tabular}{l}
    \includegraphics[width=0.45\columnwidth]{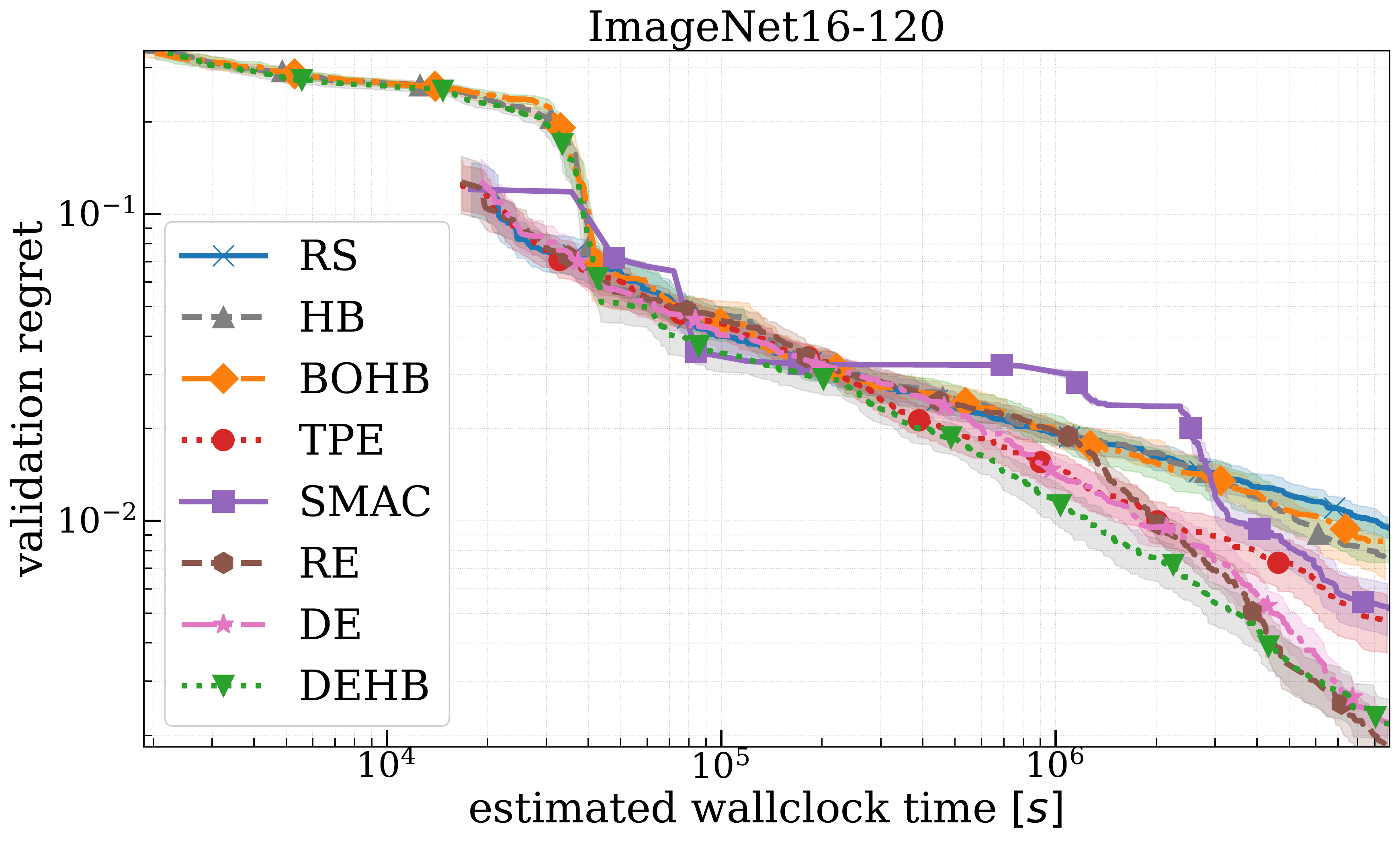} \\
\end{tabular}
\caption{Results for Cifar-10, Cifar-100, ImageNet16-120 from NAS-Bench-201 for $50$ runs of each algorithm. The search space contains $6$ categorical parameters.}
\label{fig:sub-nas201}
\end{figure}

To alleviate issues of direct applicability of weight sharing algorithms to NAS-Bench-101, ~\citep{dong2020bench} proposed NAS-Bench-201. This benchmark contains a \textit{fixed cell search space} having DAGs with 4 nodes as the cell structure, and the edges of the DAG cells representing operations. The search space by design contains $6$ discrete/categorical hyperparameters. 
NAS-Bench-201 provides a lookup table for 3 datasets: Cifar-10, Cifar-100 and ImageNet16-120, along with a fidelity level as number of training epochs. 
The search space for all 3 datasets include 15,625 cells/architectures. 
From the validation regret performances in Figure \ref{fig:sub-nas201}, it is clear that DEHB quickly converges to strong solutions which are a few orders of magnitude better than BOHB and RS (in terms of regret).
DE and RE are both competitive with RE converging slightly faster than DE. 
Notably, DEHB is the only multi-fidelity algorithm in this experiment that works well. 

NAS-Bench-201 specifies the same $6$-dimensional discrete hyperparameter space for the Cifar-10 and Cifar-100 datasets. Figure \ref{fig:sub-nas201} again shows that the evolutionary algorithms perform the best in a space defined by categorical parameters. SMAC in this scenario is the \textit{best}-of-the-rest, outside of DEHB, RE and DE. BOHB evidently struggles to be even significantly better than HB.

\subsubsection{NAS-HPO-Bench}\label{sec:app-nas-hpo}
To facilitate HPO research involving feed-forward neural networks, \citep{klein2019tabular} introduced NAS-HPO-Bench with a search space composed of hyperparameters that parameterize the architecture of a 2-layer feed-forward network\footnote{additionally, a linear output layer}, along with hyperparameters for its training procedure. The primary difference between NAS-HPO from the OpenML surrogates benchmark is that in the latter, a random forest model was used as a surrogate to approximate the performance for configurations. NAS-HPO-Bench is designed in the same vein as the other NAS benchmarks discussed earlier. For the total of 9 discrete hyperparameters (4 for architecture + 5 for training), all 62208 configurations resulting from a grid search over the search space were evaluated to yield a tabular representation for configuration and performance mapping. The benchmark provides such lookup tables for 4 popular UCI regression datasets: \textit{Protein Structure}, \textit{Slice Localization}, \textit{Naval Propulsion} and \textit{Parkinsons Telemonitoring}. NAS-HPO-Bench also provides the number of training epochs as a fidelity level. 

\begin{figure}[htbp]
\centering
\begin{tabular}{ll}
    \includegraphics[width=0.45\columnwidth]{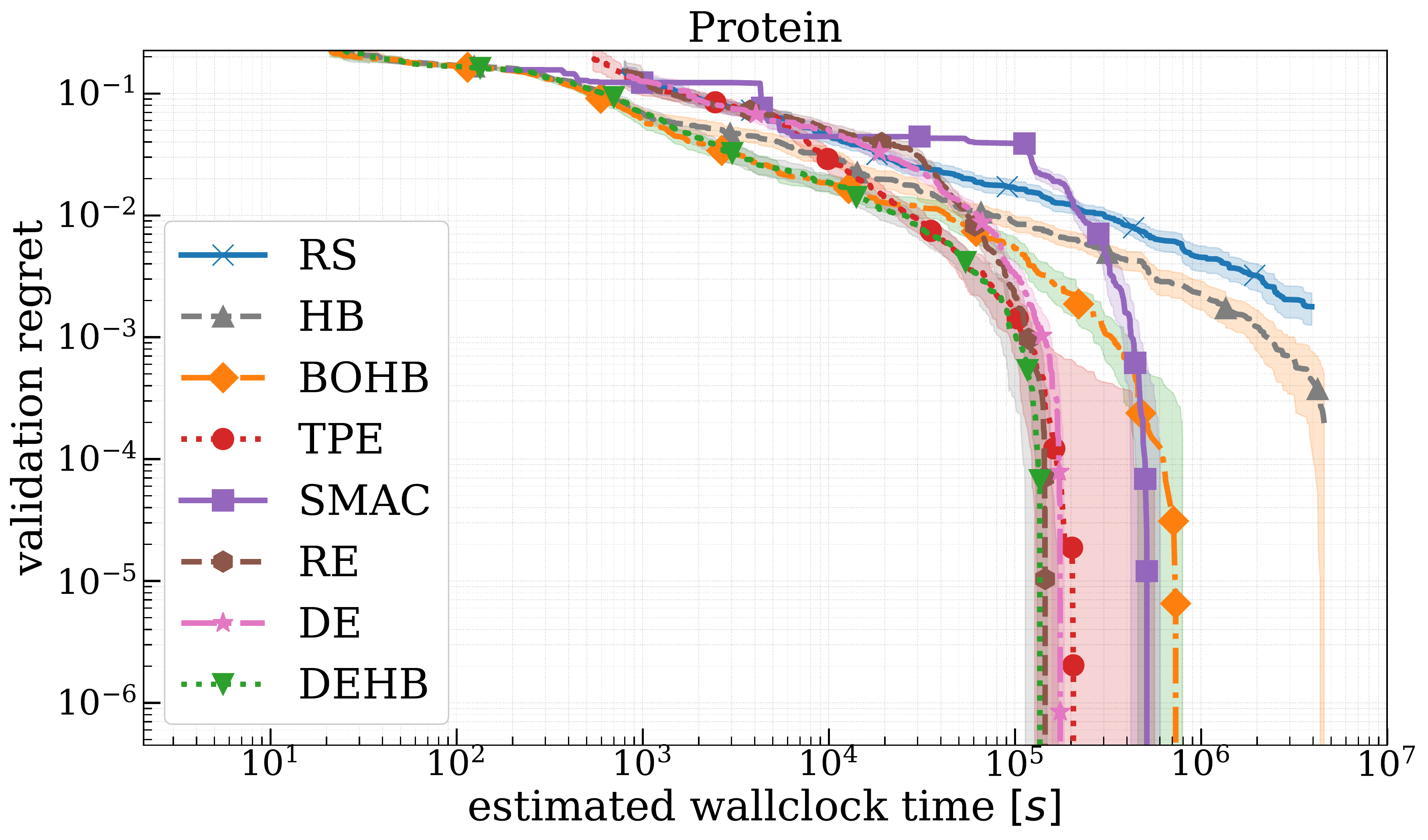} &
    \includegraphics[width=0.45\columnwidth]{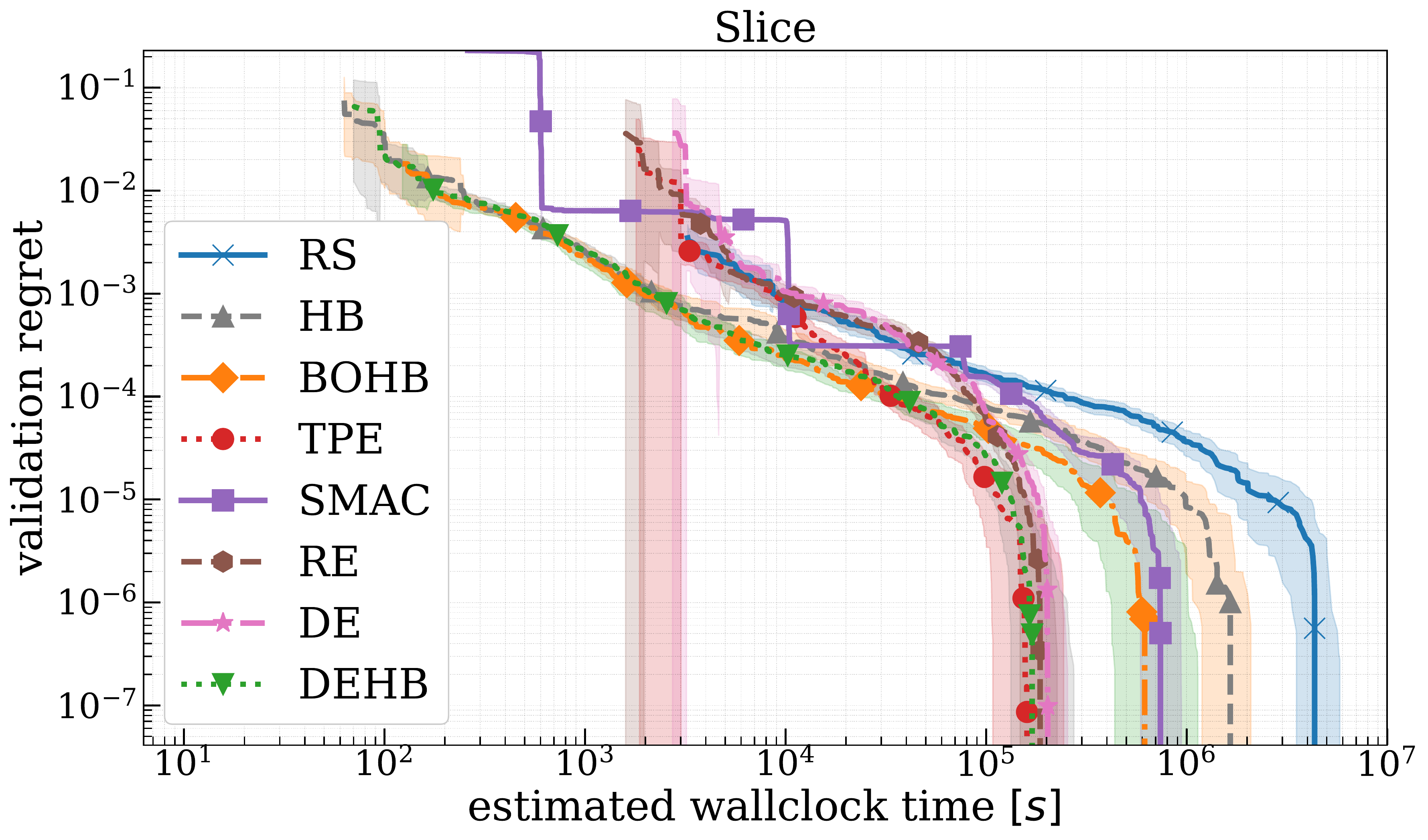} \\
    \includegraphics[width=0.45\columnwidth]{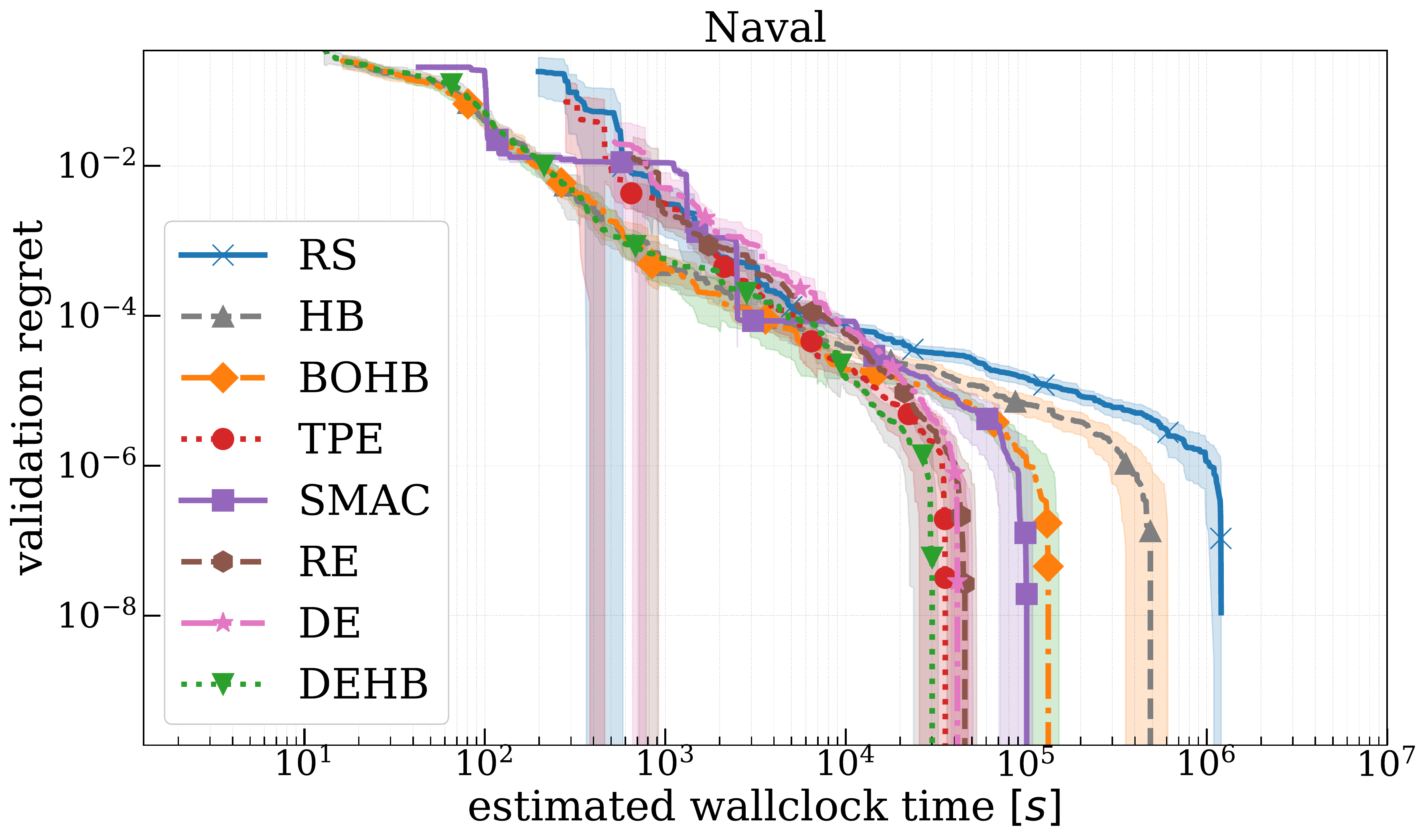} &
    \includegraphics[width=0.45\columnwidth]{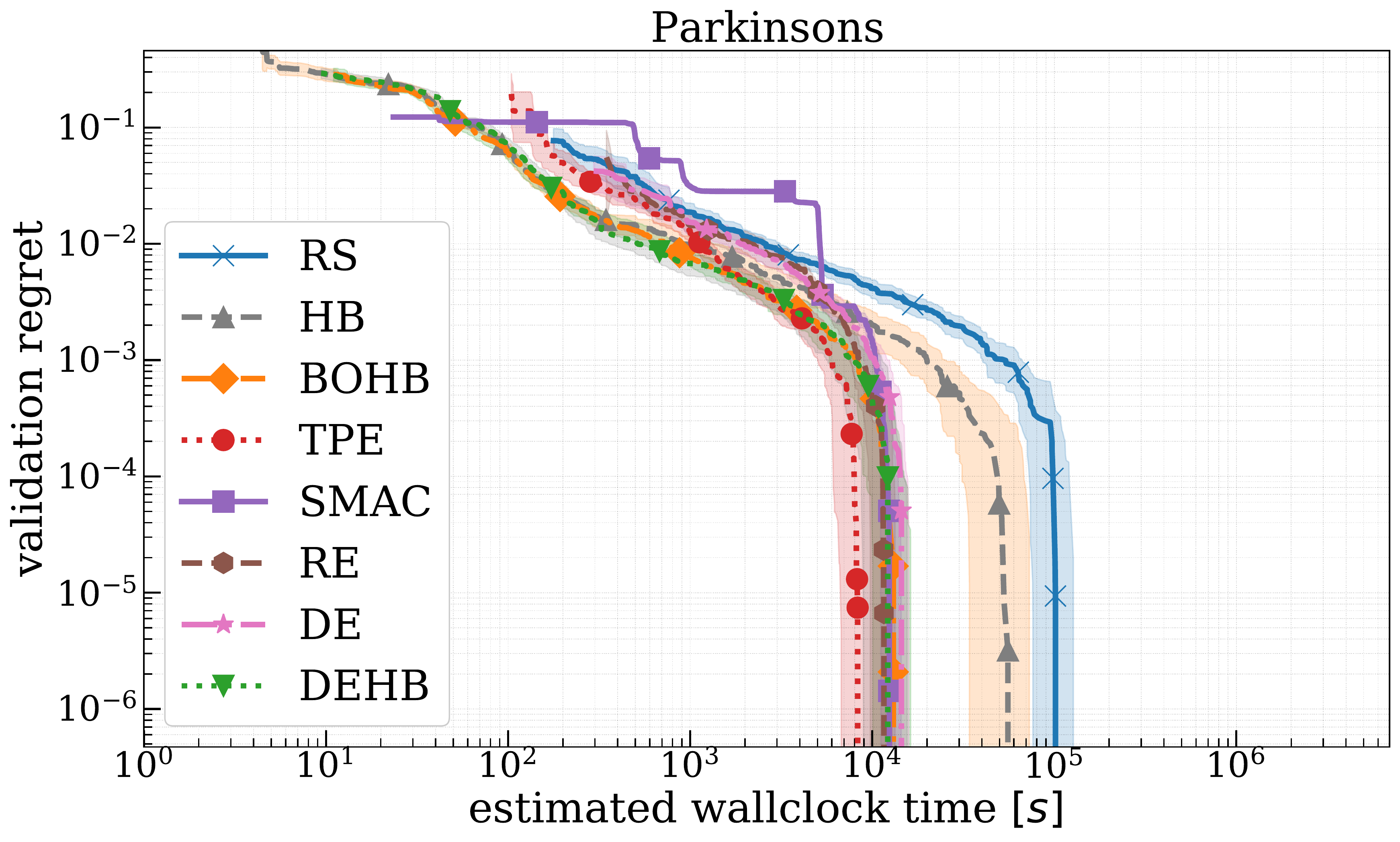} \\
\end{tabular}
\caption{Results for the Protein Structure, Slice Localization, Naval Propulsions, Parkinsons Telemonitoring datasets from NAS-HPO-Bench for 50 runs of each algorithm. The search space contains $9$ hyperparameters.}
\label{fig:sub-nashpo}
\end{figure}


Figure \ref{fig:sub-nashpo} illustrates the performance of all algorithms on the $4$ datasets provided in NAS-HPO-Bench. As it appears, barring RS and HB, all other algorithms are able to obtain similar final scores for the benchmark with respect to the validation set. BOHB and DEHB both diverge from HB and start improving early on. However, DEHB continues to improve and is able to converge the fastest. TPE, RE, DE all compete with each other in terms of convergence rate, while BOHB and SMAC show similar convergence speeds.

\subsection{Comparison of DEHB to BO-based multi-fidelity methods} \label{sec:app-bo-exps}
\begin{figure}[!ht]
\centering
\begin{tabular}{ll}
    \includegraphics[width=0.45\columnwidth]{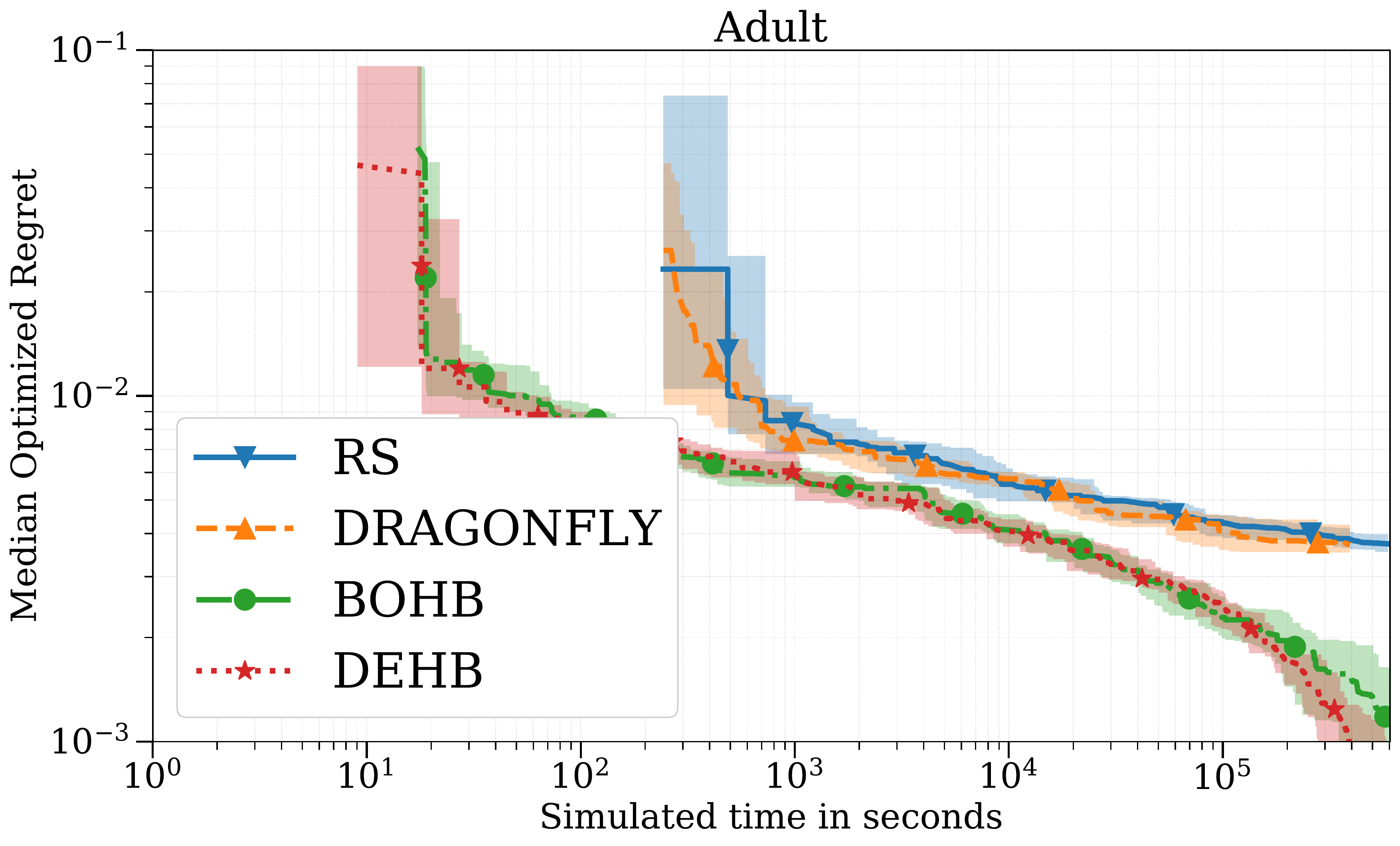} &
    \includegraphics[width=0.45\columnwidth]{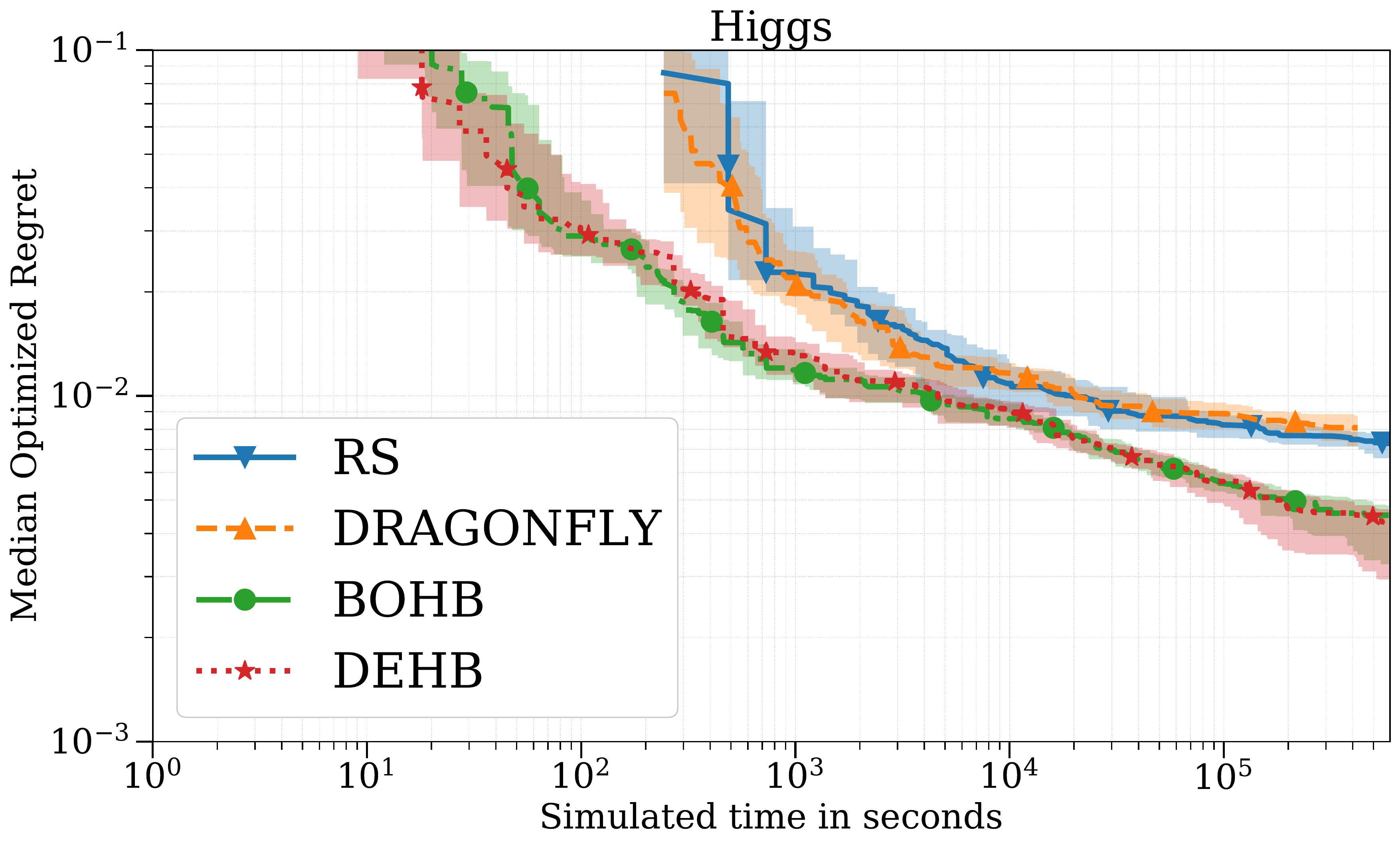} \\
    \includegraphics[width=0.45\columnwidth]{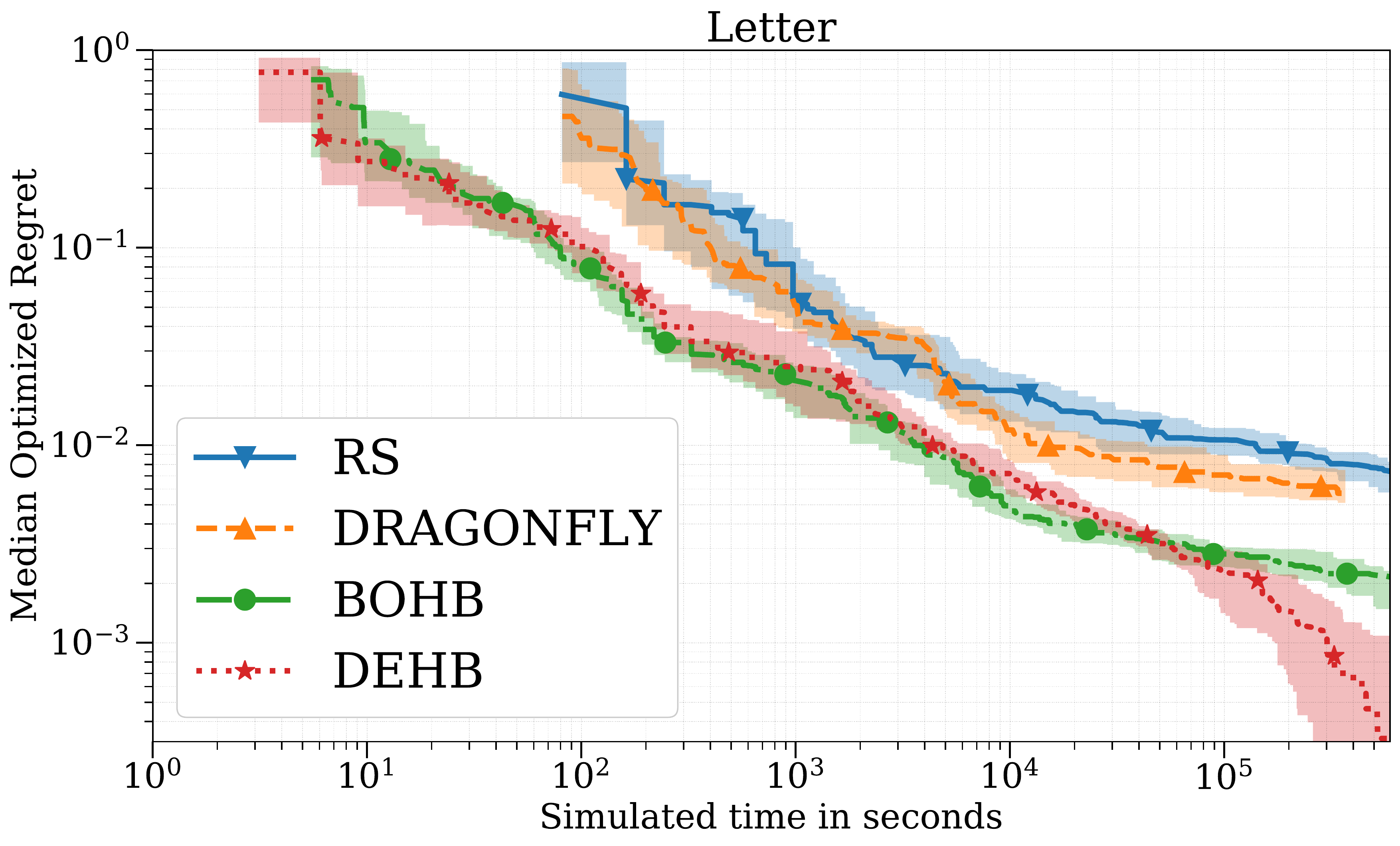} &
    \includegraphics[width=0.45\columnwidth]{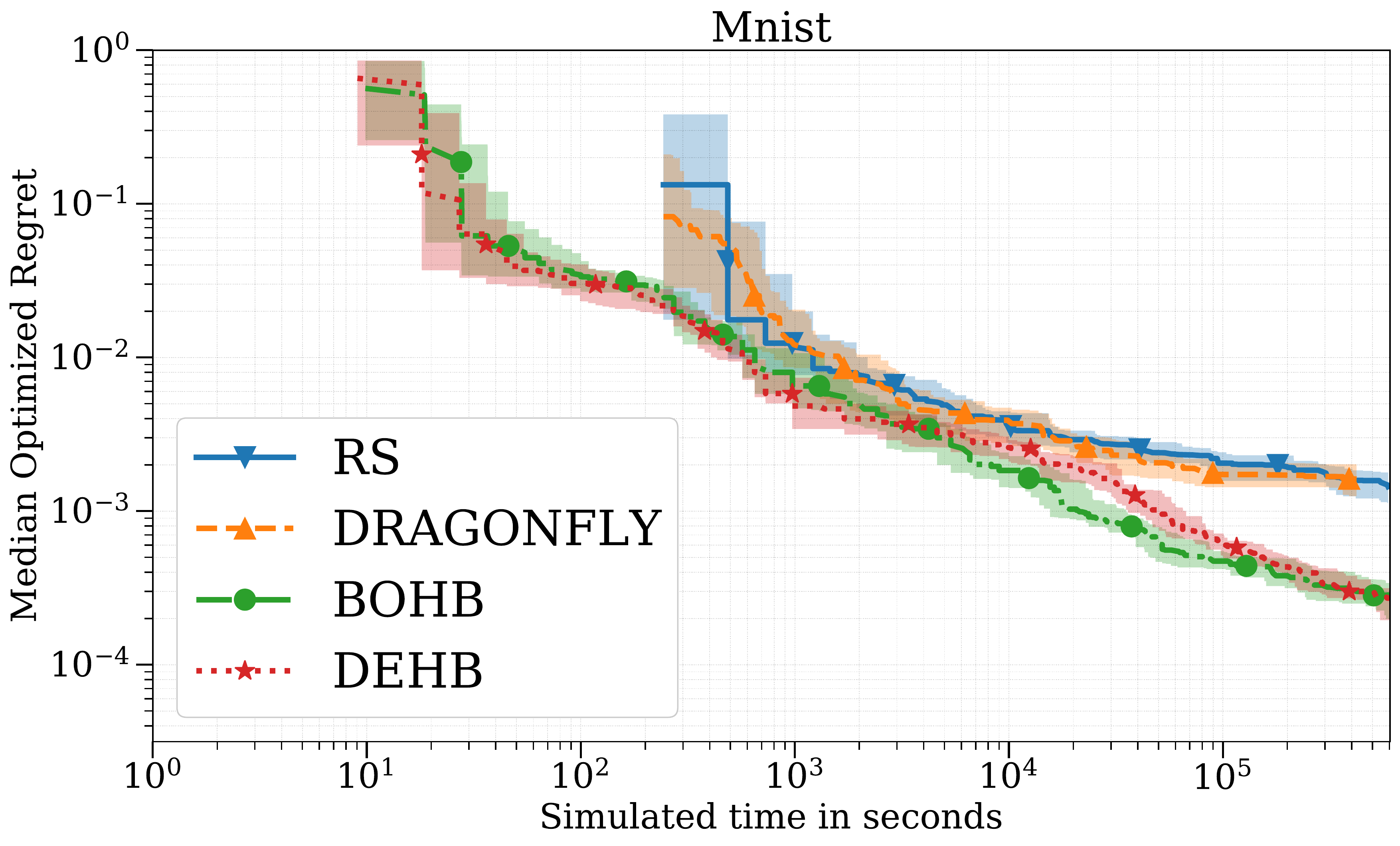} \\
    \includegraphics[width=0.45\columnwidth]{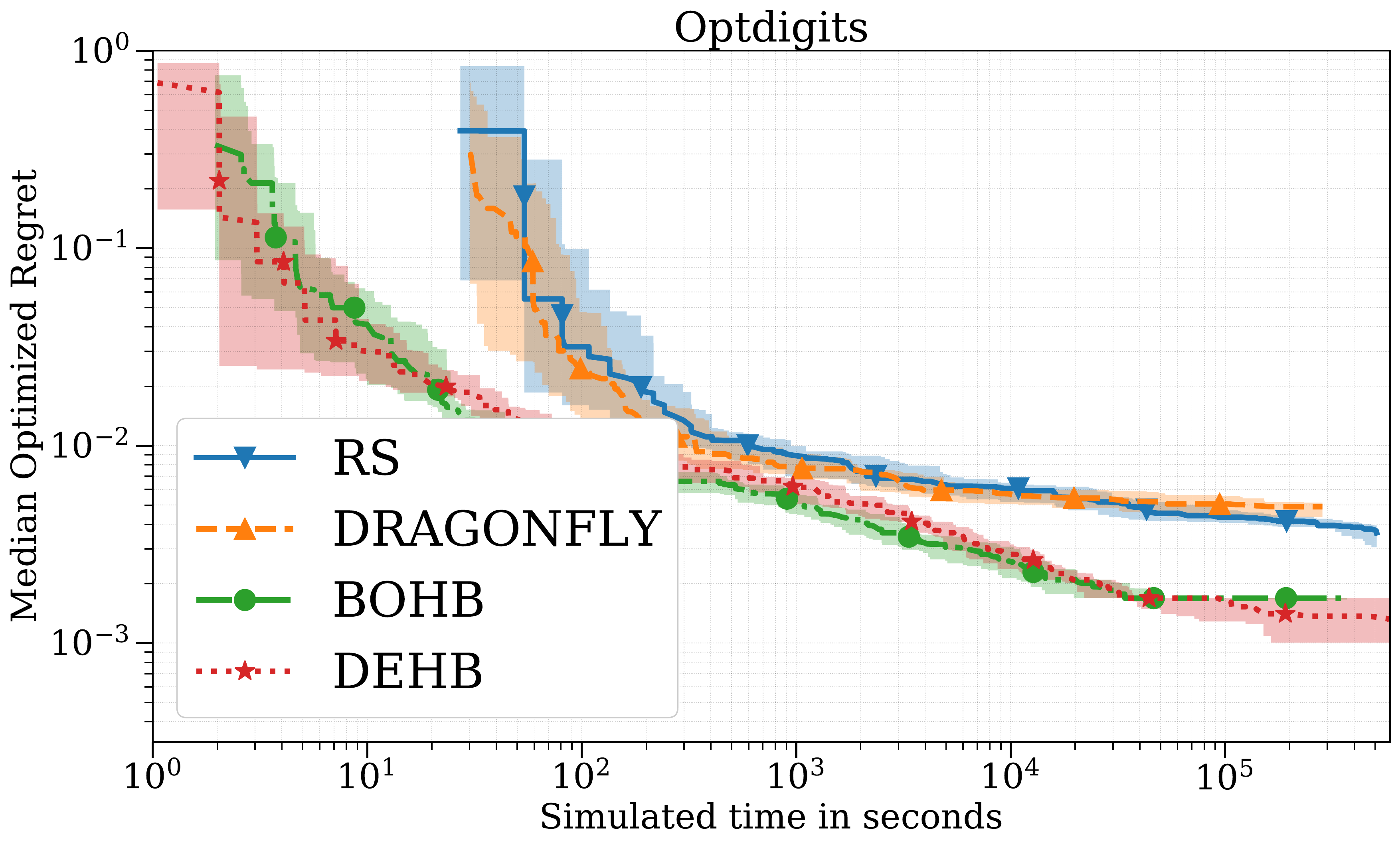} &
    \includegraphics[width=0.45\columnwidth]{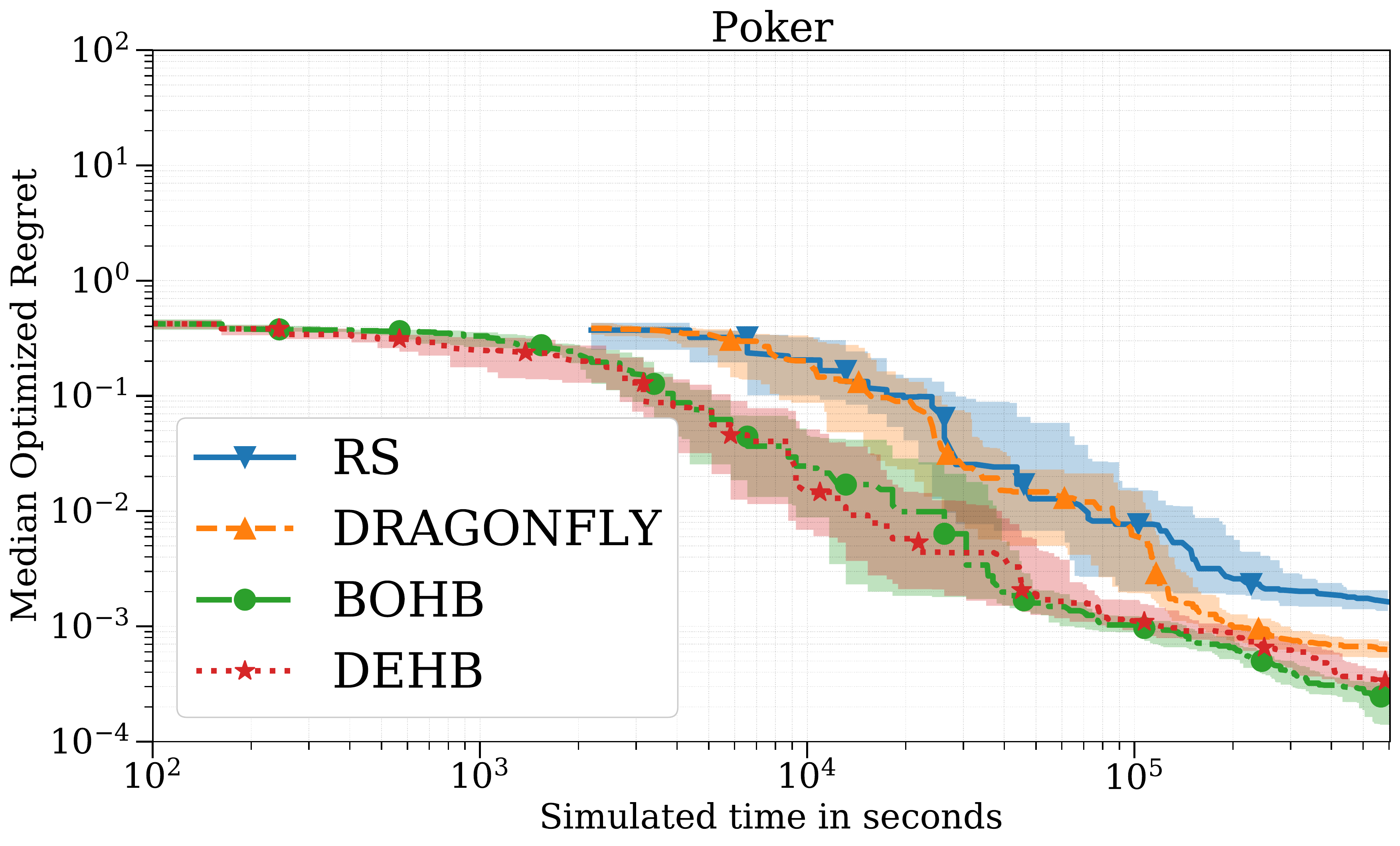} \\
    \includegraphics[width=0.45\columnwidth]{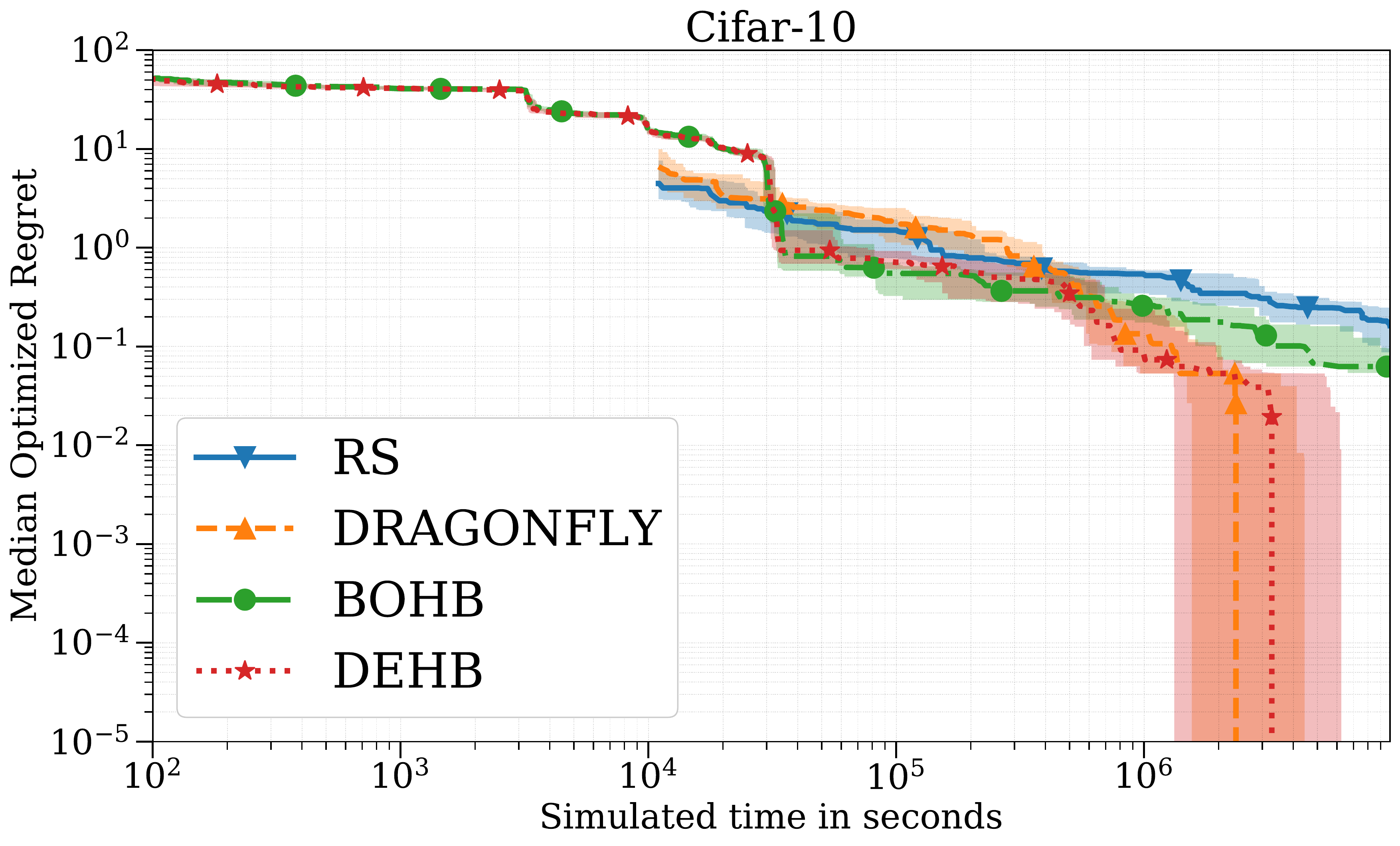} &
    \includegraphics[width=0.45\columnwidth]{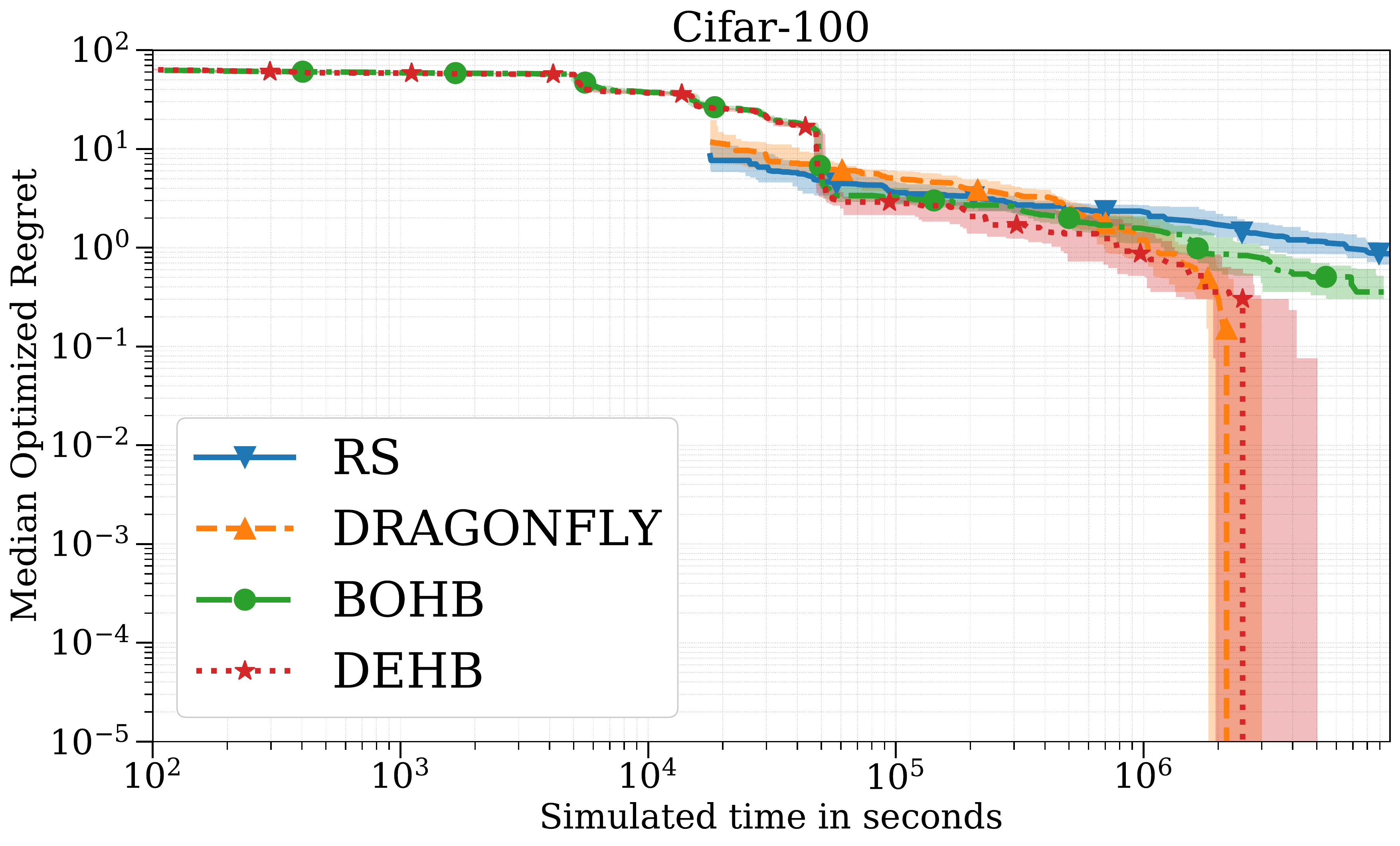} \\
    \includegraphics[width=0.45\columnwidth]{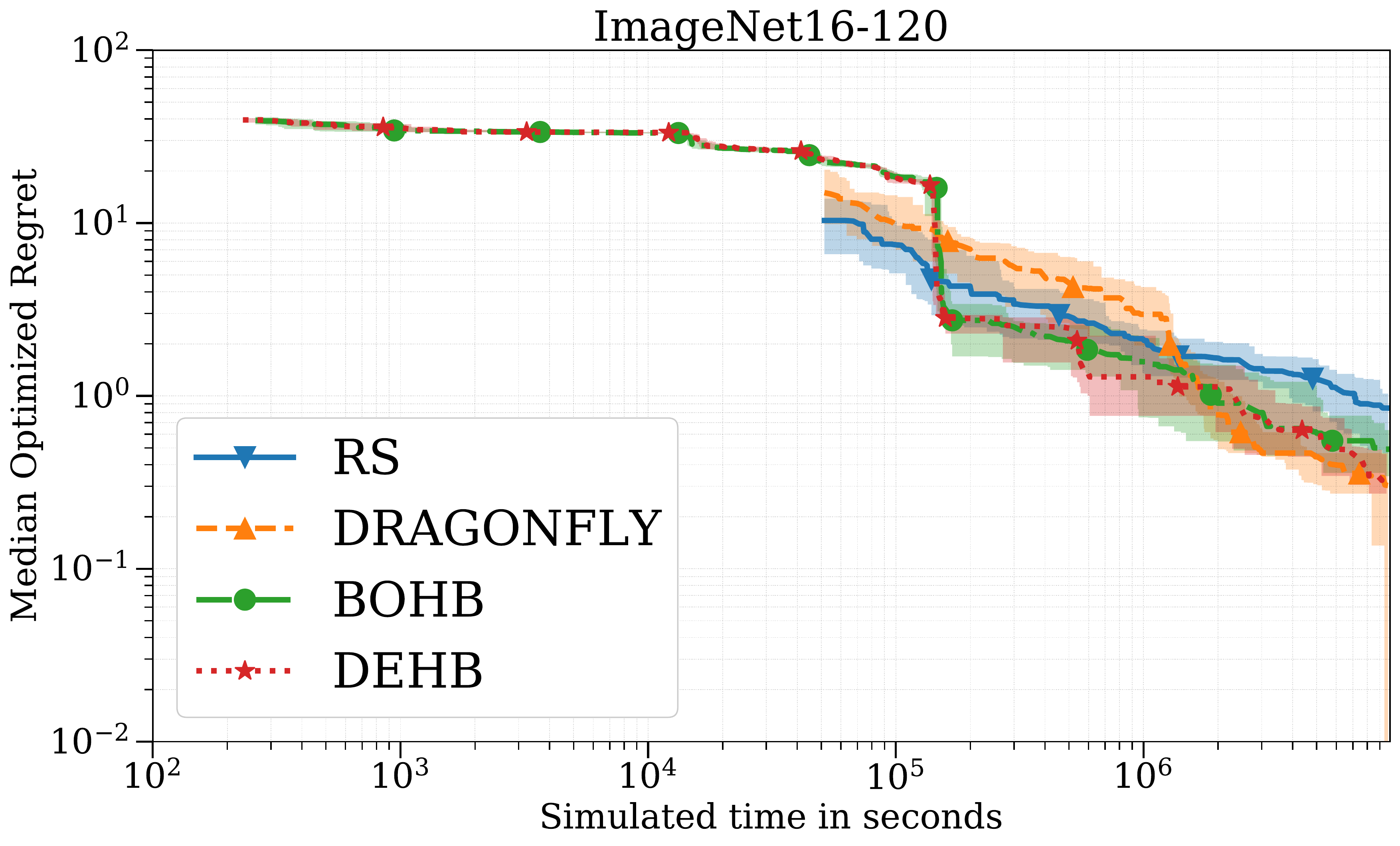} &
    \includegraphics[width=0.45\columnwidth]{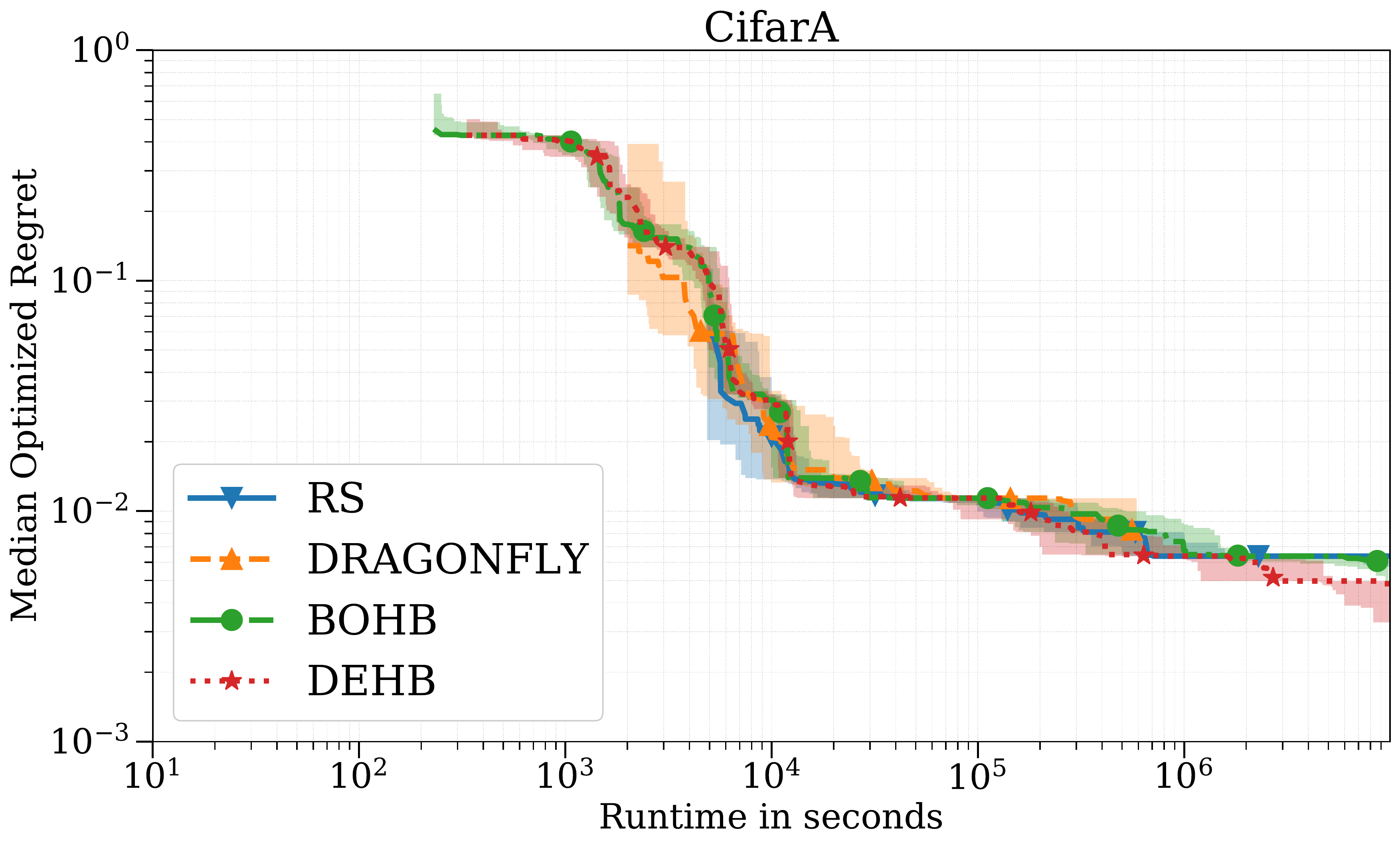} \\
    \includegraphics[width=0.45\columnwidth]{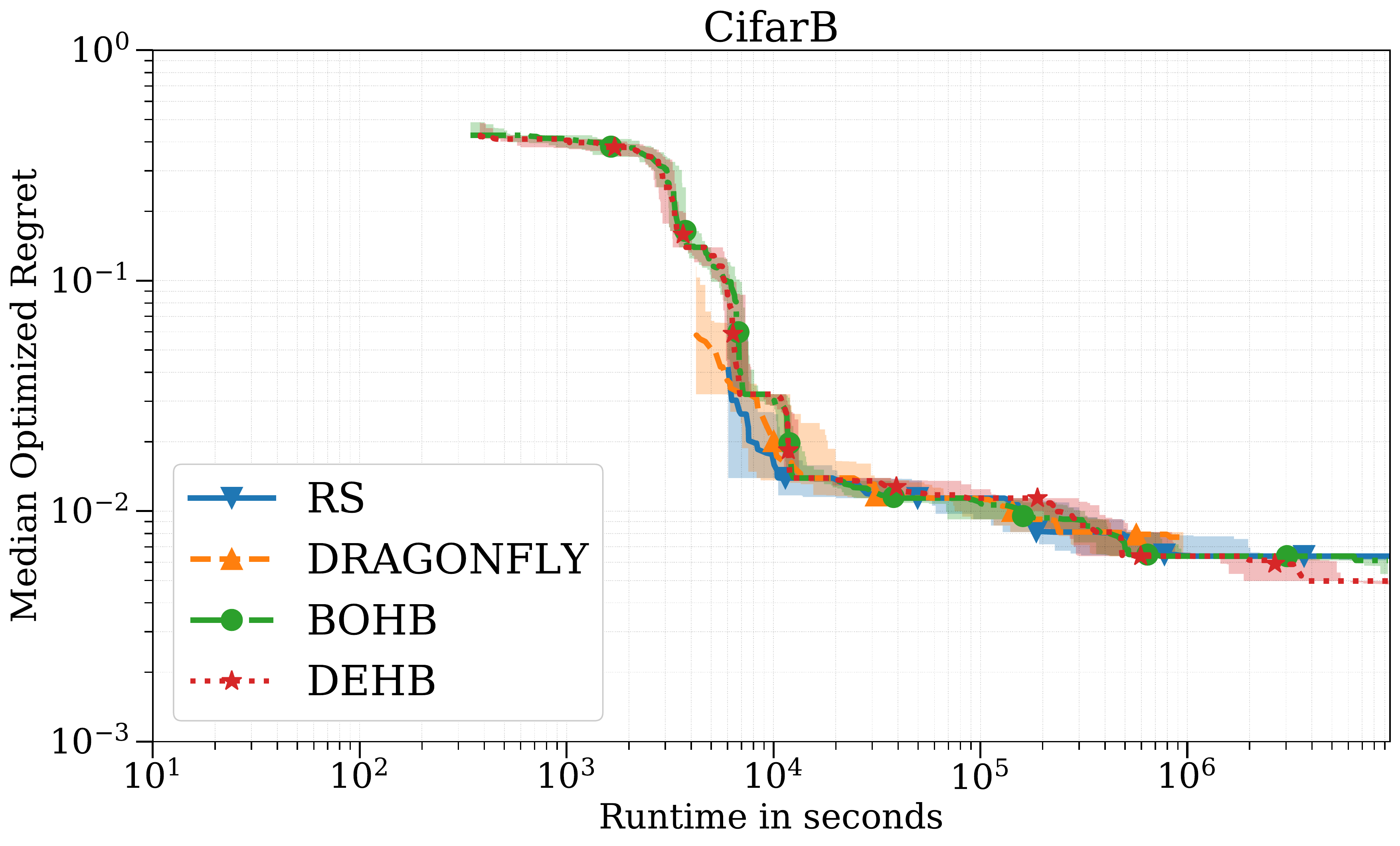} &
    \includegraphics[width=0.45\columnwidth]{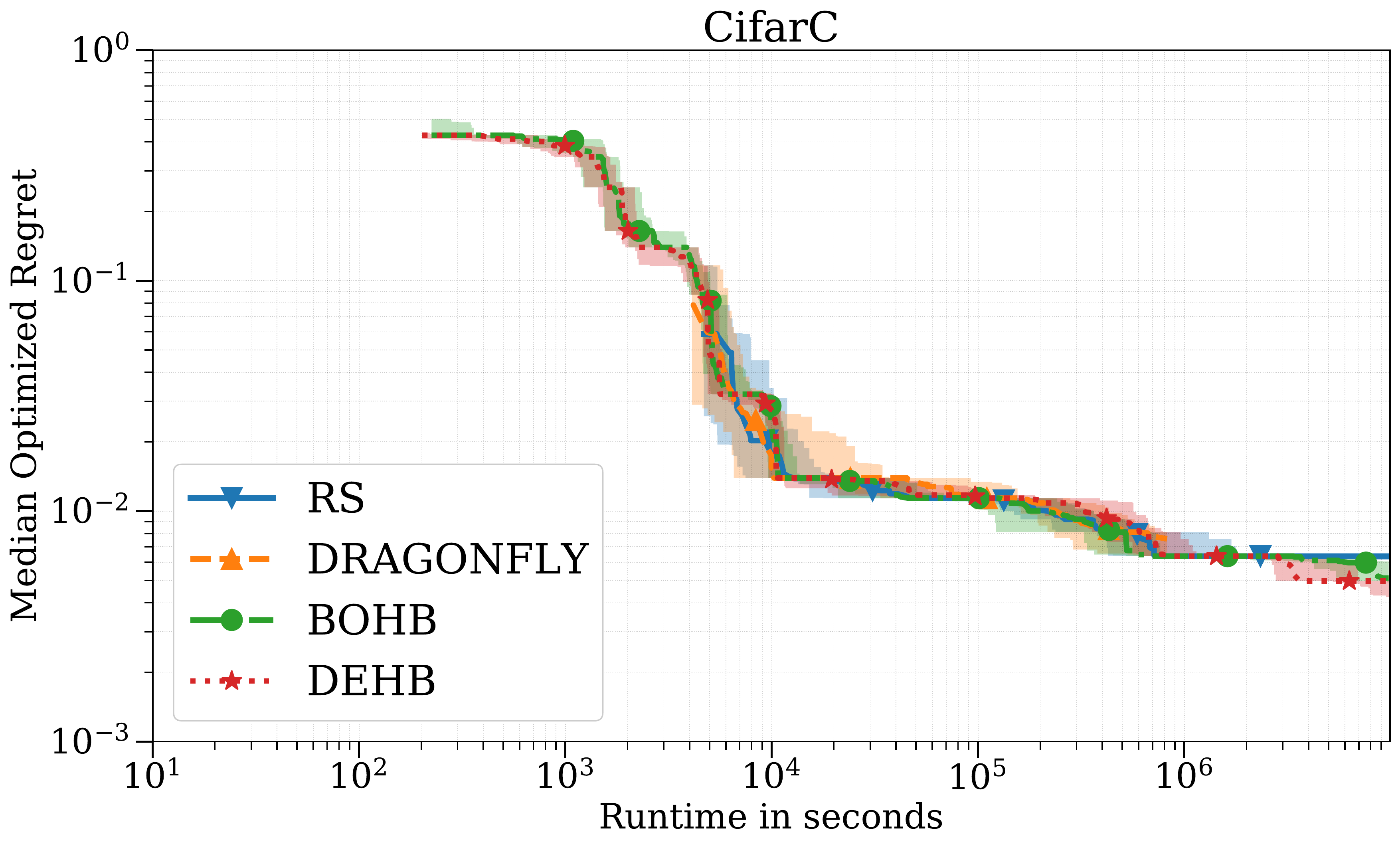} \\
    \includegraphics[width=0.45\columnwidth]{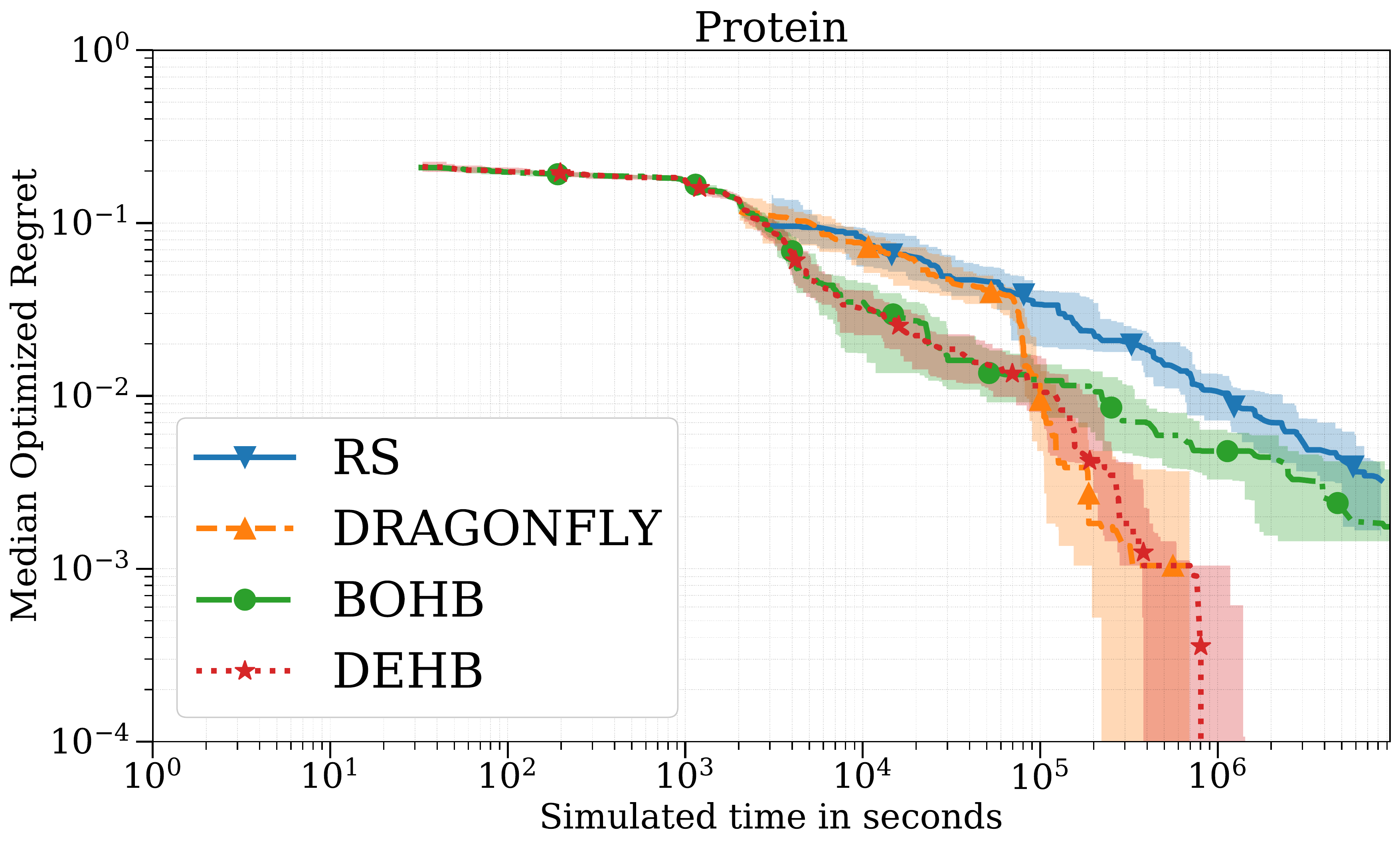} &
    \includegraphics[width=0.45\columnwidth]{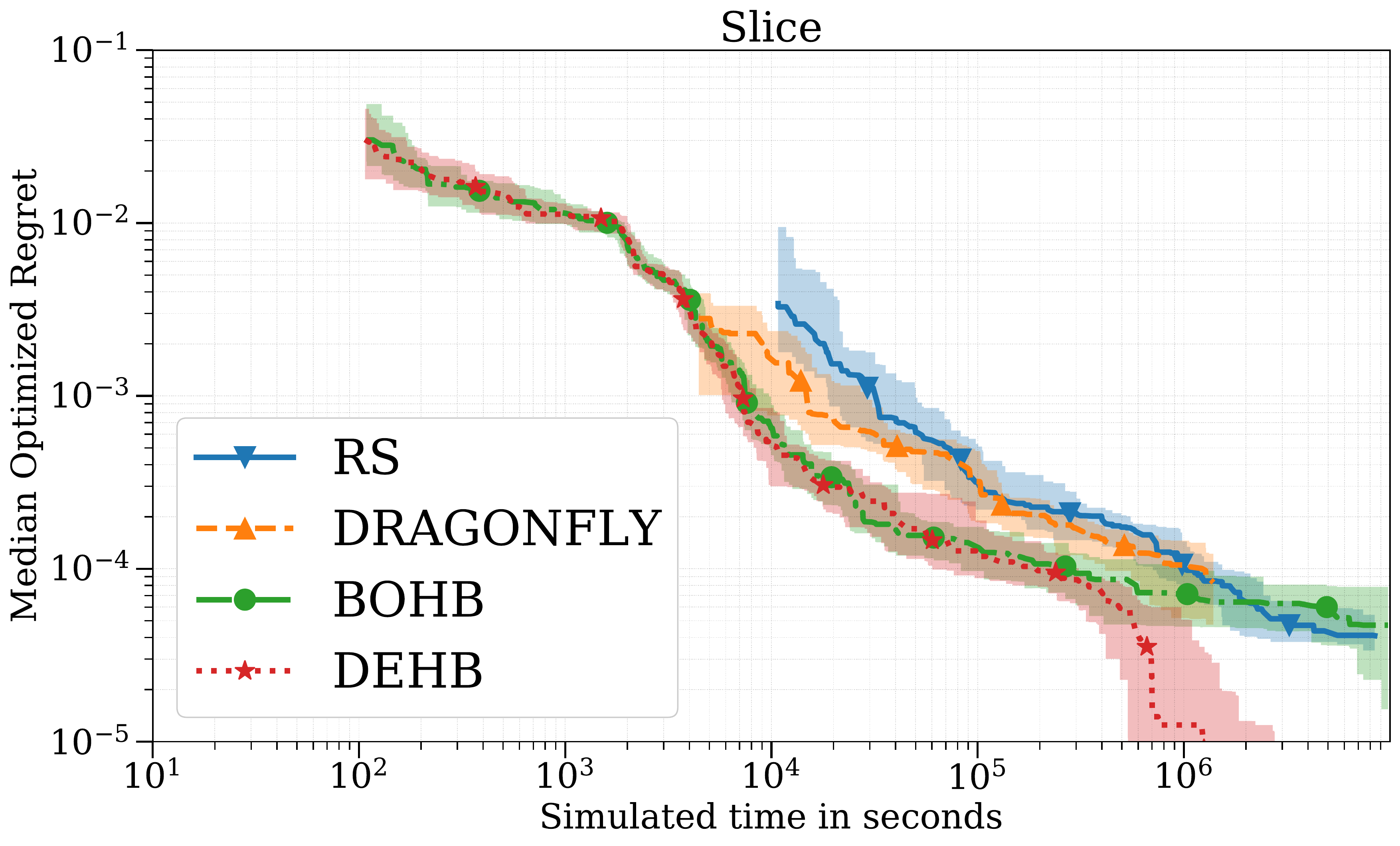} \\
    \includegraphics[width=0.45\columnwidth]{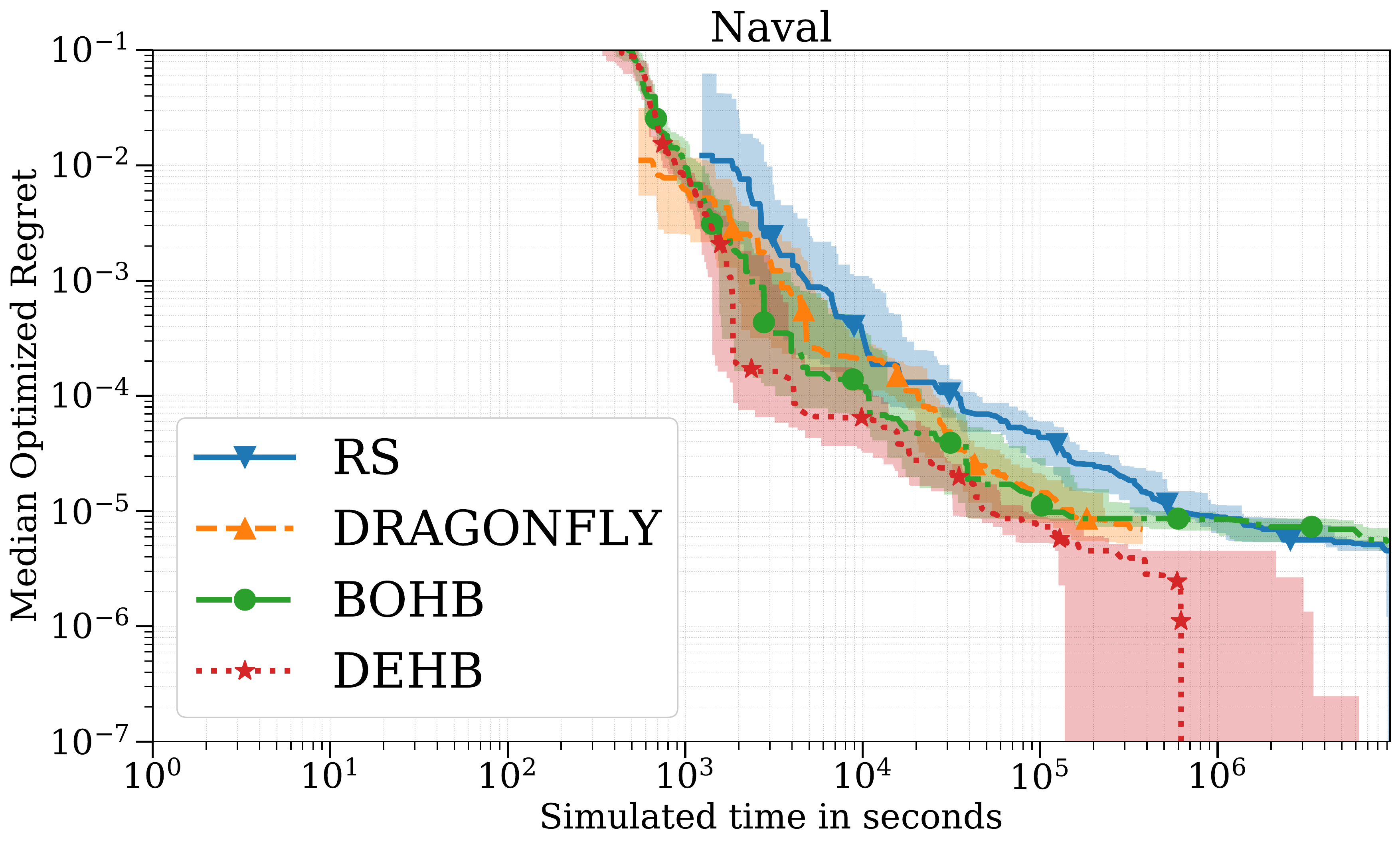} &
    \includegraphics[width=0.45\columnwidth]{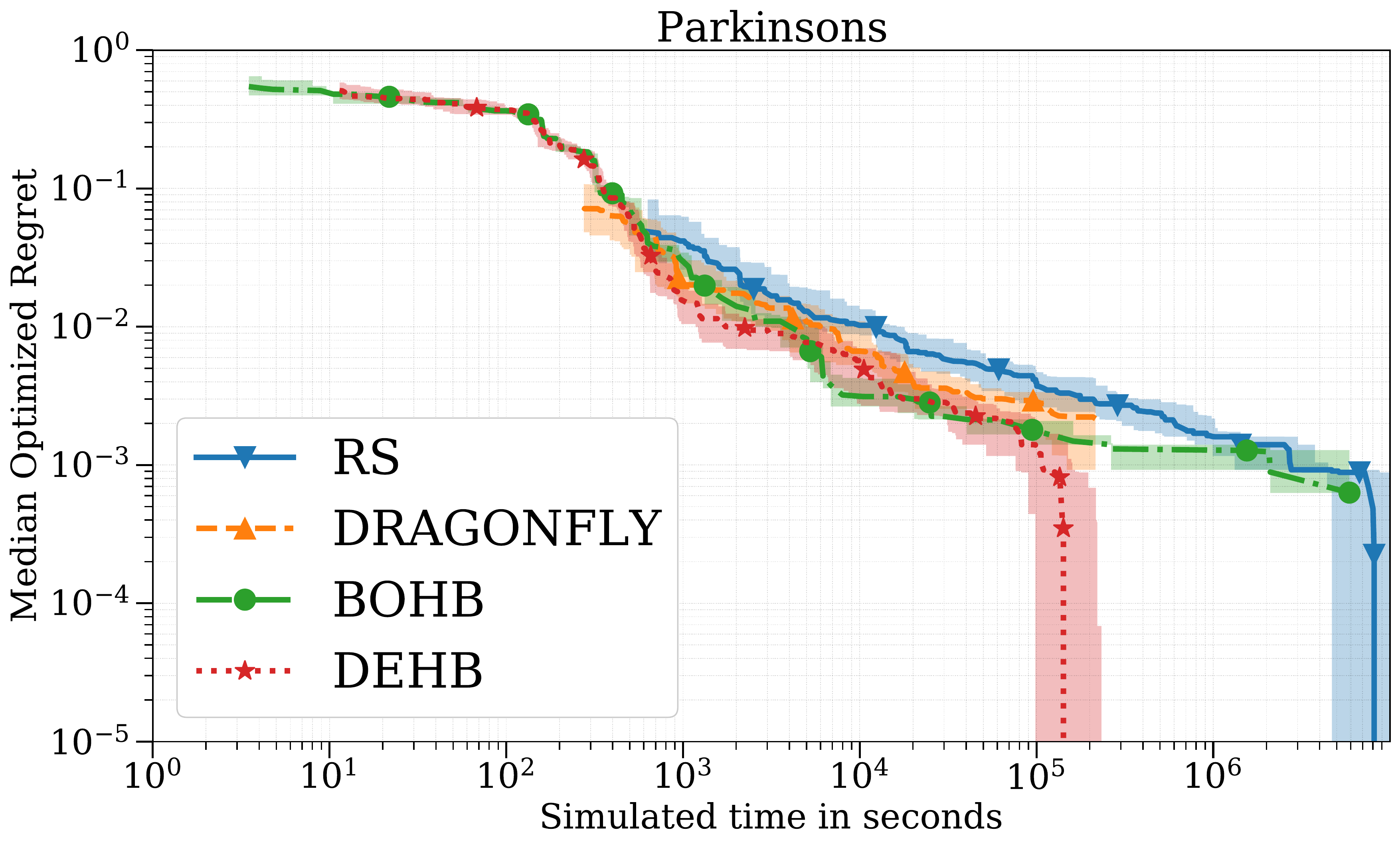} \\
\end{tabular}
\caption{Comparison of GP-based multi-fidelity BO (Dragonfly), KDE based multi-fidelity BO (BOHB), DE based multi-fidelity (DEHB) methods for some of the benchmarks, averaged over $32$ runs of each algorithm.}
\label{fig:bo-dragonfly}
\end{figure}

BOHB~\citep{falkner-icml18a} showed that its KDE based BO outperformed other GP-based BO methods. Hence, BOHB was treated as the primary challenger to DEHB as a robust, general multi-fidelity based HPO solver. In this section we run experiments on the benchmarks detailed in the previous sections, to compare DEHB to another popular multi-fidelity BO optimizer, Dragonfly~\citep{dragonfly2020} (in addition to BOHB). Dragonfly implements BOCA (\cite{kandasamy2017multi}) which performs BO with low-cost approximations of function evaluations on fidelities treated as a continuous domain. However, this GP-based BO method had longer execution time compared to other algorithms for the tabular/surrogate benchmarks. In Figure \ref{fig:bo-dragonfly} we therefore show average of $32$ runs for each algorithm, while having to terminate runs earlier than other algorithms for certain cases. In this experiment, we optimize the median performance of a configuration over different seeds. We observe that Dragonfly shows a high variance in performance across benchmarks whereas DEHB is consistently the best or at worse, comparable to Dragonfly. Moreover, BOHB performs clearly better than Dragonfly in $8$ out of the $16$ cases shown in Figure \ref{fig:bo-dragonfly}, while being comparable to Dragonfly in at least $4$ other benchmarks. Dragonfly comes out as the best optimizer only for the Cifar10 dataset in the NAS-201 benchmark in Figure \ref{fig:bo-dragonfly}. These experiments however, further illustrate the practicality, robustness, and generality of DEHB compared to GP-based multi-fidelity BO methods.

\subsection{Results summary} \label{sec:app-res-summary}
In the previous experiments sections, results on all the benchmarks for DEHB and all other baselines were reported demonstrating the competitive and often superior \textit{anytime} performance of DEHB. In Table \ref{table:summary}, we report the mean final validation regret achieved by all algorithms across all the $26$ benchmarks. DEHB got the \textit{best} performance in nearly $1/3$-rd of the benchmarks while reporting the \textit{second}-best performance in over $1/4$-th of all the benchmarks. The last row of Table \ref{table:summary} shows the rank of each algorithm averaged across their final performances on each benchmark. DEHB clearly is the \textit{best} performing algorithm on the whole, followed by DE, which powers DEHB under the hood. Such rankings illustrate DEHB's robustness across different search spaces, including high dimensions, discrete or mixed type spaces, and even problems where response signals from lower fidelity subspaces may not be too informative. It must be noted that for all the different problems tested for with the collection of benchmarks, DEHB is never consistently outperformed by any multi-fidelity or full-budget algorithm. 

It must be noted that based on the average rank plot in Figure \ref{fig:ranking}, BOHB appears to be better than DEHB in the \textit{early middle} section of the optimization. The underlying model-based search in BOHB can possibly explain this phenomenon. Though DEHB's underlying DE requires more function evaluations to explore the space, it remains competitive with BOHB and the latter is not significantly better across any of the used benchmarks. Moreover, as Figure \ref{fig:ranking} indicates, BOHB's relative performance worsens while DEHB continues to remain better even in comparison to the other full-budget black box optimizers such as DE, RE and TPE. BOHB's model-based search can again be attributed for this phenomenon. Many of the benchmarks used are high-dimensional and have mixed data types, which can affect BOHB's model certainty over the configuration space and require much more observations than DEHB requires. 
Overall, DEHB shows consistently good \textit{anytime} performance with strong \textit{final} performance scores too. As detailed earlier, DEHB's efficiency, simplicity and its speed allow the good use of available resources and make it a good practical and reliable tool for HPO in practice.

\begin{table*}[!ht]
\centering
\setlength{\tabcolsep}{3pt}
\scriptsize 
\begin{tabular}{||c|c c c c c c c c||} 
 \hline
 \  & RS & HB & BOHB & TPE & SMAC & RE & DE & DEHB \\ [0.5ex]  
 \hline\hline
 $Counting$ & 9.8e-2 $\pm$ & 6.1e-2 $\pm$ & 2.1e-2 $\pm$ & 8.1e-3 $\pm$ & \textbf{1.1e-6} & 2.7e-2 $\pm$ & 1.5e-2 $\pm$ & 9.7e-4 $\pm$ \\ 
 $4+4$ & 2.6e-2 & 1.9e-2 & 1.8e-2 & 4.8e-3 & $\pm$ 3.9e-6 & 7.9e-3 & 6.6e-3 & 4.6e-4 ($2$)\\
 \hline
 
 $Counting$ & 1.9e-1 $\pm$ & 1.5e-1 $\pm$ & 3.9e-3 $\pm$ & 7.6e-2 $\pm$ & \textbf{2.1e-3} & 5.0e-2 $\pm$ & 6.6e-2 $\pm$ & 1.4e-2 $\pm$ \\ 
 $8+8$ & 2.5e-2 & 2.5e-2 & 1.1e-3 & 3.1e-2 & $\pm$ 2.3e-3 & 1.2e-2 & 1.5e-2 & 3.7e-3 ($3$)\\
 \hline
 
 $Counting$ & 2.8e-1 $\pm$ & 2.4e-1 $\pm$ & 9.6e-2 $\pm$ & 1.7e-1 $\pm$ & 1.6e-1 $\pm$ & 9.2e-2 $\pm$ & 1.4e-1 $\pm$ & \textbf{6.5e-2} \\ 
 $16+16$ & 2.2e-2 & 2.2e-2 & 8.3e-3 & 2.8e-2 & 1.6e-2 & 1.5e-2 & 1.7e-2 & $\pm$ 6.3e-3 \\
 \hline
 
 $Counting$ & 3.5e-1 $\pm$ & 3.2e-1 $\pm$ & 2.4e-1 $\pm$ & 2.6e-1 $\pm$ & 3.6e-1 $\pm$ & 1.6e-1 $\pm$ & 2.2e-1 $\pm$ & \textbf{1.4e-1} \\ 
 $32+32$ & 1.4e-2 & 1.8e-2 & 1.3e-2 & 1.9e-2 & 2.e-2 & 2.e-2 & 1.8e-2 & $\pm$ 1.1e-2 \\
 \hline
 
 
 $OpenML$ & 3.8e-3 $\pm$ & 3.1e-3 $\pm$ & 1.8e-3 $\pm$ & 1.9e-3 $\pm$ & 2.8e-3 $\pm$ & 2.e-3 $\pm$ & 1.9e-3 $\pm$ & \textbf{1.1e-3} \\ 
 $adult$ & 4.5e-4 & 5.8e-4 & 6.1e-4 & 5.e-4 & 8.3e-4 & 6.3e-4 & 3.4e-4 & $\pm$ 2.0e-4 \\
 \hline
 
 $OpenML$ & 4.1e-3 $\pm$ & 3.6e-3 $\pm$ & 2.5e-3 $\pm$ & 2.3e-3 $\pm$ & 3.e-3 $\pm$ & 2.1e-3 $\pm$ & 2.0e-3 $\pm$ & \textbf{1.8e-3} \\ 
 $higgs$ & 7.3e-4 & 5.5e-4 & 5.8e-4 & 7.4e-4 & 1.3e-3 & 9.4e-4 & 4.6e-4 & $\pm$ 2.1e-4 \\
 \hline
 
 $OpenML$ & 3.8e-3 $\pm$ & 2.9e-3 $\pm$ & 2.e-3 $\pm$ & 7.4e-4 $\pm$ & 1.0e-3 $\pm$ & 6.2e-4 $\pm$ & 8.2e-4 $\pm$ & \textbf{5.5e-4} \\ 
 $letter$ & 1.1e-3 & 7.5e-4 & 1.5e-3 & 2.8e-4 & 9.9e-4 & 3.3e-4 & 1.5e-4 & $\pm$ 3.4e-4 \\
 \hline
 
 $OpenML$ & 1.8e-3 $\pm$ & 1.6e-3 $\pm$ & 1.3e-3 $\pm$ & 9.4e-4 $\pm$ & 1.1e-3 $\pm$ & 9.2e-4 $\pm$ & \textbf{5.3e-5} & 9.5e-4 $\pm$ \\ 
 $mnist$ & 3.1e-4 & 2.0e-4 & 4.3e-4 & 5.2e-5 & 4.6e-4 & 7.1e-5 & $\pm$ 7.2e-5 & 7.6e-5 ($4$)\\
 \hline
 
 $OpenML$ & 3.2e-3 $\pm$ & 2.8e-3 $\pm$ & 1.7e-3 $\pm$ & 1.4e-3 $\pm$ & 1.5e-3 $\pm$ & 1.0e-3 $\pm$ & 8.1e-4 $\pm$ & \textbf{7.9e-4} \\ 
 $optdigits$ & 5.4e-4 & 5.3e-4 & 8.1e-4 & 4.3e-4 & 8.8e-4 & 5.5e-4 & 3.8e-4 & $\pm$ 2.5e-4 \\
 \hline
 
 $OpenML$ & 1.1e-3 $\pm$ & 7.6e-4 $\pm$ & 3.e-4 $\pm$ & 4.4e-4 $\pm$ & 2.1e-4 $\pm$ & 4.e-4 $\pm$ & 4.4e-4 $\pm$ & \textbf{1.9e-4} \\ 
 $poker$ & 3.1e-4 & 1.8e-4 & 1.6e-4 & 1.1e-4 & 1.7e-4 & 1.9e-4 & 1.6e-4 & $\pm$ 5.4e-5 \\
 \hline
 
 $BNN$ & 4.7 $\pm$ & 4.3 $\pm$ & \textbf{3.8} $\pm$ & 4.0 $\pm$ & 4.4 $\pm$ & 6.6 $\pm$ & 5.0 $\pm$ & 4.0 $\pm$ \\ 
 $Boston$ & 9.4e-1 & 7.e-1 & 3.1e-1 & 4.2e-1 & 5.5e-1 & 9.5e+0 & 2.8e+0 & 4.9e-1 ($2$)\\
 \hline
 
 $BNN$ & 4.0 $\pm$ & 3.7 $\pm$ & \textbf{3.3} $\pm$ & 3.4 $\pm$ & 3.3 $\pm$ & 5.1 $\pm$ & 4.9 $\pm$ & 3.5 $\pm$ \\ 
 $Protein$ & 9.6e-1 & 5.2e-1 & 1.8e-1 & 2.7e-1 & 2.2e-1 & 5.3e+0 & 2.6e+0 & 3.7e-1 ($4$)\\
 \hline
 
 $Cartpole$ &4.7e+2 $\pm$ & 3.9e+2 $\pm$ & 1.9e+2 $\pm$ & 2.9e+2 $\pm$ & \textbf{1.8e+2} & 3.8e+2 $\pm$ & 4.9e+2 $\pm$ & 2.e+2 $\pm$ \\ 
 $(RL)$ & 1.4e+2 & 1.1e+2 & 4.4e+1 & 6.7e+1 & $\pm$ 1.9e+1 & 1.3e+2 & 1.1e+2 & 7.3e+1 ($3$)\\
 \hline
 
 $NAS 101$ & 2.9e-3 $\pm$ & 3.e-3 $\pm$ & 2.6e-3 $\pm$ & 2.8e-3 $\pm$ & 4.0e-3 $\pm$ & 2.3e-3 $\pm$ & \textbf{1.2e-3} & 2.2e-3 $\pm$ \\ 
 $CifarA$ & 7.6e-4 & 6.0e-4 & 1.2e-3 & 1.1e-3 & 1.5e-3 & 1.2e-3 & $\pm$ 1.1e-3 & 1.4e-3 ($2$)\\
 \hline
 
 $NAS 101$ & 3.1e-3 $\pm$ & 3.2e-3 $\pm$ & 2.8e-3 $\pm$ & 3.e-3 $\pm$ & 2.9e-3 $\pm$ & \textbf{2.4e-3} & 2.6e-3 $\pm$ & 2.6e-3 $\pm$ \\ 
 $CifarB$ & 5.3e-4 & 3.8e-4 & 7.4e-4 & 5.8e-4 & 1.4e-3 & $\pm$ 1.0e-3 & 6.9e-4 & 1.1e-3 ($2$)\\
 \hline
 
 $NAS 101$ & 3.2e-3 $\pm$ & 3.1e-3 $\pm$ & 2.6e-3 $\pm$ & 2.7e-3 $\pm$ & 6.4e-3 $\pm$ & 2.3e-3 $\pm$ & \textbf{1.7e-3} & 2.0e-3 $\pm$ \\ 
 $CifarC$ & 3.5e-4 & 5.4e-4 & 7.6e-4 & 8.3e-4 & 1.3e-3 & 1.4e-3 & $\pm$ 1.1e-3 & 1.2e-3 ($2$)\\
 \hline
 
 $NAS 1s1$ & 1.6e-3 $\pm$ & 1.5e-3 $\pm$ & 1.7e-3 $\pm$ & 1.3e-3 $\pm$ & 2.7e-3 $\pm$ & 1.1e-3 $\pm$ & \textbf{9.4e-4} & 1.4e-3 $\pm$ \\
 $SS1$ & 8.6e-4 & 9.6e-4 & 1.1e-3 & 1.1e-3 & 9.9e-4 & 1.1e-3 & $\pm$ 9.1e-4 & 7.8e-4 ($4$)\\
 \hline

 $NAS 1s1$ & 1.3e-3 $\pm$ & 9.8e-4 $\pm$ & 8.6e-4 $\pm$ & 8.2e-4 $\pm$ & 7.2e-4 $\pm$ & 3.e-4 $\pm$ & \textbf{2.3e-4} & 6.4e-4 $\pm$ \\
 $SS2$ & 6.4e-4 & 4.8e-4 & 5.0e-4 & 7.e-4 & 4.3e-4 & 3.e-4 & $\pm$ 2.5e-4 & 5.5e-4 ($3$)\\
 \hline

 $NAS 1s1$ & 3.5e-3 $\pm$ & 3.4e-3 $\pm$ & 3.9e-3 $\pm$ & 3.5e-3 $\pm$ & 3.8e-3 $\pm$ & 2.8e-3 $\pm$ & \textbf{2.3e-3} & 2.6e-3 $\pm$ \\
 $SS3$ & 9.3e-4 & 9.2e-4 & 7.2e-4 & 8.9e-4 & 8.6e-4 & 1.3e-3 & $\pm$ 1.0e-3 & 1.1e-3 ($2$)\\
 \hline
 
 $NAS 201$ & 2.7e-3 $\pm$ & 2.3e-3 $\pm$ & 2.0e-3 $\pm$ & 7.2e-4 $\pm$ & 4.1e-4 $\pm$ & 1.0e-4 $\pm$ & 2.3e-4 $\pm$ & \textbf{7.8e-5} \\
 $Cifar10$ & 1.1e-3 & 7.6e-4 & 1.4e-3 & 1.3e-3 & 5.8e-4 & 5.6e-4 & 1.2e-3 & $\pm$ 1.7e-4 \\
 \hline
 
 $NAS 201$ & 8.1e-3 $\pm$ & 6.1e-3 $\pm$ & 5.7e-3 $\pm$ & 1.9e-3 $\pm$ & 1.3e-3 $\pm$ & 8.e-5 $\pm$ & \textbf{0.e+0} & 1.3e-4 $\pm$ \\
 $Cifar100$ & 3.5e-3 & 3.2e-3 & 4.0e-3 & 2.8e-3 & 2.3e-3 & 2.4e-4 & $\pm$ 0.e+0 & 2.9e-4 ($3$)\\
 \hline
 
 $NAS 201$ & 9.3e-3 $\pm$ & 7.9e-3 $\pm$ & 7.3e-3 $\pm$ & 4.8e-3 $\pm$ & 5.4e-3 $\pm$ & \textbf{2.0e-3} & 2.3e-3 $\pm$ & 2.2e-3 $\pm$ \\
 $ImageNet$ & 3.6e-3 & 3.9e-3 & 4.1e-3 & 3.7e-3 & 3.6e-3 & $\pm$ 1.4e-3 & 8.8e-4 & 1.6e-3 ($2$)\\
 \hline
 
 $NASHPO$ & 7.4e-3 $\pm$ & 4.2e-3 $\pm$ & 2.9e-4 $\pm$ & -4.7e-4 $\pm$ & 3.9e-4 $\pm$ & -1.1e-3 $\pm$ & \textbf{-1.1e-3} & -1.0e-3 $\pm$ \\
 $Protein$ & 4.5e-3 & 2.7e-3 & 1.1e-3 & 1.9e-3 & 2.5e-3 & 3.5e-4 & $\pm$ 3.1e-4 & 5.9e-4 ($3$)\\
 \hline
 
 $NASHPO$ & 2.8e-5 $\pm$ & 2.9e-6 $\pm$ & -1.0e-5 $\pm$ & -3.1e-5 $\pm$ & -1.9e-5 $\pm$ & -3.5e-5 $\pm$ & \textbf{-4.3e-5} & -2.3e-5 $\pm$ \\
 $Slice$ & 3.6e-5 & 2.2e-5 & 2.7e-5 & 1.6e-5 & 1.9e-5 & 1.2e-5 & $\pm$ 7.8e-6 & 1.6e-5 ($4$) \\
 \hline
 
 $NASHPO$ & 6.8e-6 $\pm$ & 2.5e-6 $\pm$ & -2.5e-6 $\pm$ & -4.8e-6 $\pm$ & -6.e-6 $\pm$ & -6.7e-6 $\pm$ & \textbf{-7.e-6} & -6.e-6 $\pm$ \\ 
 $Naval$ & 6.2e-6 & 4.3e-6 & 2.8e-6 & 3.3e-6 & 2.6e-6 & 1.0e-6 & $\pm$ 8.6e-7 & 2.3e-6 ($4$)\\
 \hline
 
 $NASHPO$ & -6.6e-4 $\pm$ & -1.0e-3 $\pm$ & \textbf{-3.4e-3} & -2.3e-3 $\pm$ & -2.6e-3 $\pm$ & -2.9e-3 $\pm$ & -3.2e-3 $\pm$ & -2.4e-3 $\pm$ \\ 
 $Parkinsons$ & 1.1e-3 & 1.0e-3 & $\pm$ 7.4e-4 & 1.1e-3 & 8.4e-4 & 7.5e-4 & 6.8e-4 & 8.8e-4 ($5$) \\
 \hline
 \hline
 $Avg.\ rank$ & \textbf{7.46} & \textbf{6.54} & \textbf{4.42} & \textbf{4.35} & \textbf{4.73} & \textbf{3.16} & \textbf{2.96} & \textbf{2.39} \\ 
 \hline
 \hline
\end{tabular}
\vspace*{0.2cm}
\caption{Final mean validation regret $\pm$ standard deviation for $50$ runs all algorithms tested for all benchmarks. Performance scores for DEHB is accompanied with its (rank) among other algorithms. The last row shows the \textit{average relative rank} of each algorithm based on their final performance on each benchmark.}
\label{table:summary}
\end{table*}

\section{Ablation Studies}\label{sec:ablation}
DEHB was designed as an easy-to-use tool for HPO and NAS. This necessitated that DEHB contains as few hyperparameters as possible that require tuning or that which makes DEHB sensitive to them. Given that the HB parameters inherent to DEHB are contingent on the problem being solved, that leaves only the DE components' hyperparameters to be set adequately. We perform ablation of the mutation and crossover rates to observe how it fairs for DEHB's design for the suite of benchmarks we experiment on.

\subsection{Varying F}

\begin{figure*}[!ht]
\centering
\begin{tabular}{lll}
    \includegraphics[width=0.22\paperwidth]{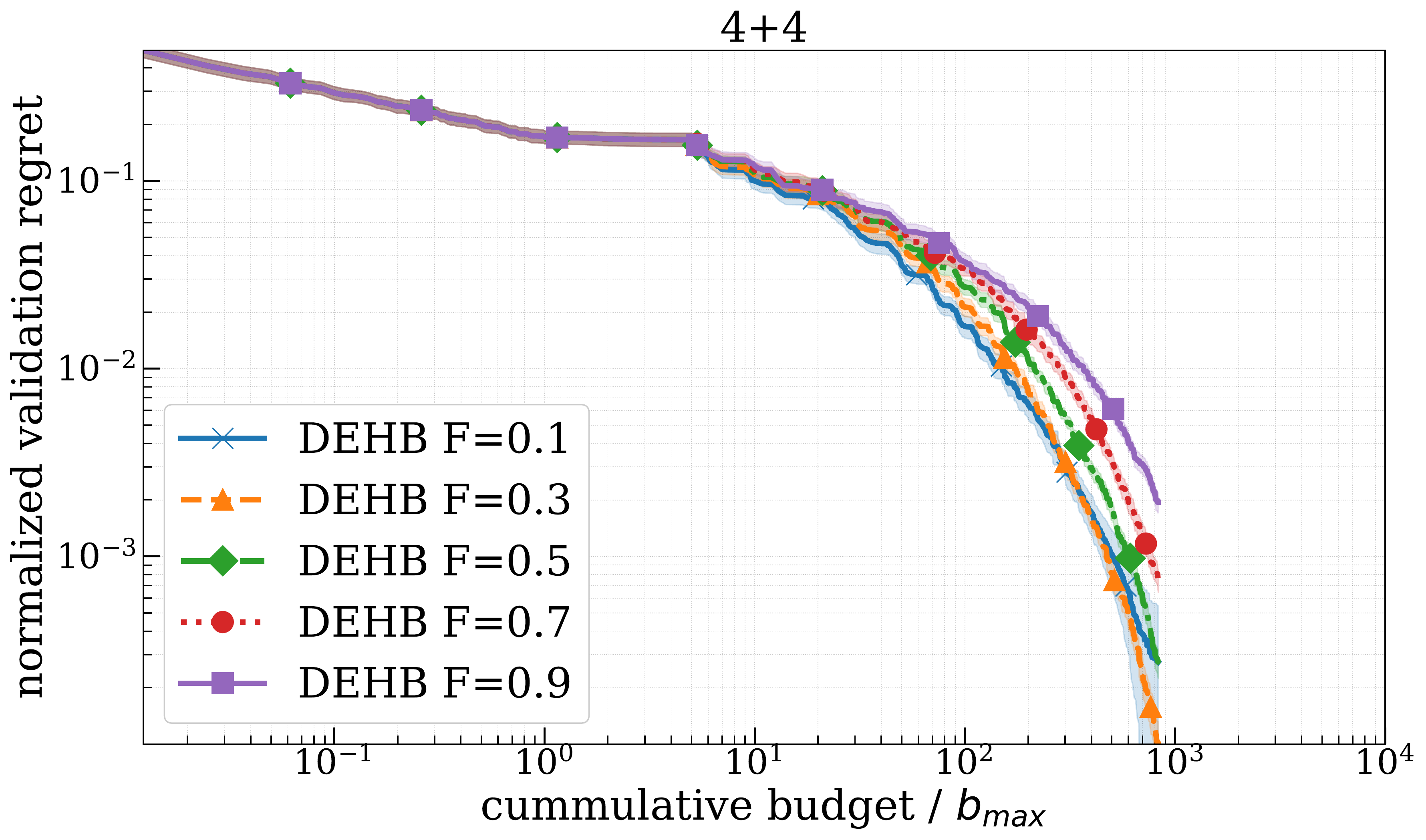} &
    \includegraphics[width=0.22\paperwidth]{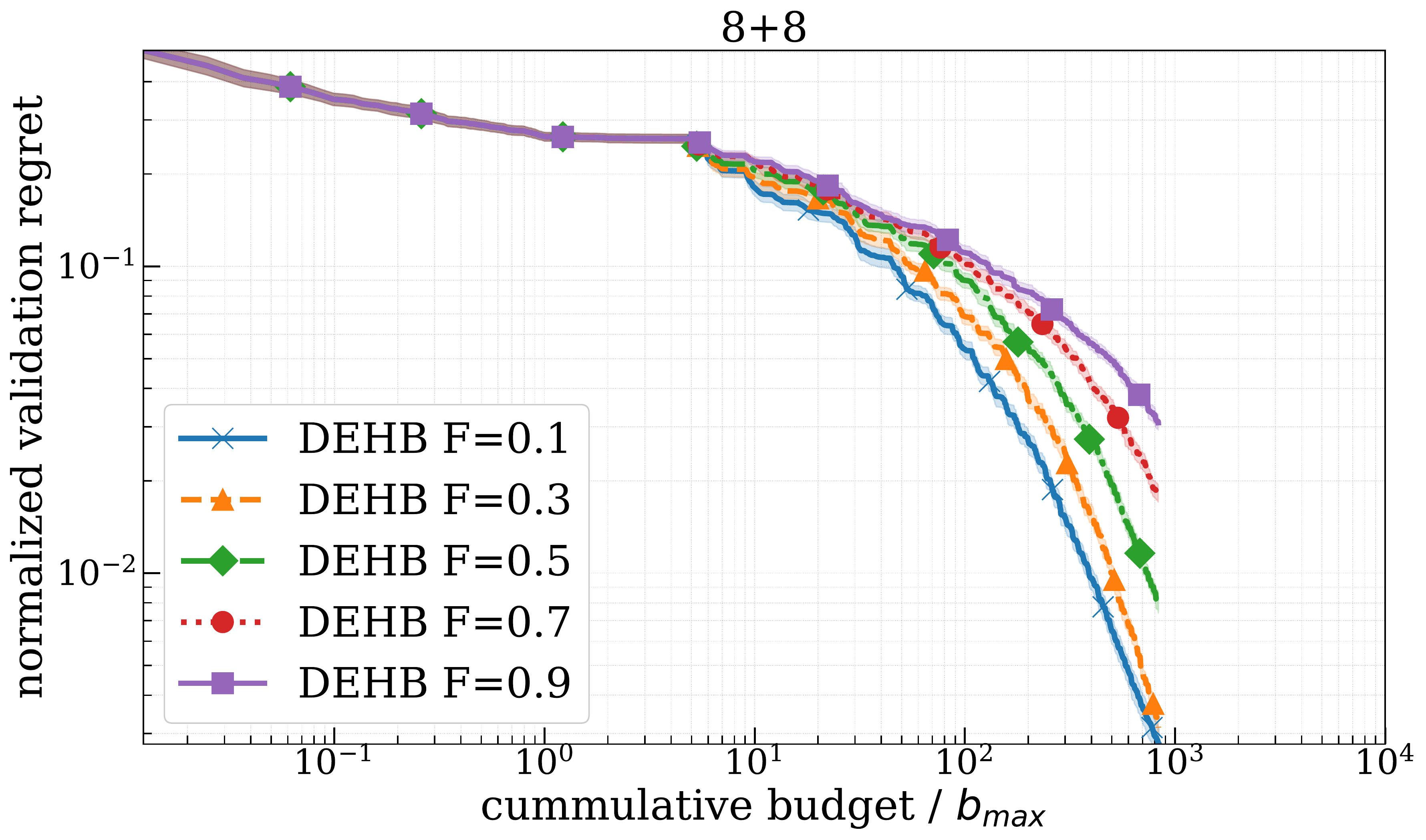} &
    \includegraphics[width=0.22\paperwidth]{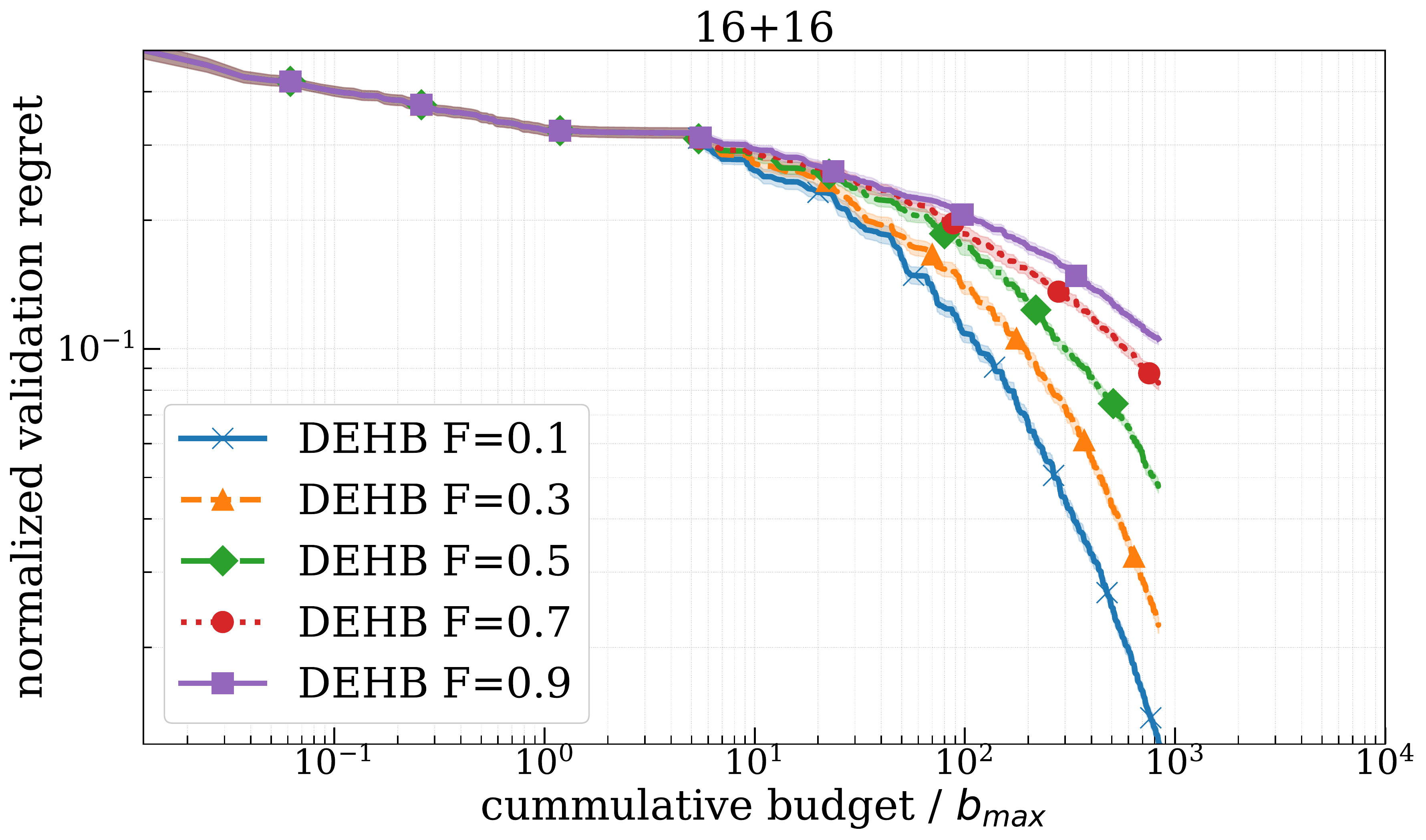} \\
    \includegraphics[width=0.22\paperwidth]{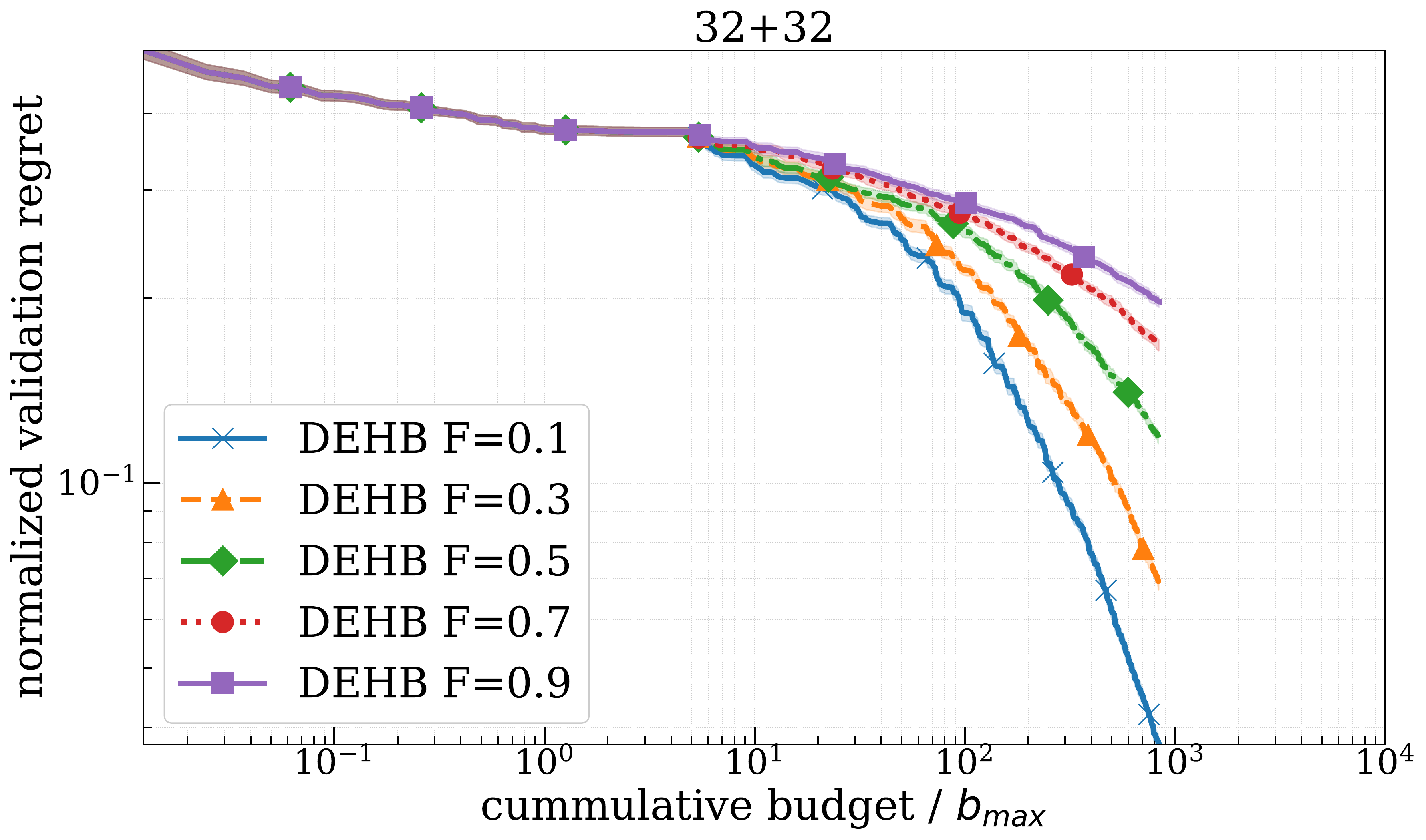} &
    \includegraphics[width=0.22\paperwidth]{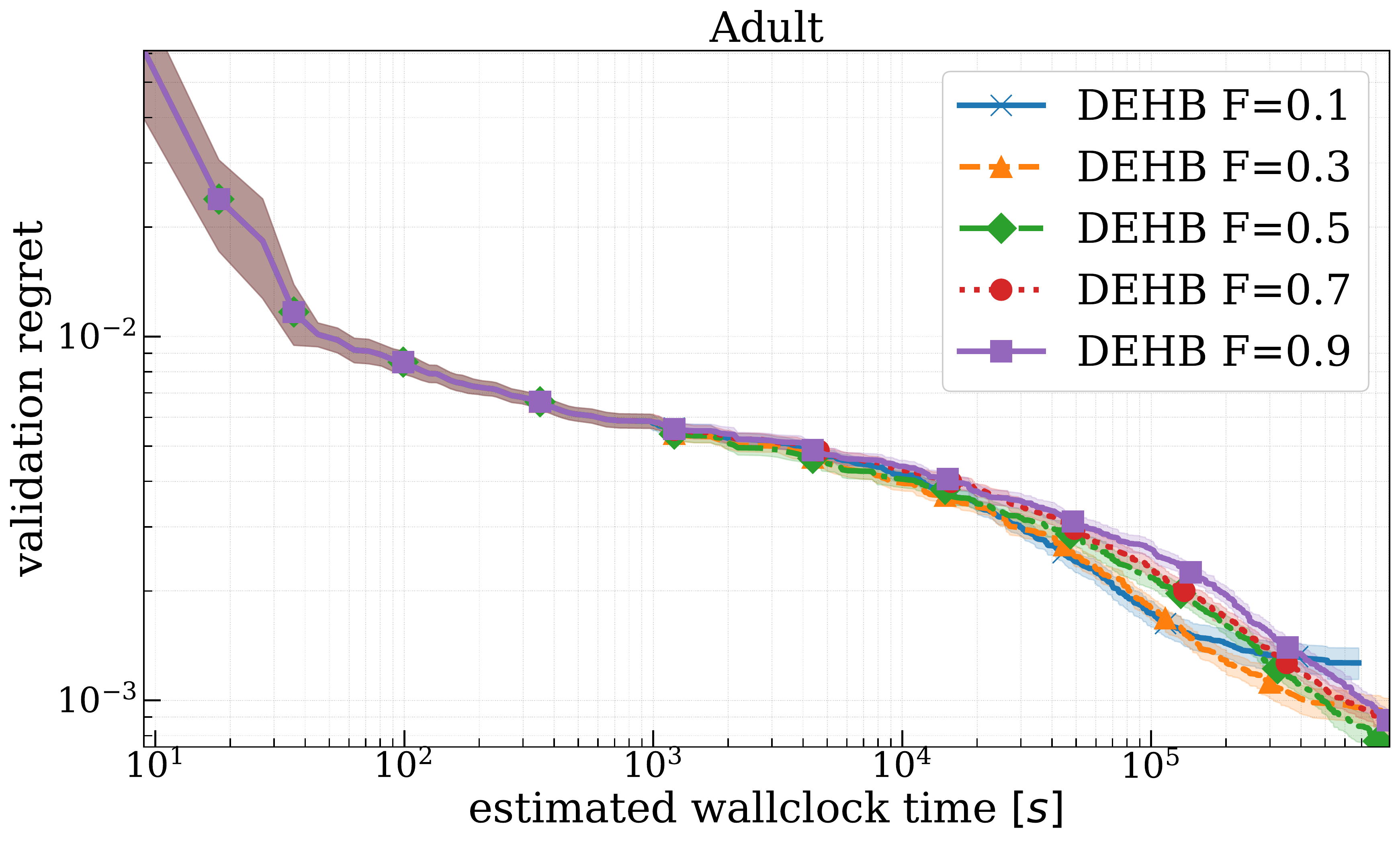} &
    \includegraphics[width=0.22\paperwidth]{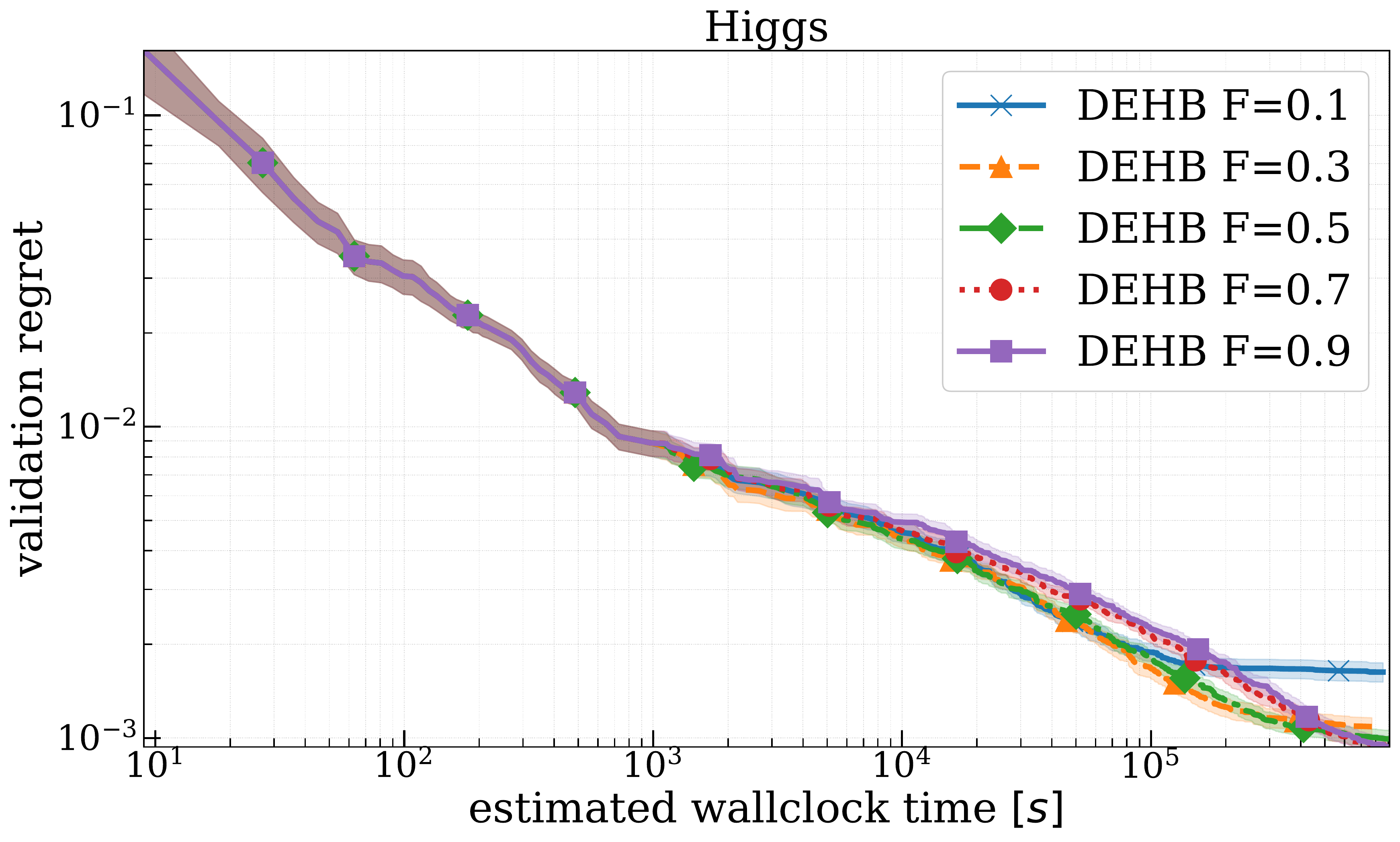} \\
    \includegraphics[width=0.22\paperwidth]{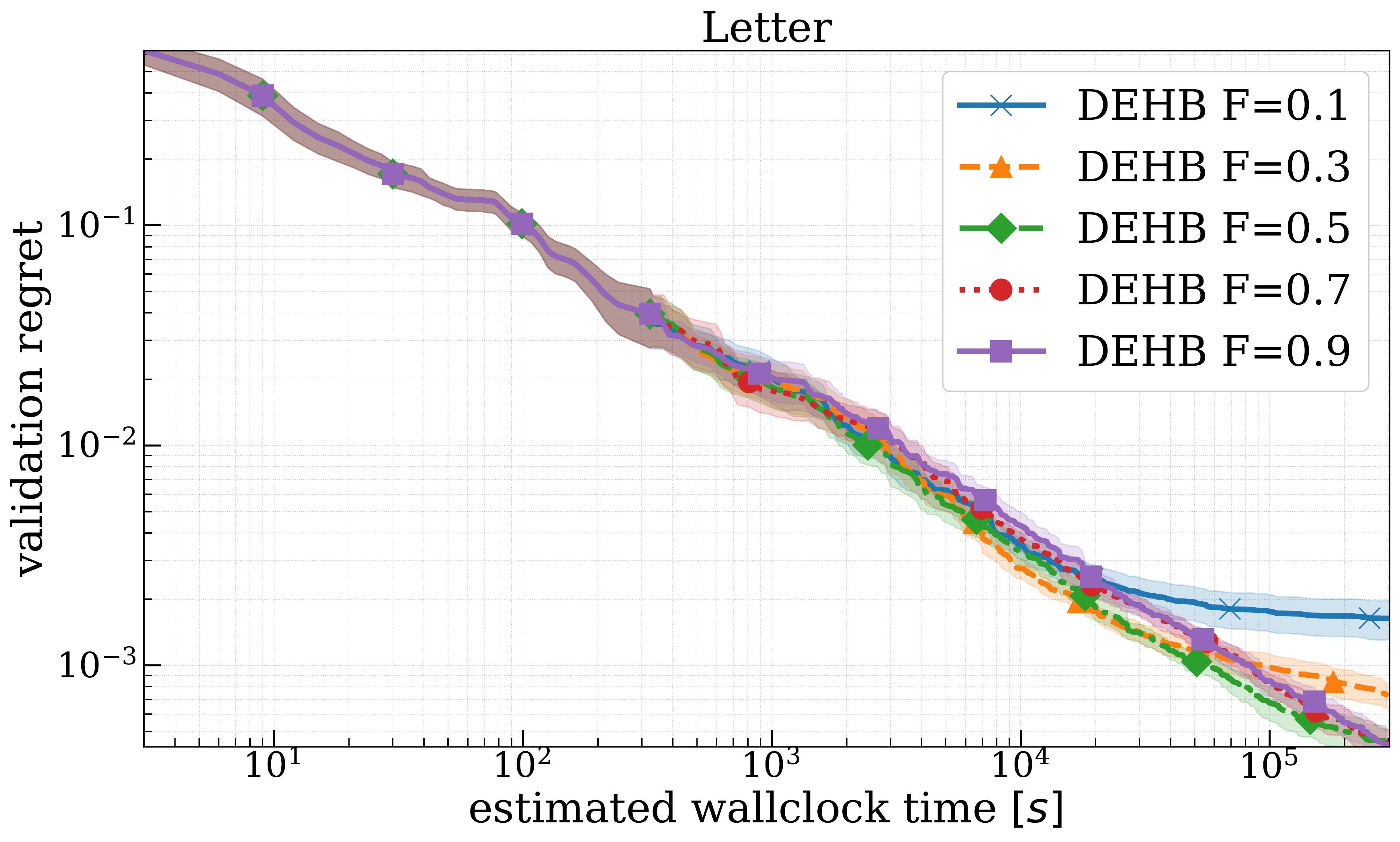} &
    \includegraphics[width=0.22\paperwidth]{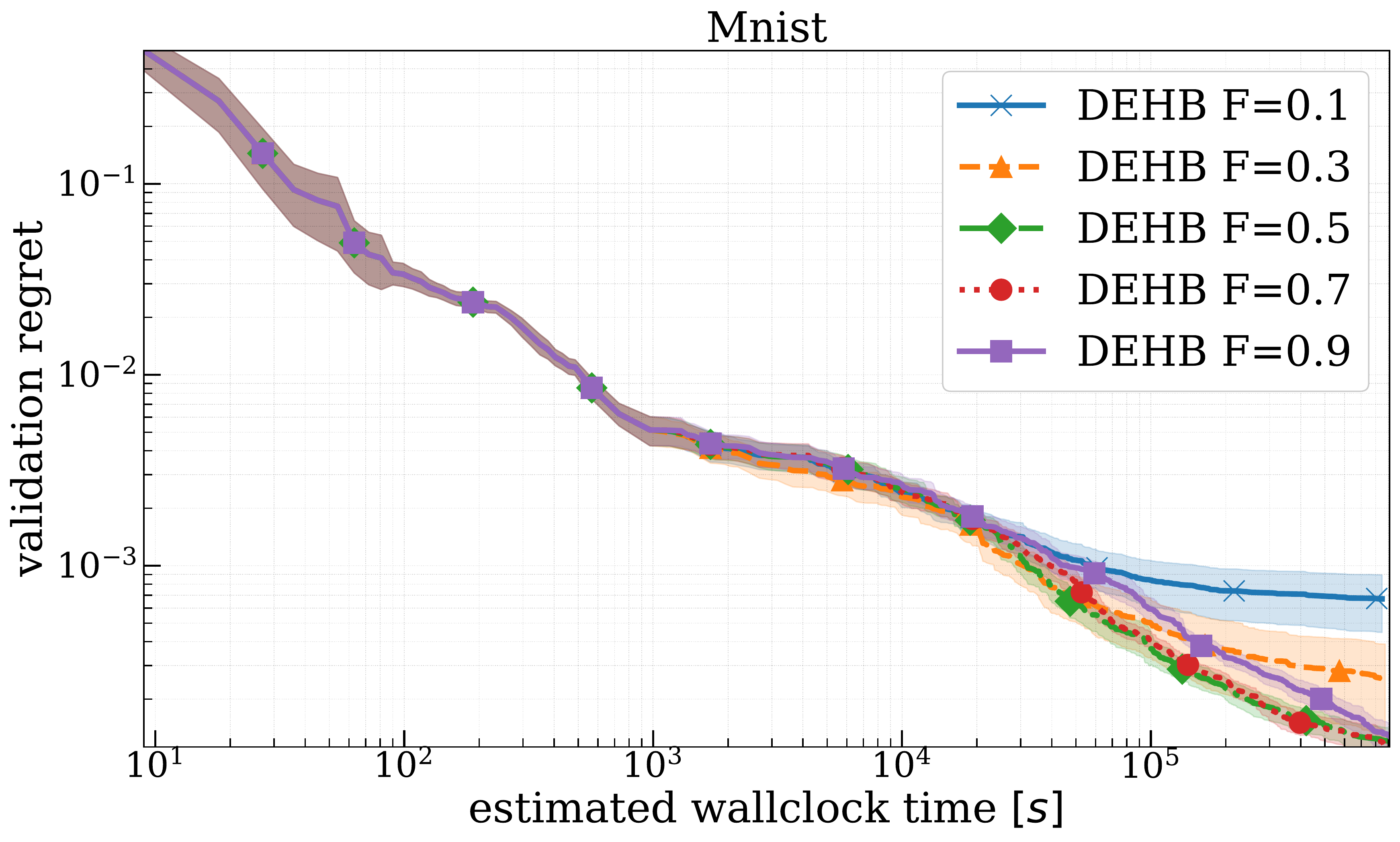} &
    \includegraphics[width=0.22\paperwidth]{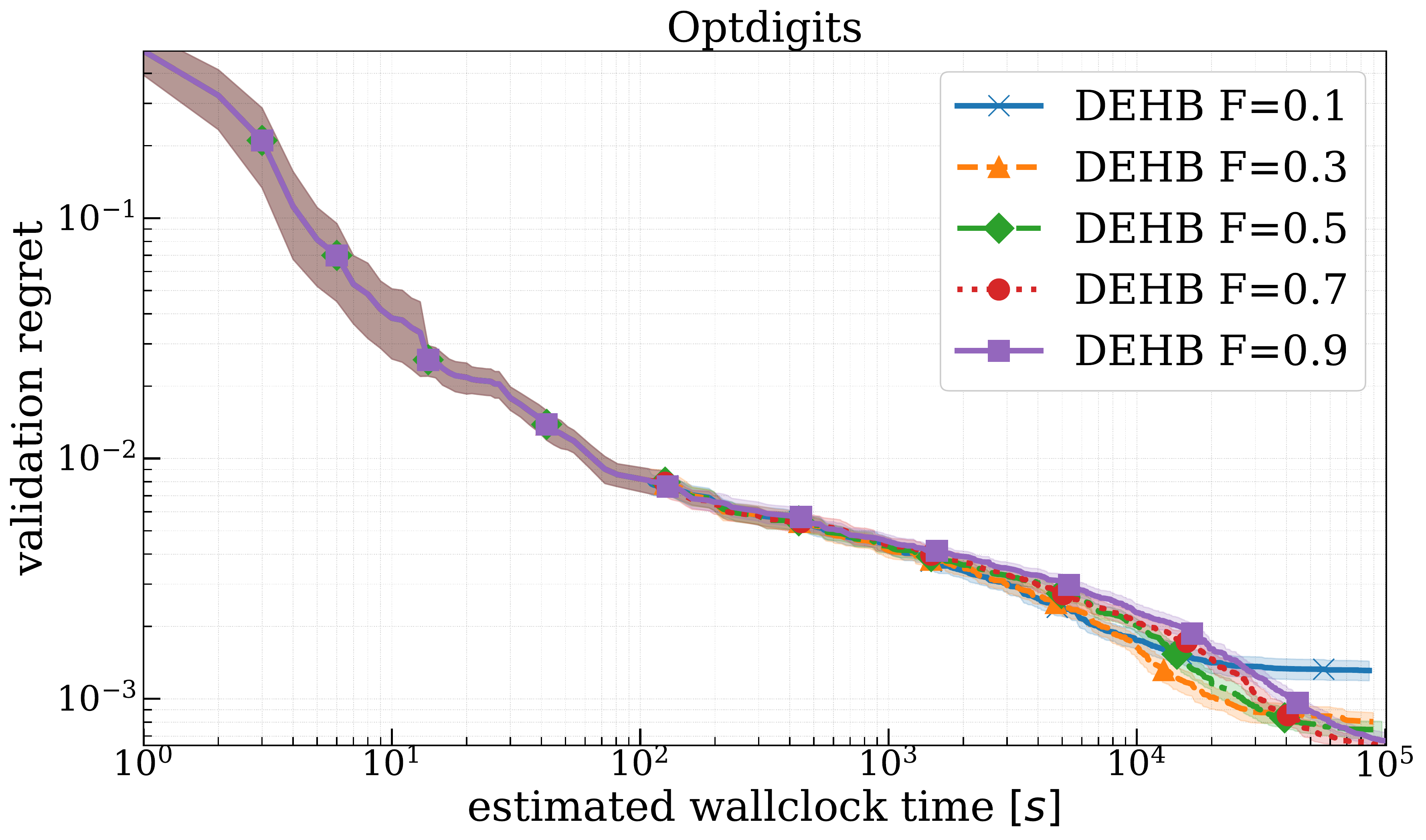} \\
    \includegraphics[width=0.22\paperwidth]{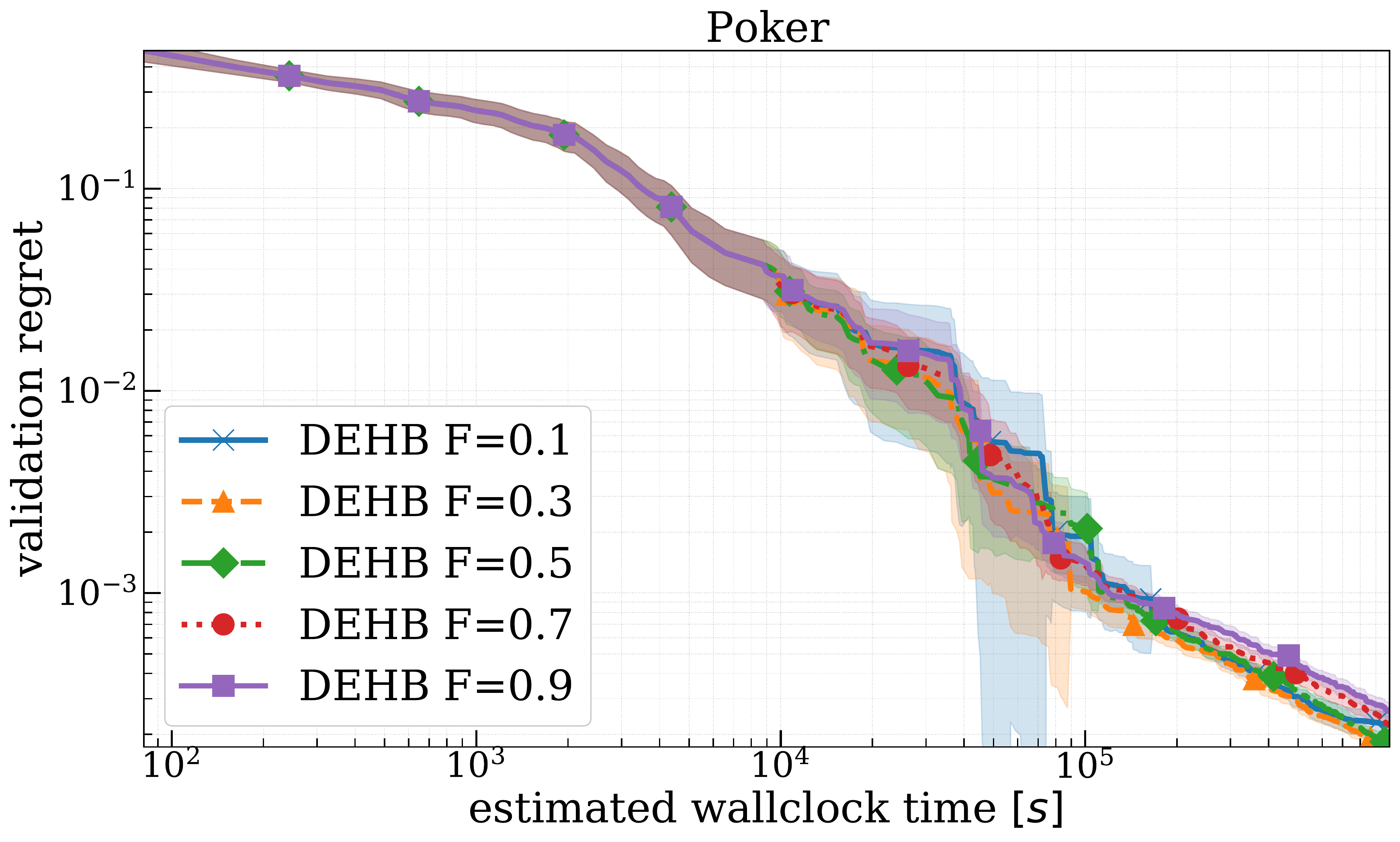} &
    \includegraphics[width=0.22\paperwidth]{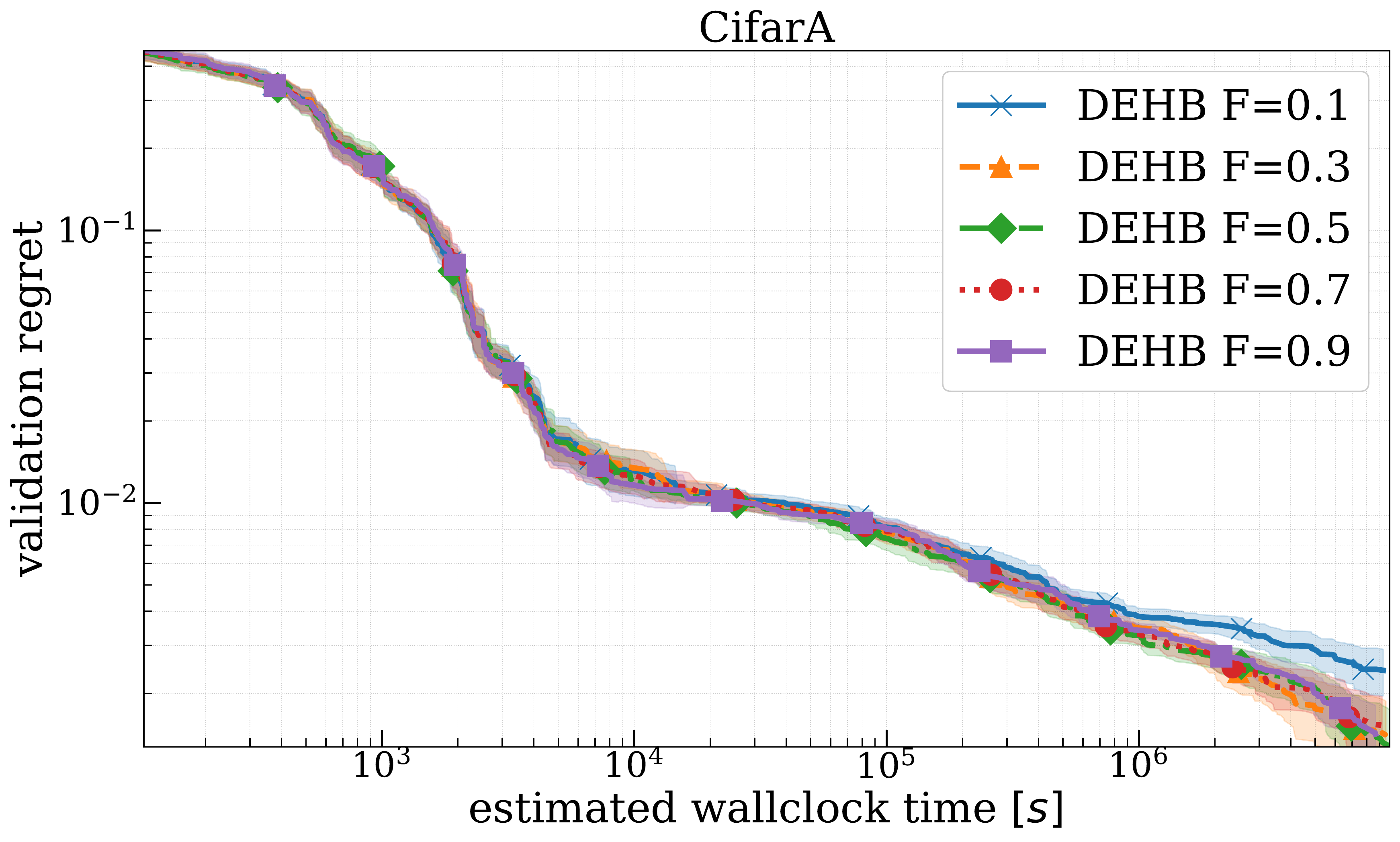} &
    \includegraphics[width=0.22\paperwidth]{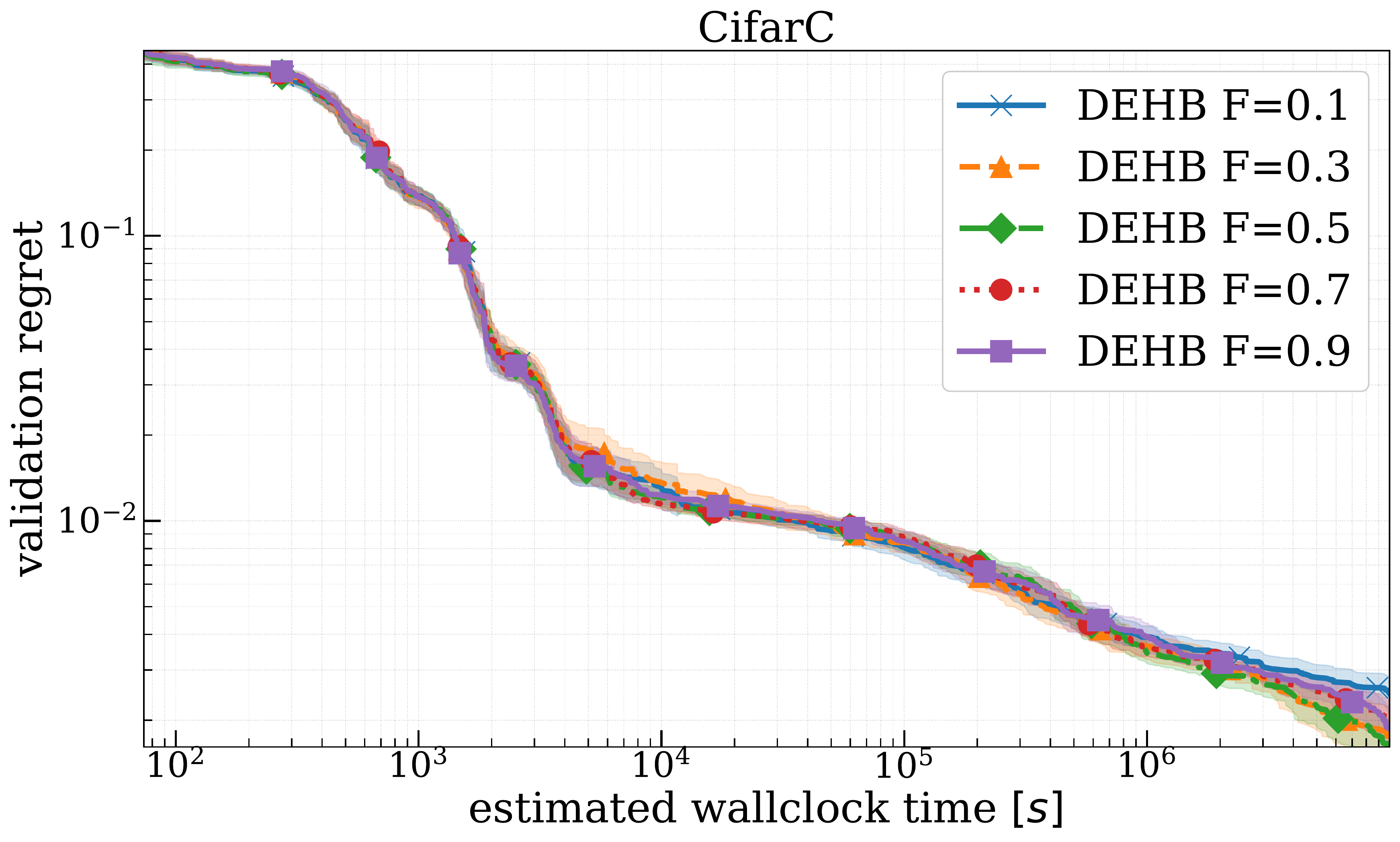} 
\end{tabular}
\caption{Ablation study for \textit{mutation factor} $F$ for DEHB, with \textit{crossover probability} fixed at $p=0.5$}
\label{fig:ablation-F}
\end{figure*}

The crossover probability $p$ was fixed at $0.5$, while mutation factor $F$ was varied with the values ${0.1, 0.3, 0.5, 0.7, 0.9}$. The studies are carried out on NAS-Bench-101, OpenML Surrogate and the toy Stochastic Counting Ones benchmarks. The results are reported in Figure \ref{fig:ablation-F}.

We observe that a low $F$ of $0.1$ allows more exploitative power to the DE search for a well correlated benchmark such as Counting Ones, while $F=0.9$ performs the worst. However, for the other benchmarks $F=0.1$ performs the worst with all other $F$ performing similarly. As a result we choose the conservative option of $F=0.5$ to ensure one general setting performs acceptably across all benchmarks.

\subsection{Varying CR}

\begin{figure*}[!ht]
\centering
\begin{tabular}{lll}
    \includegraphics[width=0.22\paperwidth]{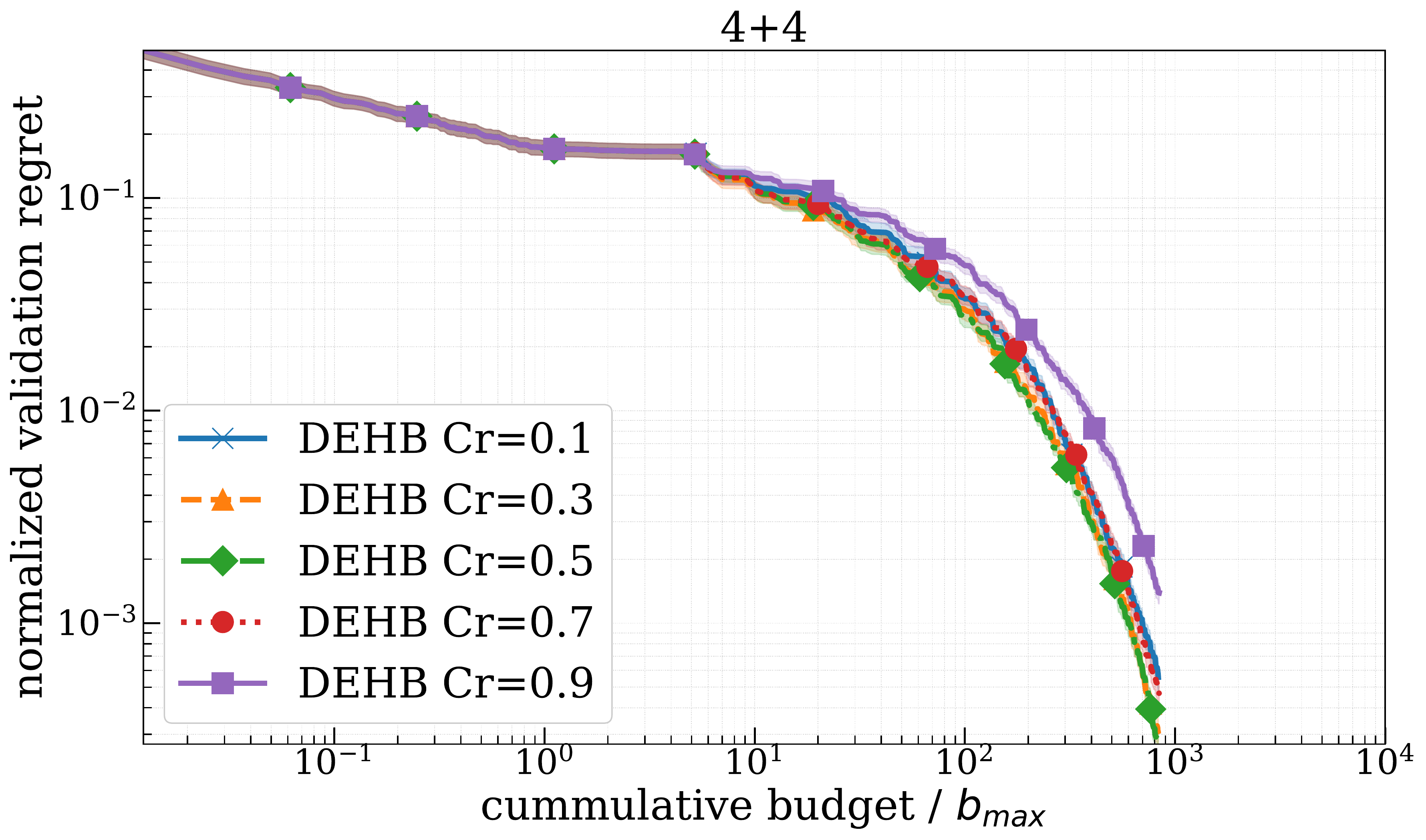} &
    \includegraphics[width=0.22\paperwidth]{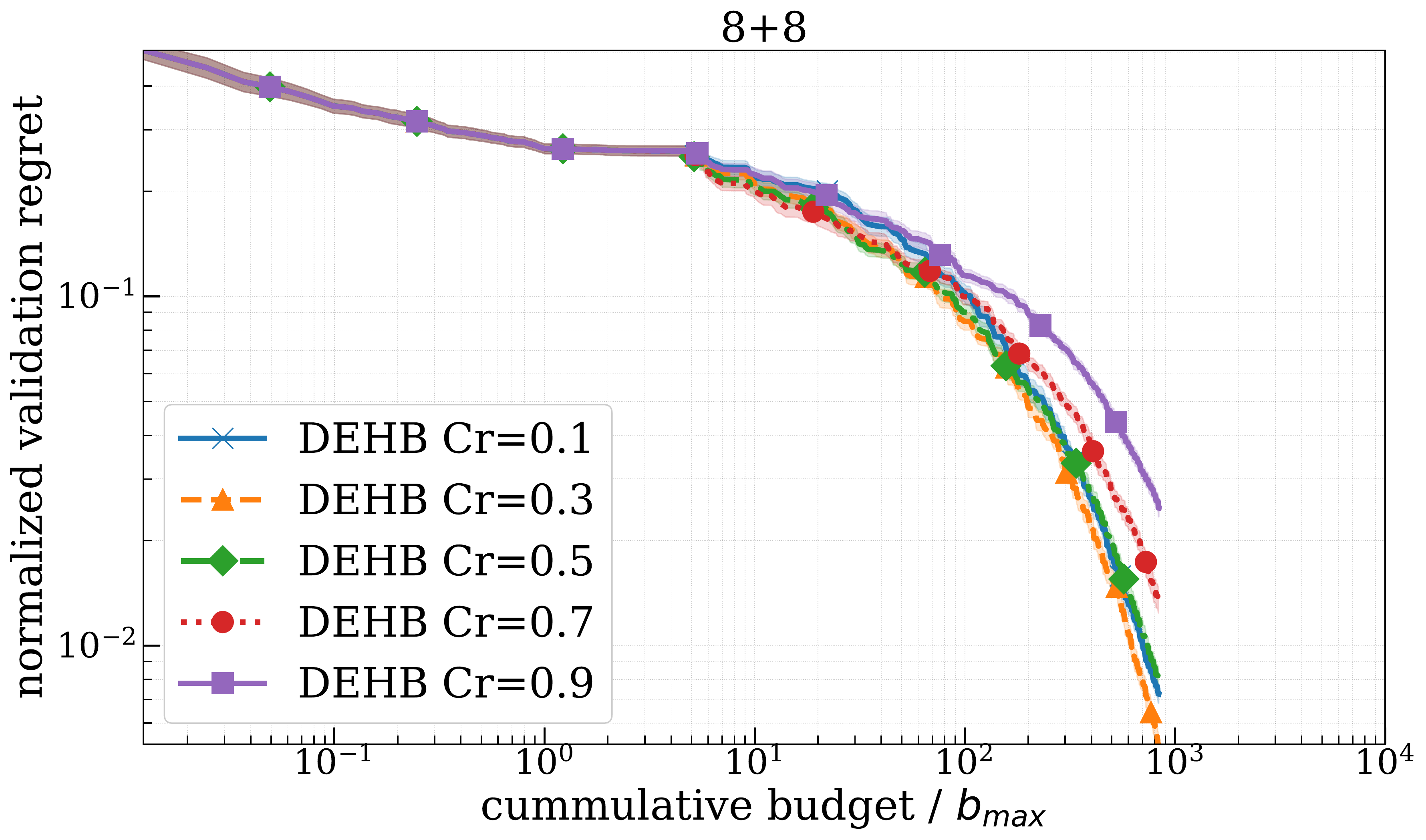} &
    \includegraphics[width=0.22\paperwidth]{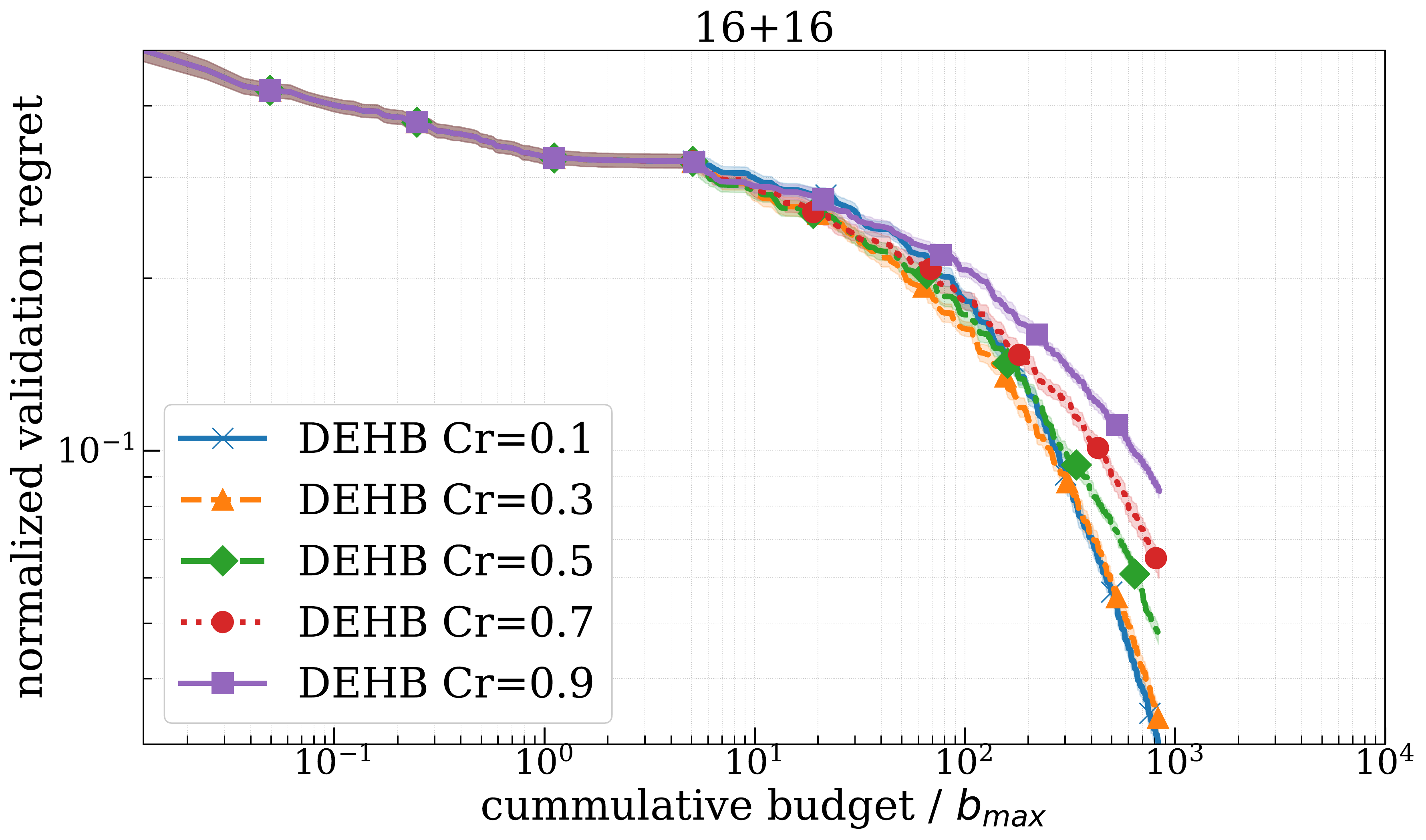} \\
    \includegraphics[width=0.22\paperwidth]{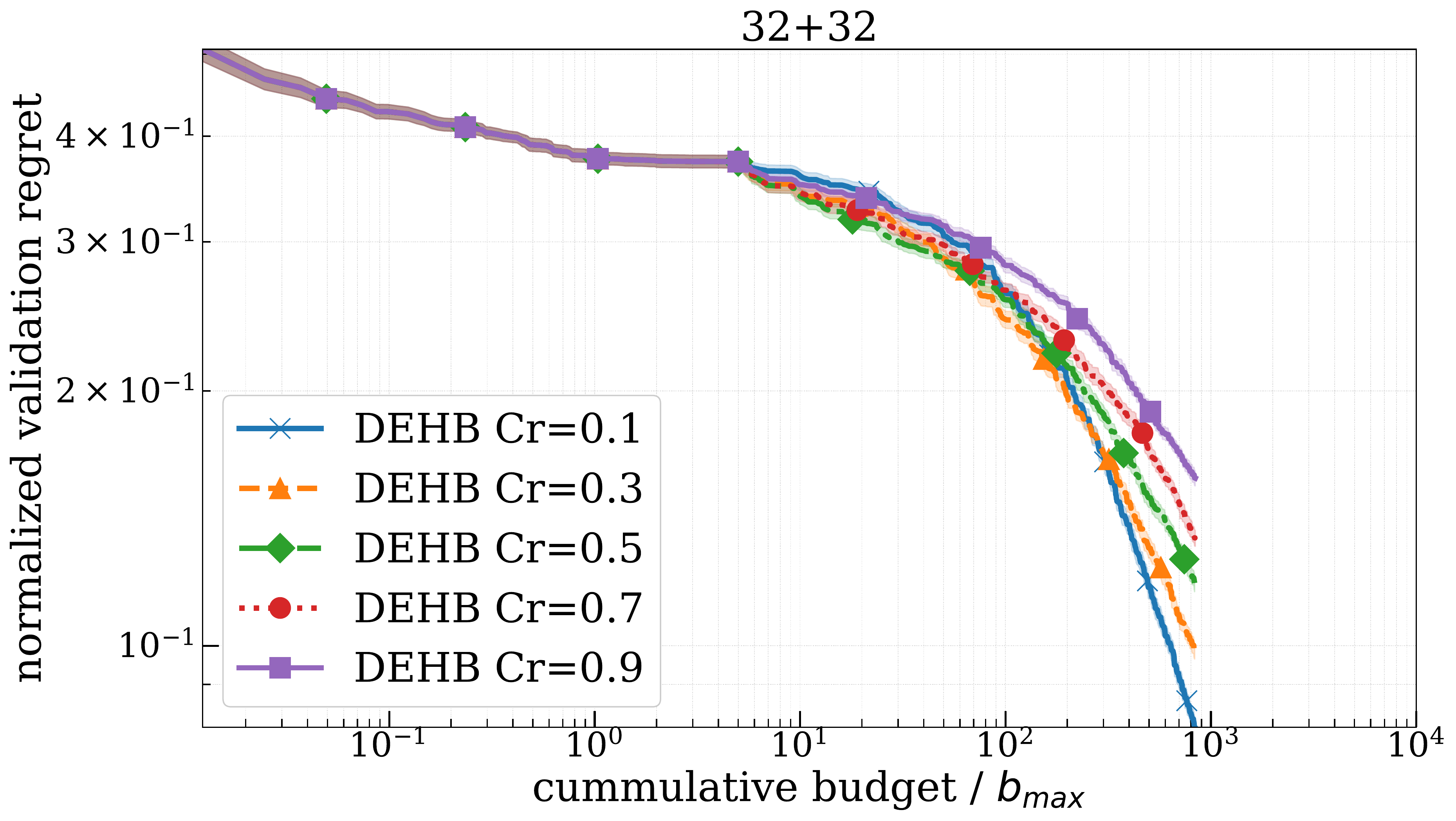} &
    \includegraphics[width=0.22\paperwidth]{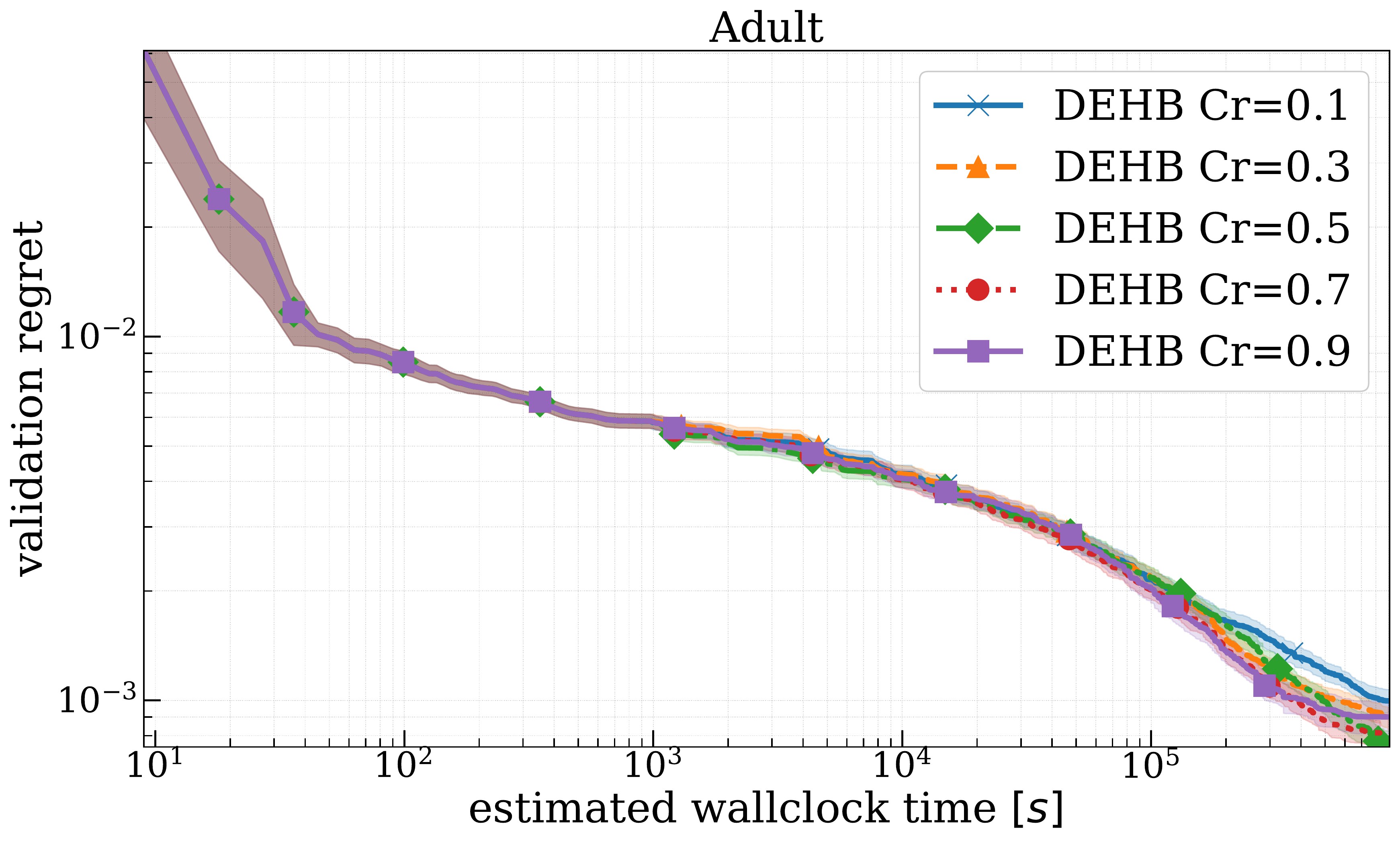} &
    \includegraphics[width=0.22\paperwidth]{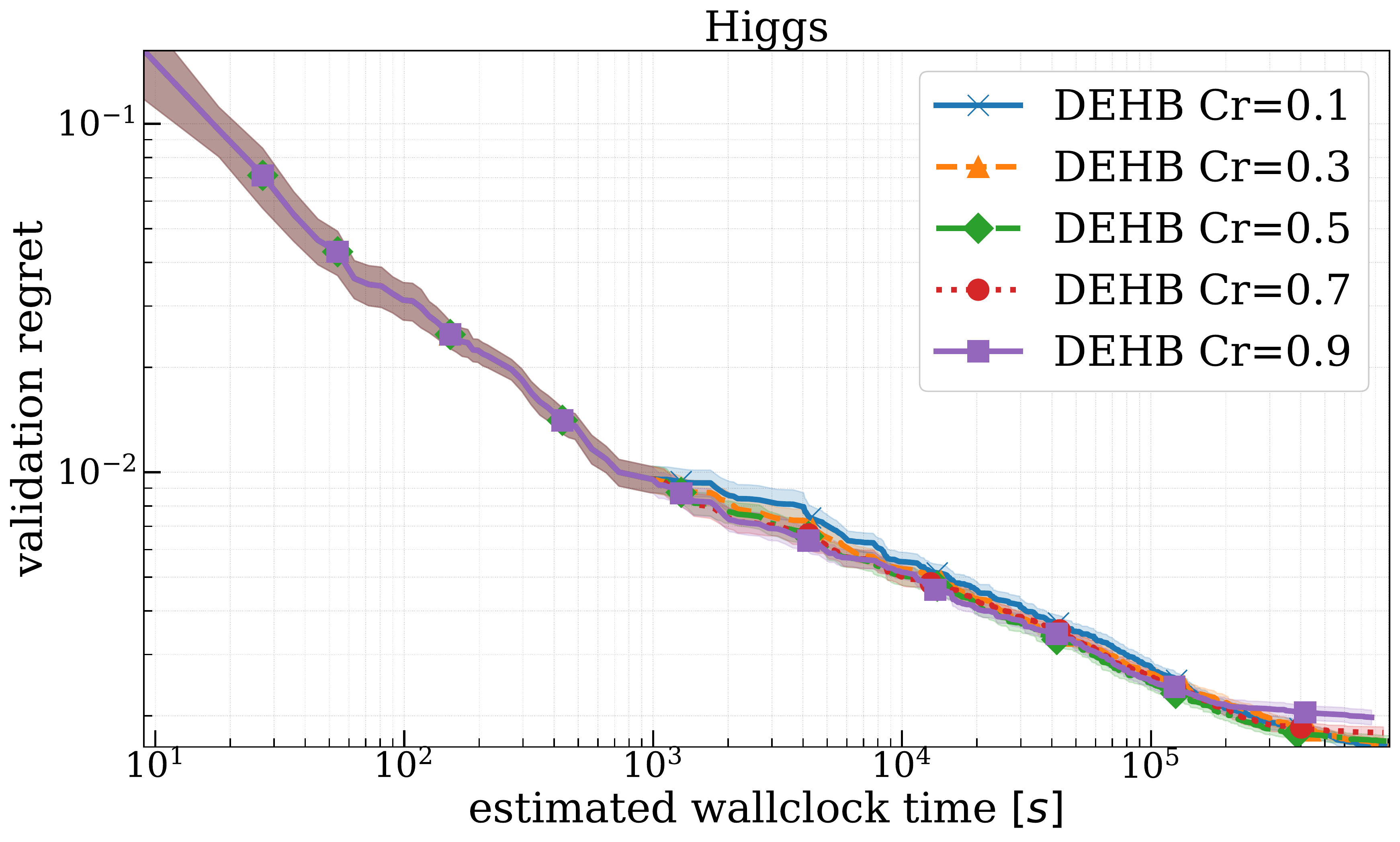} \\
    \includegraphics[width=0.22\paperwidth]{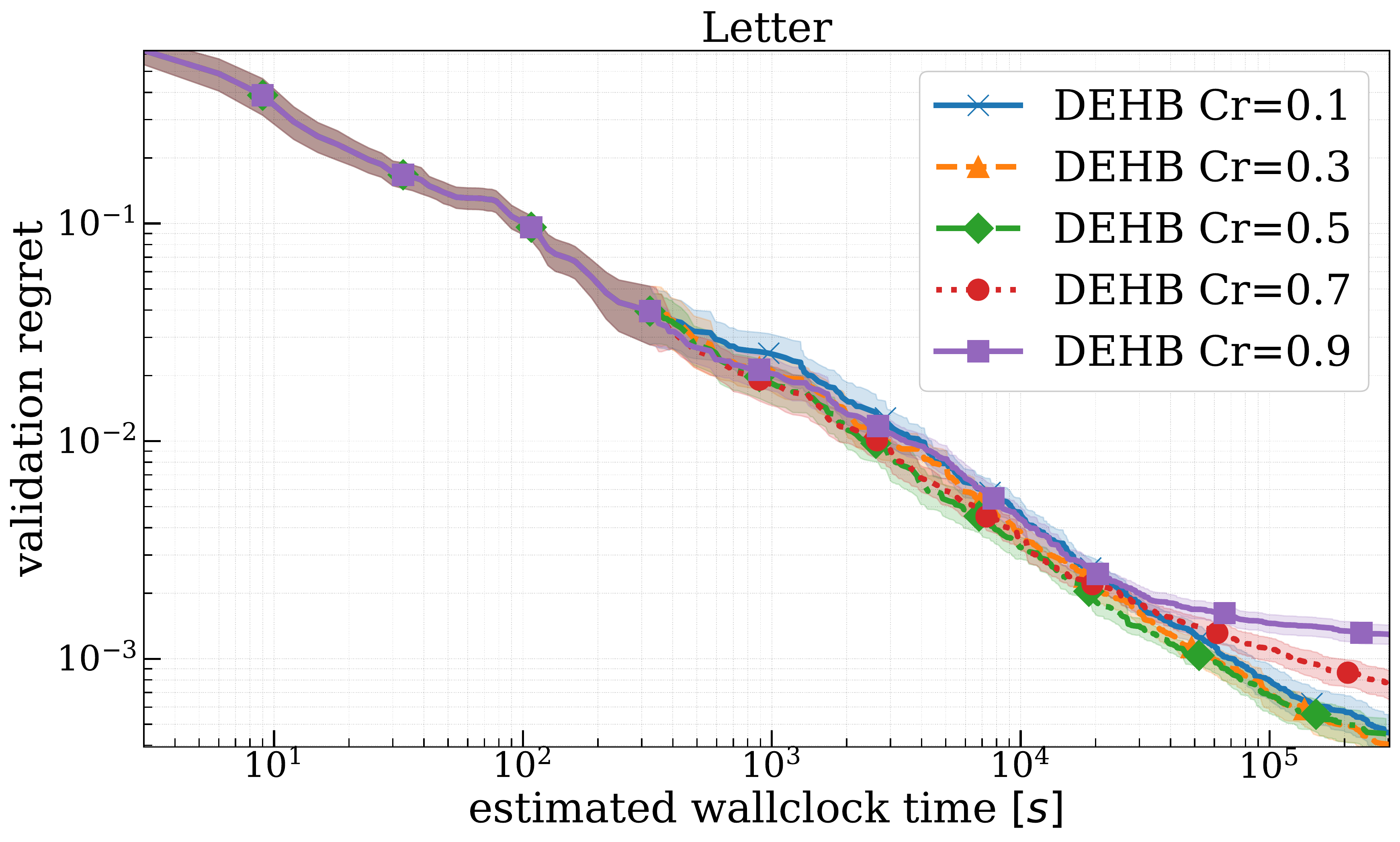} &
    \includegraphics[width=0.22\paperwidth]{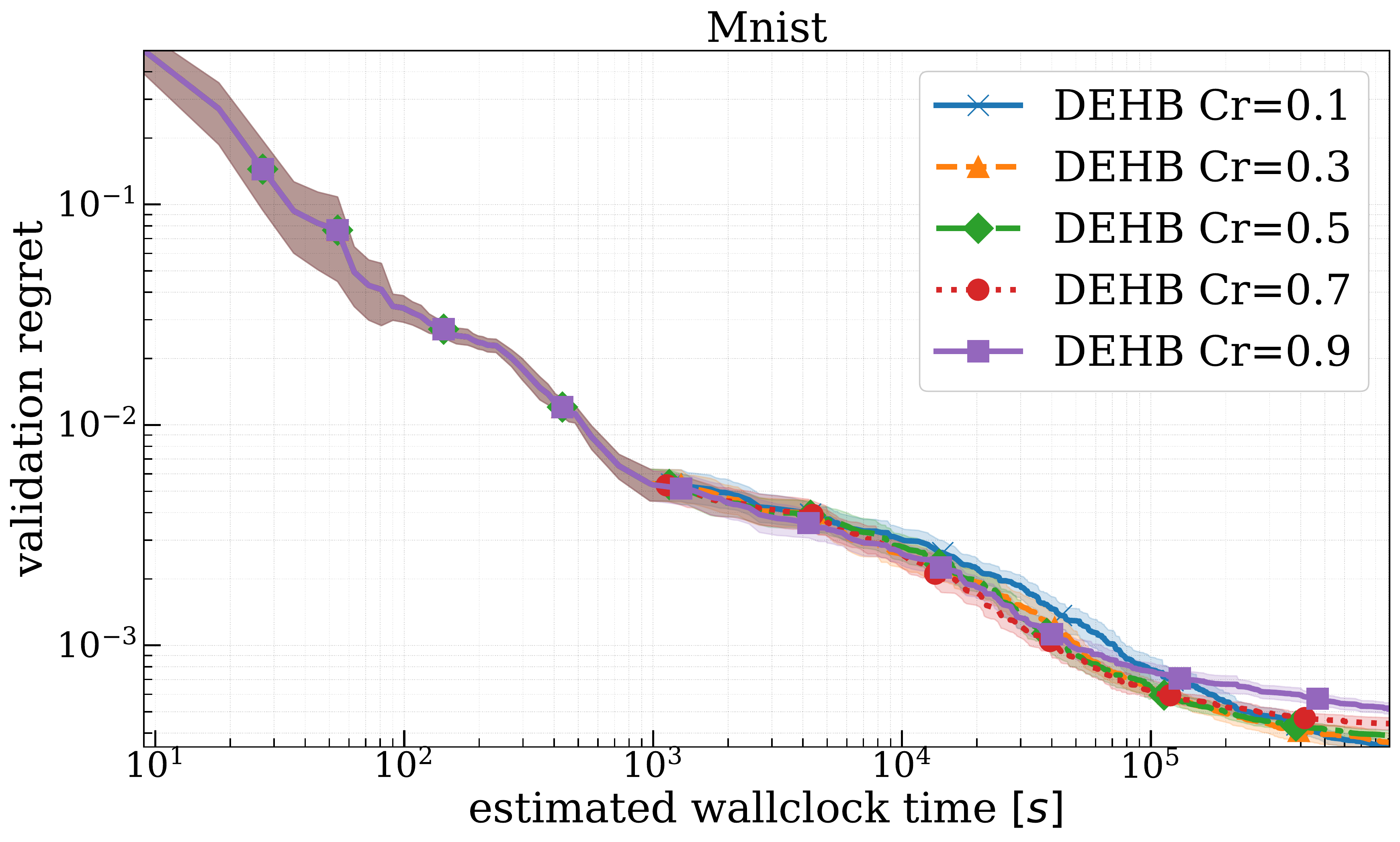} &
    \includegraphics[width=0.22\paperwidth]{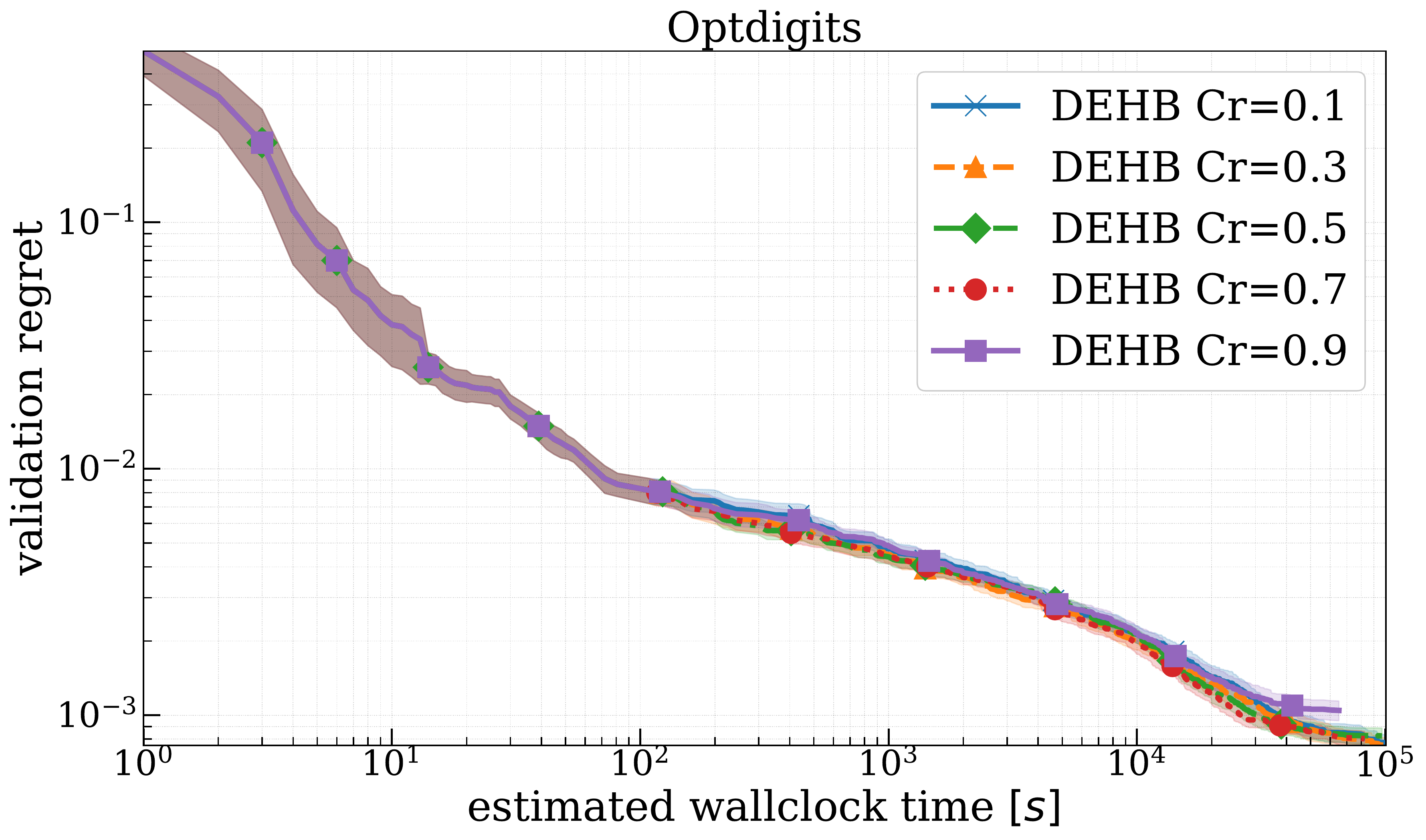} \\
    \includegraphics[width=0.22\paperwidth]{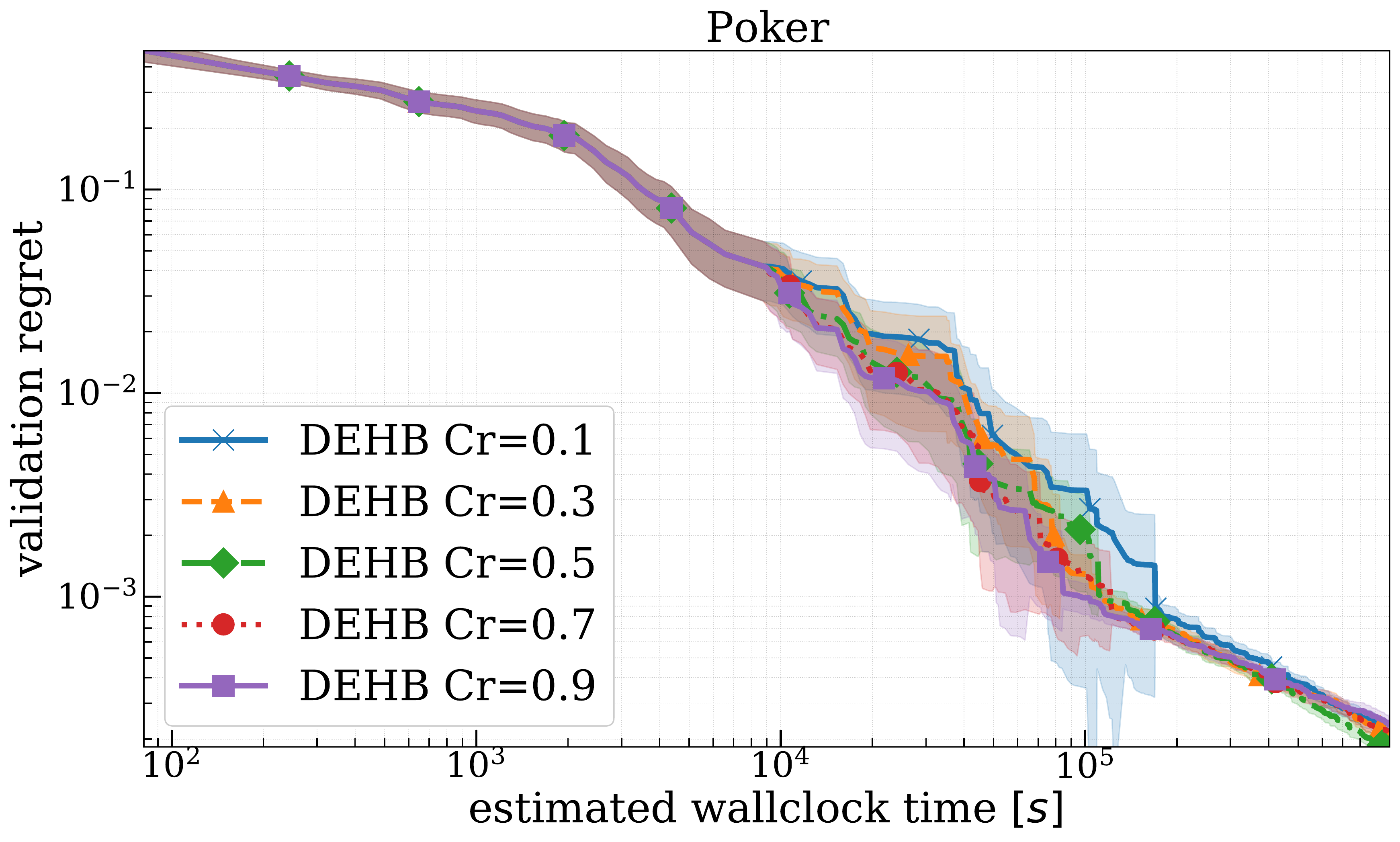} &
    \includegraphics[width=0.22\paperwidth]{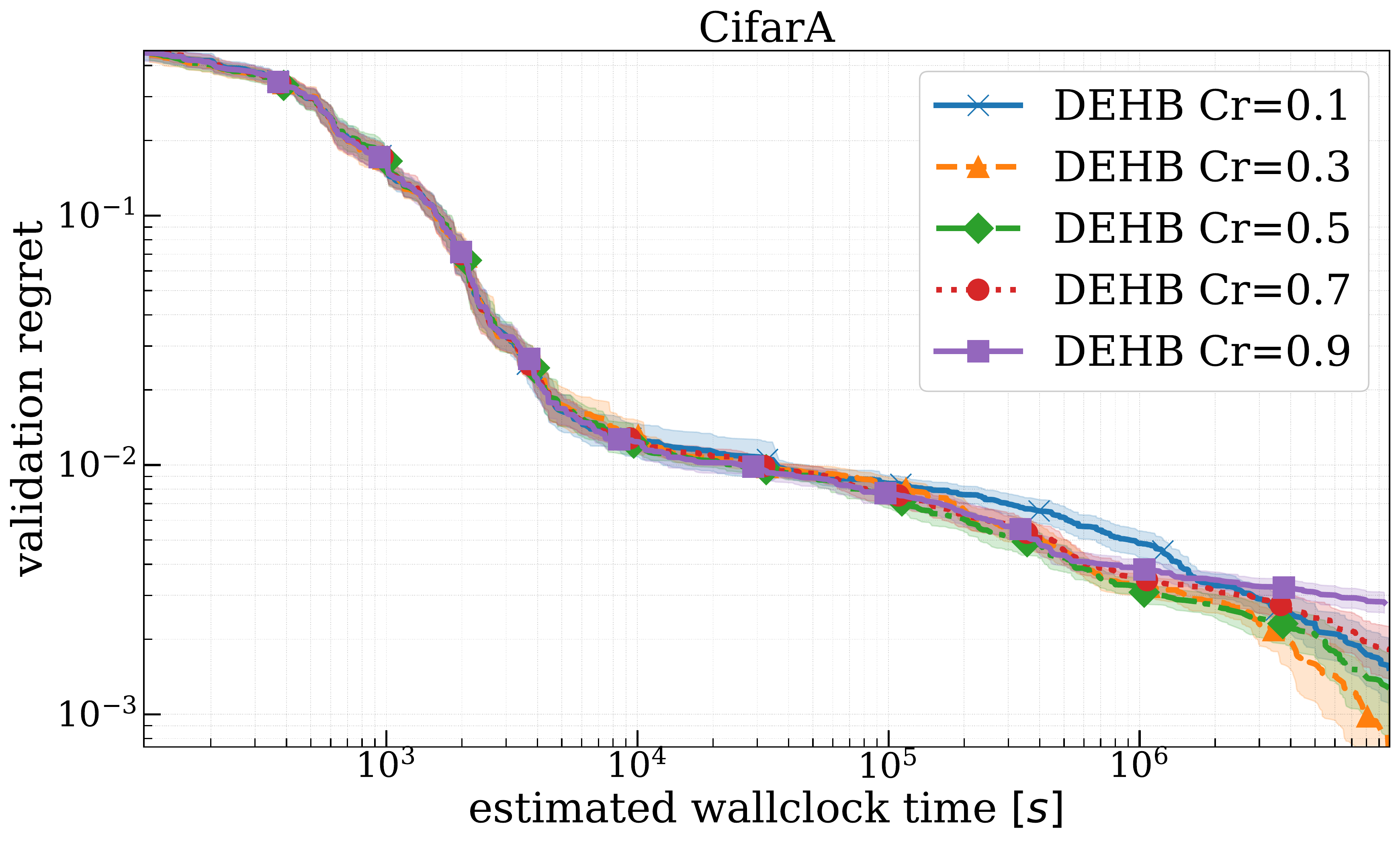} &
    \includegraphics[width=0.22\paperwidth]{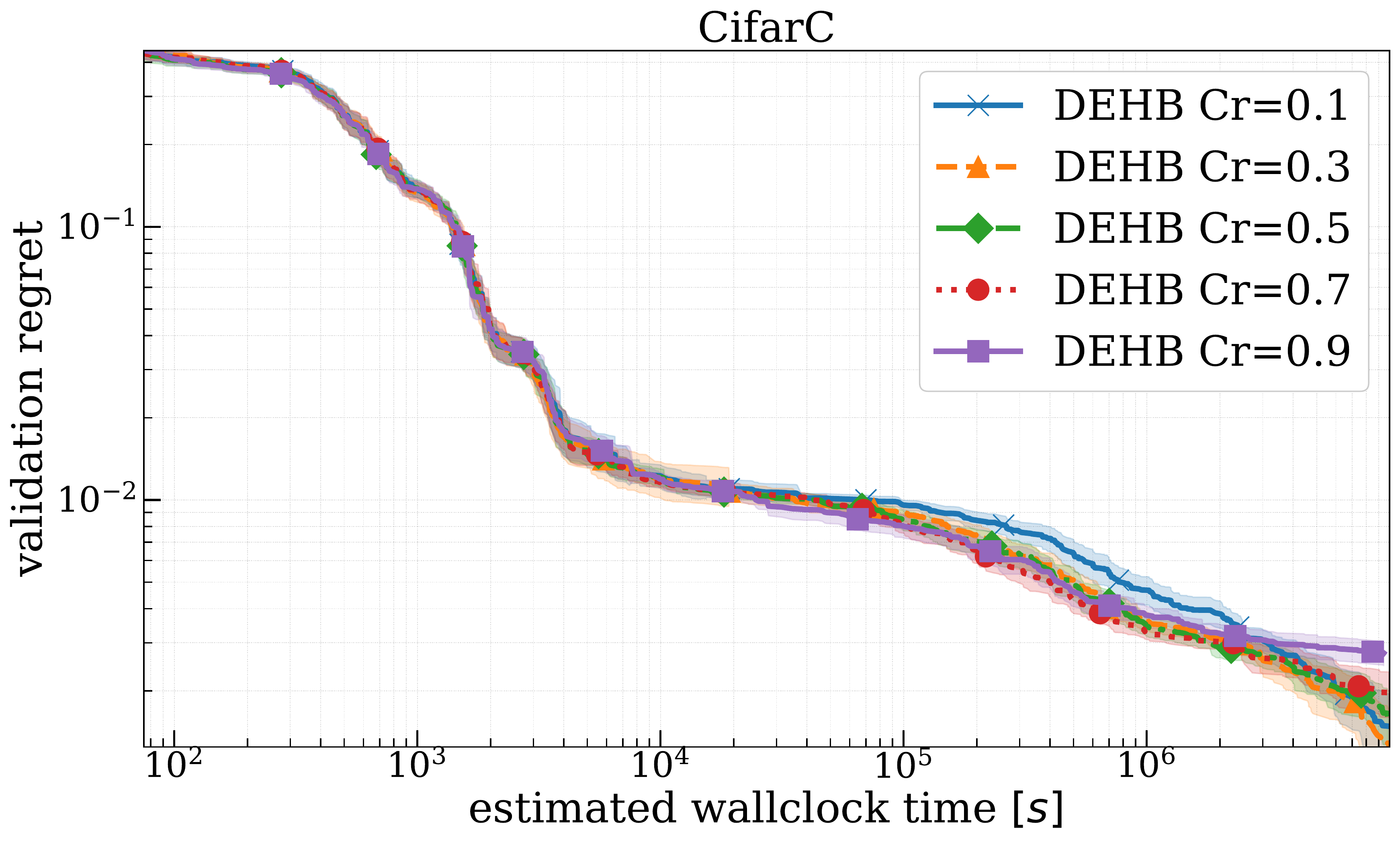} 
\end{tabular}
\caption{Ablation study for \textit{crossover probability} $p$ for DEHB, with \textit{mutation factor} fixed at $F=0.5$}
\label{fig:ablation-Cr}
\end{figure*}

The scaling factor $F$ was fixed at $0.5$, while crossover factor $p$ was varied with the values ${0.1, 0.3, 0.5, 0.7, 0.9}$. The studies were carried out on NAS-Bench-101, OpenML Surrogate and the toy Stochastic Counting Ones benchmarks. The results are reported in Figure \ref{fig:ablation-Cr}.

The lower the $p$ value, the less likely are the random mutant traits to be incorporated into the population. For Counting Ones, we observe that a high $p$ slows down convergence, whereas low $p$ speeds up convergence. However, for the others, $p=0.5$ is consistently the best performer. Hence, we chose $p=0.5$ for the design of DEHB.

|



\end{document}